\documentclass[twocolumn,switch,times]{article}

\usepackage{a4wide}
\usepackage[section]{placeins}
\usepackage[utf8]{inputenc}
\usepackage[T1]{fontenc}
\usepackage{titlesec}
\titlespacing\section{0pt}{12pt plus 3pt minus 3pt}{1pt plus 1pt minus 1pt}
\titlespacing\subsection{0pt}{10pt plus 3pt minus 3pt}{1pt plus 1pt minus 1pt}
\titlespacing\subsubsection{0pt}{8pt plus 3pt minus 3pt}{1pt plus 1pt minus 1pt}

\usepackage{graphicx}
\usepackage{caption}
\usepackage{subcaption}
\usepackage{float}
\usepackage{newfloat}
    \DeclareFloatingEnvironment[name={Supplementary Figure}]{suppfigure}
\usepackage{sidecap}
    \sidecaptionvpos{figure}{c}
\usepackage[font=small]{caption}
\usepackage{etoolbox}

\usepackage{booktabs}
\usepackage{multirow}
\usepackage{tikz}
\usepackage{ctable}
\usepackage{manyfoot}
\usepackage{makecell}
\usepackage{threeparttable}
\usepackage{rotating}
\usepackage{multicol}

\usepackage{amsmath}
\usepackage{amssymb}
\usepackage{bm}
\usepackage{gensymb}
\usepackage{bbm}

\usepackage[authoryear, round]{natbib}
\usepackage{fancyhdr}
\usepackage{preprint}
\usepackage{soul}
\usepackage{lineno}
\usepackage{eurosym}
\usepackage{color}
    \definecolor{red}{rgb}{1, 0, 0}
    \definecolor{green}{rgb}{0, 1, 0}
    \definecolor{blue}{rgb}{0, 0, 1}

\usepackage[hidelinks]{hyperref}
    \hypersetup{
        colorlinks=true,
        linkcolor=black,
        urlcolor=cyan,
        citecolor=cyan,
        anchorcolor=black
    }
\usepackage[title]{appendix}
    
\usepackage{authblk}

\usepackage{placeins}

\title{A Framework for Individual Tree Growth Reconstruction Using Multi-Platform Laser Scanning}

\author[1,\small\textdagger,$^{*}$]{Daniella Tavi}
\author[1,\small\textdagger]{Valtteri Soininen}

\author[1,2]{Lassi Ruoppa}
\author[1]{Jesse Muhojoki}
\author[1]{Juha Hyyppä}

\affil[1]{Department of Remote Sensing and Photogrammetry, Finnish Geospatial Research Institute FGI, The National Land Survey of Finland, Vuorimiehentie 5, Espoo, FI-02150, Finland}
\affil[2]{Department of Computer Science, School of Science, Aalto University, P.O. Box 11000, Aalto, FI-00076, Finland}

\begin{document}

\twocolumn[\begin{@twocolumnfalse}
    \maketitle
    \begin{abstract}

    Accurate individual tree-level forest monitoring using laser scanning data requires reliable tree delineation, consistent tree correspondence across multitemporal point clouds, and accurate estimation of tree attributes and their change. Reconstructing tree growth in boreal forests is challenging due to the scarcity of historical stem-level data, propagation of errors from older sensors into change estimation, and growth rates with a magnitude of measurement uncertainty. This study investigates a framework for estimating individual tree diameter at breast height (DBH) and stem volume growth using 136 point clouds acquired between 2014--2025 with 11 scanners on airborne (ALS), mobile (MLS), and terrestrial laser scanning (TLS) platforms across boreal forest test sites. Trees were delineated from an MLS point cloud using deep learning-based segmentation which was then transferred to the remaining point clouds, resulting in reliable multitemporal tree correspondence. Stem curves were derived from MLS/TLS data, with ALS data used for height estimation, enabling DBH and volume estimation and time series. A height growth-based scaling model was used to reconstruct stem attributes across time and estimate growth. Results showed that modeled growth achieved higher agreement with manual growth estimates than differencing independently estimated attributes from point clouds. The modeled-manual 5- and 10-year growth RMSEs were 55--111\% and 26--67\% for DBH, and 31--87\% and 21--67\% for volume, respectively, depending on plot difficulty. The scaling model was temporally robust, with errors remaining stable or stabilizing after 5--6 years, reaching maximum RMSEs of 8--12\% for DBH and 12--23\% for volume after 12 years. Combining MLS/TLS-derived stem measurements with multitemporal ALS-derived heights provided a robust framework for individual tree growth estimation without requiring multiple under-canopy scans.
    
    \end{abstract}

    \keywords{Individual tree, Change detection, Forest monitoring, Time series, Airborne laser scanning, Terrestrial laser scanning, Mobile laser scanning}
    \vspace{0.5cm}
\end{@twocolumnfalse}]

\renewcommand{\thefootnote}{\small\ensuremath{\ast}}
\footnotetext[1]{Corresponding author}
\renewcommand{\thefootnote}{}
\footnotetext[2]{\textit{Email addresses}: \newline daniella.tavi@nls.fi (Daniella Tavi),  valtteri.soininen@nls.fi (Valtteri Soininen), lassi.ruoppa@nls.fi (Lassi Ruoppa), jesse.muhojoki@nls.fi (Jesse Muhojoki), juha.hyyppa@nls.fi (Juha Hyyppä) }
\renewcommand{\thefootnote}{\small\textsuperscript{\textdagger}}
\footnotetext[3]{These authors contributed equally.}

\renewcommand{\thefootnote}{\arabic{footnote}}
\setcounter{footnote}{0}

\section{Introduction}
Forest monitoring provides up-to-date information on forest structure and change. Accurate information on forest state is essential for sustainable forest management, biodiversity conservation, carbon accounting, and understanding forest responses to environmental change \citep{kettunen2012socio, fridman2014adapting, landsberg2011physiological}. Light detection and ranging (LiDAR) data now enable forest monitoring at the individual tree level, enabling the derivation of attributes such as tree height, volume, and trunk dimensions directly from point clouds \citep{Hyypp1999, zhen2016trends}. Tree-level information derived from LiDAR also supports a wide range of forest monitoring and management applications. For example, it can be used for dead wood assessment \citep{Marchi2018}, logging monitoring \citep{Vastaranta2014, coops2023framework}, urban tree detection \citep{Wallace2021, Ruoppa2026, haala1999extraction}, and tree species classification \citep{Taher2026, persson2004tree}. It is also essential for estimating above-ground biomass, quantifying carbon sequestration, and supporting operational forest inventories and long-term monitoring of carbon sinks. As a result, high-quality tree-level information has become an increasingly important component of forest management and decision-making \citep{Perry2026, Keefe2022}.

The increasing availability of repeated airborne LiDAR acquisitions has further expanded opportunities for monitoring forest change beyond single-date characterization \citep{yu2005measuring, yu2006change, yu2004automatic}. National-scale laser scanning programs are evolving toward denser and more frequent acquisitions \citep{Hyypp2024}, enabling more detailed analyses of forest dynamics and supporting the transition from stand-level inventories toward operational individual-tree-based forest monitoring. In addition, multitemporal LiDAR data enable monitoring tree growth \citep{yu2005measuring, duncanson2018monitoring}, mortality \citep{ni2025remote, bueno2025aboveground}, and responses to environmental disturbances \citep{atkins2020application, duncanson2018monitoring}. Overall, such developments strengthen the role of LiDAR-based approaches in precision forestry, long-term forest monitoring, and improving the understanding of forest dynamics under changing environmental and climatic conditions.

Delineating individual trees from point clouds depicting forest environments, commonly referred to as individual tree segmentation (ITS), is a prerequisite for tree-level forest monitoring \citep{Hyypp1999}. As ITS is also essential for a wide range of other applications, it remains one of the most extensively studied segmentation tasks in forestry \citep{Hyyppa2001,Kaartinen2012, Aubry-Kientz2019, Cao2023, Ruoppa2026, vauhkonen2012comparative, eysn2015benchmark, persson2002detecting}. Traditionally, ITS has relied on unsupervised algorithms that delineate trees based on a set of manually defined heuristics. These algorithms can be broadly divided into two categories based on whether they operate on 2D or 3D representations. Most 2D approaches project the point cloud onto the $xy$-plane to create a raster image, and subsequently delineate trees using local maxima combined with watershed or region growing \citep{Hyyppa2001, Koch2006, Yu2011, Dalponte2016}. In contrast, 3D methods comprise a more diverse set of approaches, including clustering \citep{Lee2010, Ferraz2016}, graph-based segmentation \citep{Strimbu2015, Xi2022}, and hybrid strategies that combine 2D and 3D processing \citep{Wang2008, Duncanson2014, Ayrey2017}. Although 3D ITS algorithms are generally more accurate than 2D approaches \citep{Hakula2023, Cao2023}, they are also significantly more computationally demanding \citep{Ruoppa2026}. Consequently, 2D methods have remained popular in large-area ITS \citep{Hyypp2024}.

Following the success of deep learning in several other computer vision tasks, recent ITS research has largely shifted toward deep learning-based approaches, which generally outperform unsupervised algorithms by a significant margin \citep{Xiang2024, Ruoppa2026}. Early studies typically either transformed point clouds into 2D depth images and applied image segmentation models \citep[see e.g.][]{Windrim2019, Straker2023}, or combined deep learning with conventional heuristic-based algorithms \citep[see e.g.][]{Krisanski2021, Wielgosz2023}. More recently, several fully deep learning-based end-to-end 3D segmentation frameworks have been proposed, with most works opting for a grouping-based approach \citep{Xiang2024, Wielgosz2024, Henrich2024, Xi2025}. An alternative end-to-end paradigm is 3D mask transformers \citep{Xiang2025a, Liu2026}, which set the current state of the art on several ITS benchmarks \citep{Xiang2025a, Ruoppa2026}.

One technique for improving the consistency and efficiency of ITS is the use of prior tree location information, or tree maps \citep{Vastaranta2014, Honkanen2025, Cao2023}. Incorporating prior tree maps can improve the consistency of individual tree detection across different data sources, reduce the processing time of ITS \citep{Honkanen2025}, and help mitigate missing trees in the analyses \citep{Vastaranta2014}. \citet{Honkanen2025} applied a kd-tree-based segmentation transfer approach, similar to that used in this study, in which segment labels were assigned based on the nearest neighboring points. Their method reduced running time by 95\% compared to segmenting the point cloud independently, while also improving segmentation results. \citet{Cao2023} generated an ALS segmentation dataset by transferring segments from TLS point clouds to ALS point clouds. Segmentation transfer is particularly relevant for multitemporal forest monitoring, as it can improve tree detection in sparse or occluded point clouds by leveraging segmentations derived from higher quality datasets. This is especially important for sparse airborne laser scanning (ALS) point clouds, where canopy occlusion and lower point densities limit tree detection performance compared to under-canopy acquisitions or sensors producing denser point clouds. In this context, mobile laser scanning (MLS) and terrestrial laser scanning (TLS) appear well-suited as source datasets for segmentation transfer. However, segmentation transfer using prior knowledge of tree counts and locations has received considerably less attention than research on segmentation methods themselves.

Once individual trees are successfully delineated, the focus shifts to determining the state of their attributes at a given point in time and using time series information to estimate tree growth. This requires reliable single-time measurements or estimates of attributes derived from individual trees. Among the most commonly measured attributes are geometric characteristics such as tree height, which has been estimated from point clouds for several decades \citep{Hyypp1999, zhen2016trends}. Other geometric attributes, including diameter at breast height (DBH) and stem volume, can be estimated using statistical models or machine learning \citep[e.g.][]{Hyypp1999, Yu2011, Windrim2020, persson2002detecting, parkan2019combined}. Currently, direct extraction of stem geometry from point clouds is possible and enables DBH, stem curves, and stem volume to be measured directly from point cloud-derived tree stems. These point clouds can be collected using a variety of LiDAR platforms, including above-canopy carriers \citep{hyyppa2022direct, vandendaele2021estimation, kuvzelka2020very, brede2017comparing, wieser2017case}, under-canopy uncrewed aerial vehicles (UAVs) \citep{hyyppa2020under, hyyppa2021under, muhojoki2024benchmarking}, and ground-based terrestrial and mobile systems \citep{Hyypp2020, hyyppa2020comparison, muhojoki2024benchmarking, bienert2007tree, oveland2018comparing, lovell2003using, li2023use, olofsson2016single, thies2004three, cabo2018automatic, huang2011automated, bauwens2016forest, chen2019applicability, ryding2015assessing, maas2008automatic}. Direct stem measurements are particularly attractive for individual tree monitoring, because they reduce the dependence on predictive models and the uncertainties associated with model selection and calibration.

Different LiDAR acquisition methods provide complementary strengths for individual tree attribute estimation. Under-canopy approaches, such as TLS and MLS, capture stems with higher point density and reduced occlusion, enabling accurate characterization of stem geometry and DBH. Previous studies have reported DBH and stem curve accuracies of approximately 1--2 cm when using TLS \citep{lovell2003using, liang2012detecting, liang2013automatic, liang2018international, olofsson2016single, cabo2018automatic}. Comparable accuracies have also been achieved under favorable conditions using MLS, with reported DBH RMSE values ranging from 0.6--3.1 cm \citep{Tavi2026, muhojoki2024benchmarking, gollob2020forest, hyyppa2020under, hyyppa2020comparison, giannetti2018integrating, oveland2018comparing, bauwens2016forest, chen2019applicability, ryding2015assessing}, and stem curve accuracies below 2 cm \citep{muhojoki2024benchmarking, hyyppa2020under, hyyppa2020comparison}. In addition, repeated MLS measurements have demonstrated DBH precisions of approximately 0.3--0.5 cm, indicating high repeatability and reliability of stem-level measurements \citep{Tavi2026}.

In contrast, ALS is particularly effective for large-scale forest monitoring due to its efficiency in covering vast areas quickly and the possibility of using Global Navigation Satellite System (GNSS) for obtaining georeferenced point clouds. ALS also excels in tree height estimation, with previous studies finding ALS-derived heights to be more accurate than height estimates from manual, MLS, or TLS methods \citep{liang2019forest, Wang2019, kankare2014accuracy}, especially for dominant and co-dominant trees. Under-canopy MLS and TLS methods often produce height estimates with larger errors, particularly for tall trees in dense stands, where occlusion from neighboring trees and limited treetop visibility can introduce negative bias and inaccurate height estimation \citep{Tavi2026, muhojoki2024benchmarking, Wang2019, liang2018international, kankare2014accuracy, maas2008automatic}. However, while ALS excels in height estimation, stem-level measurements are typically less accurate than those derived from MLS and TLS. This is mainly due to canopy occlusion, lower stem point densities, and larger beam footprints near the ground, with the lowest reported DBH and stem curve RMSE values being approximately 2--4 cm \citep{muhojoki2024benchmarking, hyyppa2022direct, brede2017comparing}. Moreover, ALS-based stem detection rates remain lower than those achieved using under-canopy MLS/TLS methods, while segmentation of small trees from ALS data can be challenging \citep{hyyppa2022direct, Donager2021, Ruoppa2026}. Consequently, ALS and under-canopy MLS/TLS systems provide complementary advantages for individual tree monitoring, suggesting that combining these acquisition methods may improve the robustness and accuracy of multitemporal forest monitoring frameworks.

Quantifying how these attributes change over time introduces a distinct set of approaches. Permanent laser scanning systems can be used to detect individual tree changes \citep{wittke2024liphestream}, destructive methods such as tree-ring width measurements can provide non-repeated measurements of tree growth \citep{mohammadi2018estimation}, and repeated manual field measurements using instruments such as tape measures, calipers, and clinometers \citep{Burkhart2012} can provide accurate individual tree growth measurements often used as reference. While these approaches can yield accurate growth estimates, they are generally time-consuming and labor-intensive, limited in spatial coverage, or unsuitable for frequent large-scale monitoring. Consequently, non-destructive approaches based on multitemporal LiDAR data have become increasingly important for scalable forest growth assessment. Several studies have demonstrated the feasibility of using two-date TLS measurements to quantify changes in individual tree stem attributes, including DBH, volume, taper, and branching patterns \citep{yrttimaa2023capturing, luoma2021revealing,luoma2019examining, sheppard2017terrestrial, kaasalainen2014change}, as well as studies using ALS data for forest change monitoring \citep{kozniewski2022tracking, ma2018quantifying, arumae2020thinning, zhao2018utility, yu2005measuring, yu2006change, Soininen2022, duncanson2018monitoring}. However, while attribute change can be detected reliably in many cases, comparatively few studies have systematically evaluated the accuracy of the resulting growth estimates.

Recent studies highlight both the potential and limitations of these approaches. \citet{wang2025forest} reported 5-year DBH change estimation errors of 0.50--0.82 cm (51\%) and height change errors of 0.57--0.64 m (48\%) compared to a manual reference using two-date TLS data in a managed, even-aged forest stand. Similarly using TLS data, \citet{yrttimaa2022exploring} reported a 5-year DBH change RMSE of 65\% with an $R^2$ of 0.58, and volume growth RMSE of 69\% and $R^2$ of 0.59. \citet{luoma2021revealing} further noted when studying TLS-based volume growth that TLS-derived height estimates can be a limiting factor. With ALS-based approaches, height growth of individual trees has been studied over varying time spans, with reported $R^2$ value of 0.29 for growth over 2 years \citep{yu2005measuring}, an $R^2$ of 0.68 and RMSE of 0.43 m over 5 years \citep{yu2006change}, and an $R^2$ of 0.9 and RMSE of 0.98 m over 20 years \citep{Soininen2022}. For DBH and volume change obtained from ALS data, \citet{Soininen2022} reported an $R^2$ of 0.3--0.36 for DBH change and 0.28--0.32 for volume change when 20-year growth was estimated by differencing independently estimated attributes. \citet{Tavi2026} investigated 10-year growth by combining ALS data collected at two time points with single-date MLS data from the more recent data collection year. By using ALS-derived height change and MLS-derived stem measurements to model past stem attributes and calculate change, DBH change was estimated with an RMSE of 0.9--1.7 cm and an $R^2$ of 0.29--0.44, and volume change with an RMSE of 0.04--0.10 $\mathrm{m}^3$ and an $R^2$ of 0.60--0.84. Overall, accurate growth estimation remains challenging because errors in single-time attribute estimates propagate into growth estimates, while the slow growth of boreal forest trees, particularly in DBH, results in a low growth-to-noise ratio \citep{yrttimaa2023capturing, Tompalski2019}. Furthermore, differences in sensors and scanning practices introduce inconsistencies between multitemporal datasets, complicating long-term analyses \citep{duncanson2018monitoring}.

To address some of these limitations, a practical framework for multitemporal forest monitoring would therefore be to combine under-canopy MLS/TLS data with ALS data to leverage the complementary strengths of both acquisition methods. This approach also addresses the practical limitation of scarce historical stem-level data, requiring only a single under-canopy acquisition together with multitemporal ALS data. In such a framework, the forest state could be measured as accurately as possible in the beginning of the monitoring period by using MLS/TLS data for stem measurements and incorporating ALS data for heights. Past or future stem attributes could then be reconstructed from ALS-derived height change and total growth could be inferred. Under-canopy MLS/TLS provide dense and minimally occluded observations of tree stems, whereas ALS enables accurate observations of treetops and efficient large-area data collection for observing height growth over time \citep{Wang2019}. Previous work by \citet{Tavi2026} and \citet{Soininen2024} demonstrated the overall feasibility of using this type of framework for reconstructing past stem attributes and estimating growth, with \citet{Tavi2026} using a combination of multitemporal ALS and single-date MLS data and \citet{Soininen2024} using only ALS data. However, the broader reliability and applicability of this workflow remains insufficiently understood, particularly in regards to the temporal error propagation of the scaling model and reliability of the use of multitemporal ALS-derived height estimates from different sensors and acquisitions. Furthermore, the growth estimation accuracy of this approach has not been studied across different monitoring periods or compared to the use of two-date attribute estimates directly obtained from combined under-canopy and ALS data to determine growth.

In this study, we investigate a framework for estimating individual tree DBH and stem volume growth using multitemporal laser scanning data acquired between 2014--2025 using point clouds from 11 different sensors across MLS, TLS, and ALS platforms. A total of 136 point clouds collected from eight boreal forest test sites were analyzed using multiple state-of-the-art methods, including deep learning-based tree segmentation, segmentation transfer between point clouds, and direct stem curve extraction from point cloud data. Stem curves and DBHs were extracted from all MLS/TLS datasets, with ALS data used for height and volume estimation, enabling the construction of attribute time series. We combined single-date stem measurements with multitemporal height estimates to reconstruct stem attributes across time using a height growth-based scaling model, in which ALS-derived height change is used to scale MLS/TLS-derived stem curves forwards and backwards in time. Growth in DBH and volume could then be estimated using the proposed model-based approach. Specifically, the study addresses four research questions: 1) How effective is the application of the deep learning-based tree segmentation and segmentation transfer of the detected trees to other point clouds for maintaining tree correspondence across time? 2) How do the errors in one-time attribute estimation propagate temporally when tree attributes are modeled forwards and backwards in time using a height growth-based scaling model? 3) Does incorporating stem taper variation in the model improve the scaling model? 4) How do manually-based growth estimates, direct differencing of independently estimated attributes from point clouds, and the proposed model-based approach compare in estimating 5- and 10-year growth? These analyses enable us to assess whether retrospective individual tree growth can be reliably estimated by combining ALS-derived height growth with current MLS-derived stem curves to reconstruct past attributes, without requiring historical under-canopy MLS/TLS data.

\section{Materials}
This section describes the study area, scanners and scanning practices used during the scanning campaigns, as well as the resulting point cloud datasets.

\subsection{Test sites}
\begin{figure}[ht!]
    \centering
    \begin{subfigure}{0.49\linewidth}
        \caption{\label{fig: evo_area1}}
        \includegraphics[width=\linewidth]{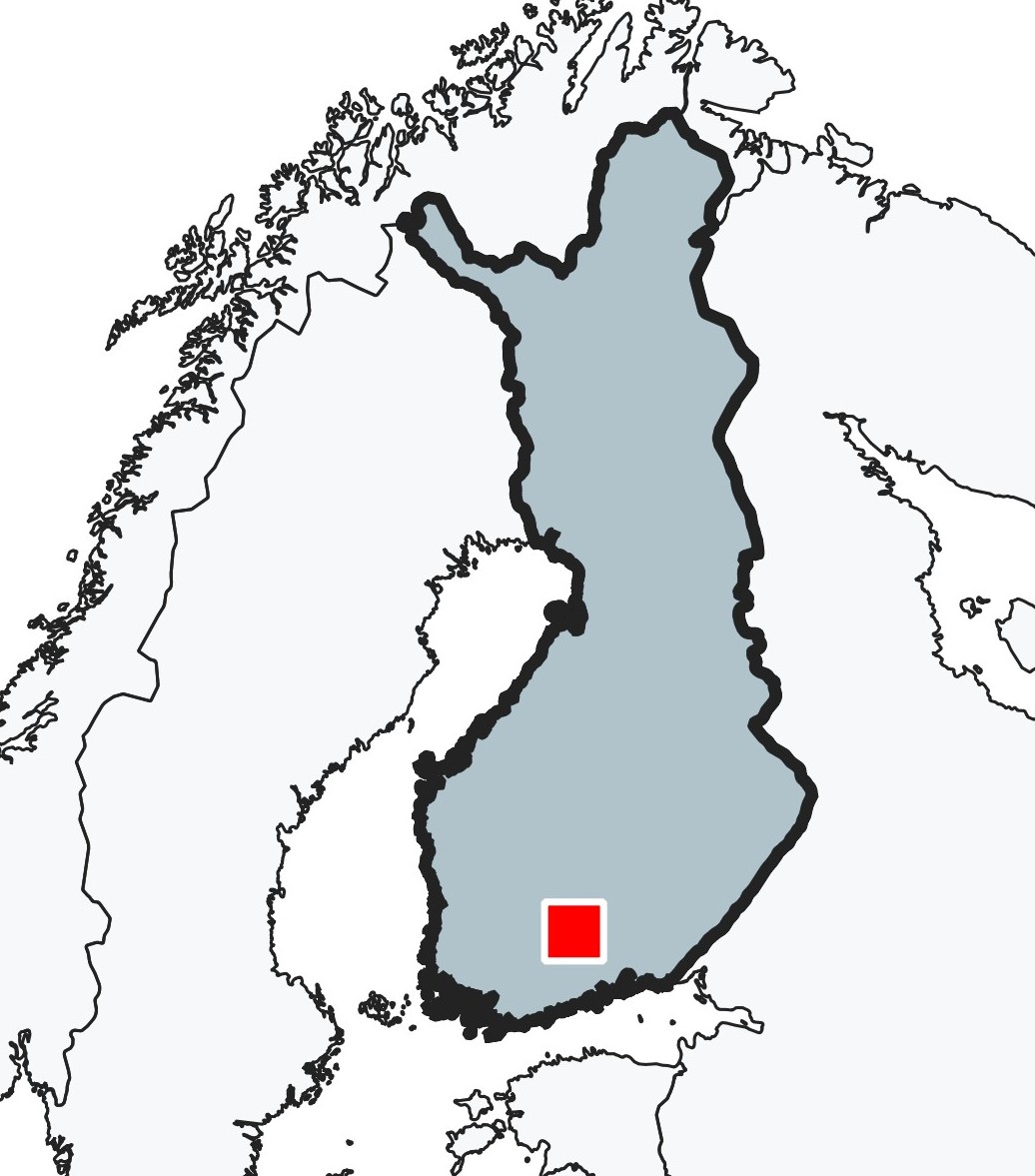}
    \end{subfigure} \\
    \begin{subfigure}{0.49\linewidth}
        \caption{\label{fig: evo_area2}}
        \includegraphics[width=\linewidth]{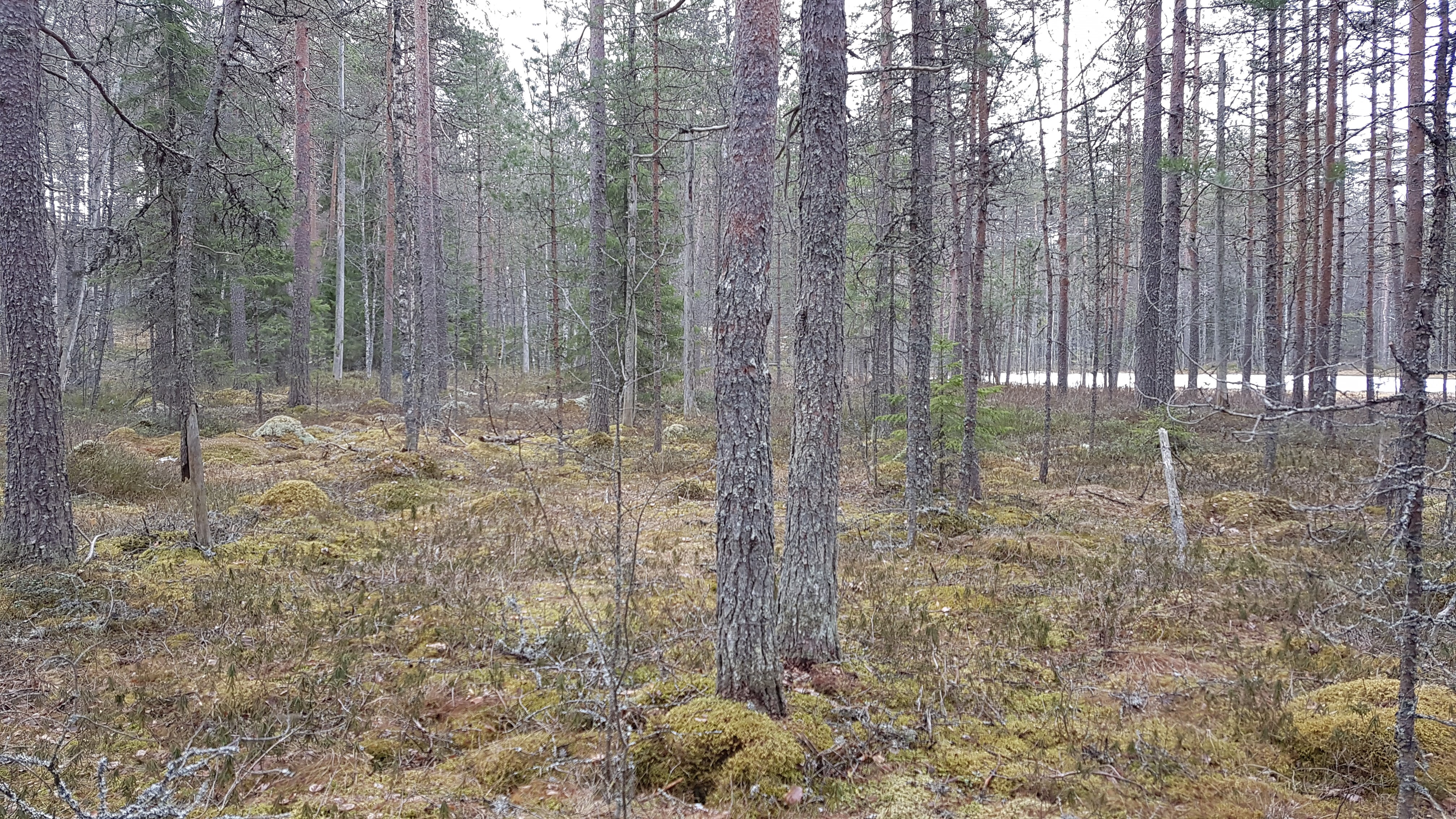}
    \end{subfigure} 
    \begin{subfigure}{0.49\linewidth}
        \caption{\label{fig: evo_area3}}
        \includegraphics[width=\linewidth]{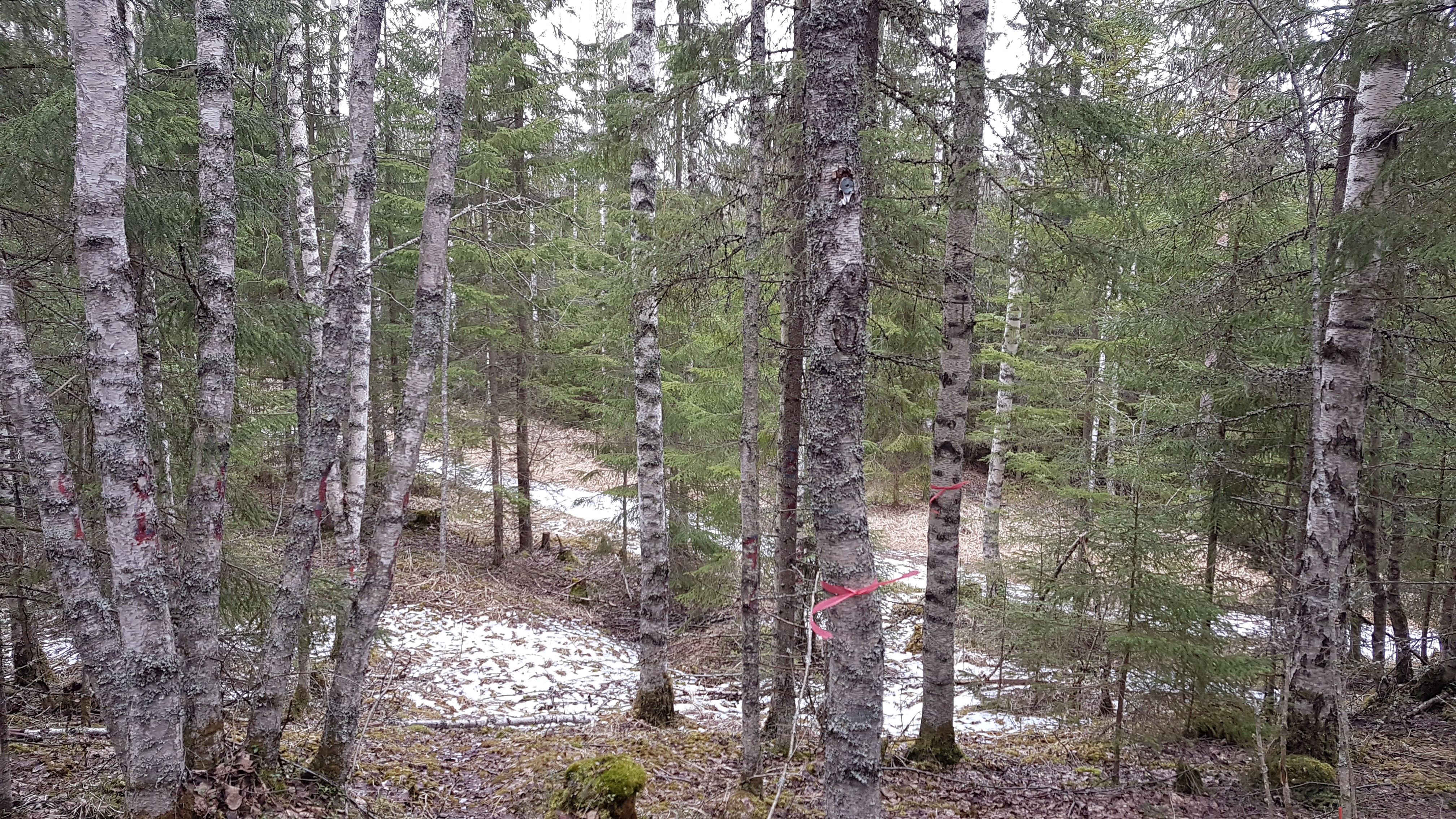}
    \end{subfigure}
\caption{\ref{fig: evo_area1}: The location of the Evo study area in Finland (red square). \ref{fig: evo_area2}: A photo from an easy plot used in this study. \ref{fig: evo_area3}: A photo from a difficult plot used in this study.}
\end{figure}

The test sites used in this study were located in Evo, southern Finland (61.19\degree N, 25.11\degree E, Fig. \ref{fig: evo_area1}), within a lightly managed boreal forest that forms part of the SCAN FOREST\footnote{\url{https://www.scanforest.fi}} research infrastructure. The forest is primarily composed of Scots pine (\textit{Pinus sylvestris}), Norway spruce (\textit{Picea abies}), silver birch (\textit{Betula pendula}), and downy birch (\textit{Betula pubescens}). Due to the similarities between the two birch species, they are combined and referred to as \textit{birch} in this paper.

Eight plots of size 32 m $\times$ 32 m were used, with four plots categorized as \textit{easy} and four as \textit{difficult}. This categorization follows previous studies conducted in the area \citep{Tavi2026, muhojoki2024benchmarking}, and is primarily based on species composition, tree density, and degree of understory vegetation present. The easy test sites were dominated by pines and characterized by sparse under-canopy structure, lower tree density, and good stem visibility. In contrast, the difficult plots contained a more diverse mixture of tree species present, with few pines, and denser under-canopy forest structure, higher stem density, more obstructing branches, and greater levels of occlusion. The average tree density was 575 trees per hectare in the easy category and 1460 trees per hectare in the difficult category, based on the total number of trees found in the test sites in 2024 according to the manual data used in this study. These two categories, therefore, represent contrasting forest conditions, enabling the proposed framework to be evaluated under both favorable and challenging environments. The plot categorization in this study is also consistent with previous findings showing that individual tree detection and attribute estimation is most accurate for pines, compared to spruces and birches \citep{Tavi2026, yrttimaa2022exploring,hyyppa2020comparison, hyyppa2020under}. This is largely attributable to species-specific differences in branching patterns, crown structure, and tree stem visibility.

The location of the Evo study area in Finland is shown in Fig. \ref{fig: evo_area1}. Figs. \ref{fig: evo_area2} and \ref{fig: evo_area3} show examples of easy and difficult plots. Table \ref{tab:plot_stats} summarizes the main plot characteristics according to the subset of trees used in this study, including species composition and mean tree attribute values. This subset of trees is based on the common set of detected trees and their stems across the different point cloud datasets.

\begin{table}[!htb]
\centering
\caption{Plot characteristics based on the subset of trees used in this study, including the species composition and mean attribute values. The DBH values were obtained from manual measurements in 2024. The mean height in 2025 was estimated using combined ALS and MLS data.}\label{tab:plot_stats}
\begin{tabular*}{\linewidth}{@{\extracolsep\fill}lll}
\toprule
 & \textbf{Easy plots} & \textbf{Difficult plots}\\
\midrule
Number of plots & 4 & 4 \\
\midrule
& \multicolumn{2}{@{}c@{}}{Mean tree count per plot}\\
\midrule
Pine & 44 & 2 \\
Spruce & 0 & 3 \\
Birch & 2 & 14 \\
Other & 0 & 8 \\
DBH below 10 cm & 1 & 2 \\
DBH below 20 cm & 8 & 12 \\
Total & 46 & 27 \\
\midrule
& \multicolumn{2}{@{}c@{}}{Mean attribute values}\\
\midrule
DBH 2024 (cm) & 25.0 & 25.2 \\
Height 2025 (m) & 20.6 & 22.8 \\
\bottomrule
\end{tabular*}
\end{table}

\subsection{ALS devices}
This study used ALS point clouds acquired in 10 different years between 2014 and 2025 using seven scanners from Teledyne (Teledyne Technologies, California, USA) and Riegl (Riegl GmbH, Horn, Austria), mounted on different airborne platforms. This section summarizes the key characteristics of the scanners themselves, while the corresponding datasets produced by the scanners are described in Section \ref{sec:data_description}. The nominal scanner specifications are presented in Table \ref{tab: als_sensors}.

The Teledyne Optech Titan is a three-wavelength multispectral scanner consisting of visible (532 nm), infrared (1550 nm), and near-infrared (1064 nm) channels. In this study, all channels were used in combination. Riegl drone scanners, VQ-480-U, VUX-1HA, and miniVUX-1UAV, were integrated into the in-house-developed HeliALS multispectral scanner. Although the device has evolved over time, the scanners in this study have been a part of the device at some point during its development. The current HeliALS configuration combines four individual scanners operating at three wavelengths combined with a hyperspectral camera, producing multispectral point clouds that can be used, for example, for tree species classification \citep{Taher2026}.

UAV-based ALS acquisitions were conducted using the Riegl VUX-1UAV and VUX-120 scanners. The VUX-1UAV datasets were acquired at a flight altitude of approximately 50 m, whereas the VUX-120 datasets were collected at approximately 250 m.

The study also incorporates data from Finland's second national laser scanning campaign, acquired using the Riegl VQ-1560i scanner. In this campaign, most of Finland's land area was scanned using sensors with a pulse density of five pulses/m\textsuperscript{2} \citep{nls_data}. The Evo study area was scanned in 2019, while the campaign as a whole was flown during 2019--2025. A third iteration of the national laser scanning campaign began in 2026 and uses scanners with a pulse density of 20 pulses/m\textsuperscript{2}.

\begin{table*}[htb!]
    \caption{\label{tab: als_sensors}Nominal specifications of the used ALS sensors. The accuracy and precision values are calculated at $1\sigma$. NIR = near infrared, FOV = field of view.}
    \centering
    \begin{threeparttable}
    \begin{tabular*}{\textwidth}{@{\extracolsep\fill}lllll@{}}
        \toprule
        \textbf{Device/property}& \textbf{Accuracy/precision} (mm) & \textbf{Beam divergence} (mm) & \textbf{Wavelength} (nm) & \textbf{Max. FOV} (\degree) \\
        \midrule
        Riegl VQ-480-U&   25/25\tnote{a} & 0.3\tnote{1}& 1550& 60\\
        Teledyne Optech Titan& \makecell[l]{alt/7500 (horz.),\\50--100 mm (elev.)/\\8 mm}& 0.35/0.35/0.7\tnote{2}& 1550/1064/532& 60\\
        Rieql VUX-1UAV& 10/5\tnote{a}& 0.5\tnote{1}& 1550& 330\\
        Riegl miniVUX-1UAV & 15/10\tnote{b} & 2.7 $\times$ 0.85\tnote{1}& NIR & 360\\
        Riegl VQ-1560i& 20/20\tnote{c}& 0.25\tnote{1}& NIR& 58\\
        Riegl VUX-1HA& 5/3\tnote{d}& 0.5\tnote{1}& 1550  & 360\\
        Riegl VUX-120& 10/5\tnote{a}& 0.4\tnote{1}& NIR& 100\\
        \bottomrule
    \end{tabular*}
    \begin{tablenotes}
    \setlength{\columnsep}{0.8cm}
    \setlength{\multicolsep}{0cm}
    \begin{multicols}{3} 
        \item[a] @ 150 m
        \item[b] @ 50 m
        \item[c] @ 250 m
        \item[d] @ 30 m
        \item[1] 1/$\mathrm{e^2}$
        \item[2] 1/e
    \end{multicols}
    \end{tablenotes}
    \end{threeparttable}
\end{table*}

\subsection{MLS and TLS scanners}
MLS and TLS point clouds were collected in seven different years between 2014 and 2025. TLS data were acquired in 2014, 2019, 2021, and 2023, while MLS data were collected in 2020, 2024, and 2025. The TLS scanners used in this study were from Leica (Leica Geosystems AG, Heerbrugg, Switzerland), while the MLS systems were produced by FARO (FARO, California, USA) and GeoSLAM (currently part of FARO). The nominal sensor specifications are summarized in Table \ref{tab: tlsmls_sensors}.

\begin{figure*}[htb!]
    \centering
    \begin{subfigure}{0.33\textwidth}
        \caption{\label{fig:trajectories1} TLS}
        \includegraphics[width=\linewidth]{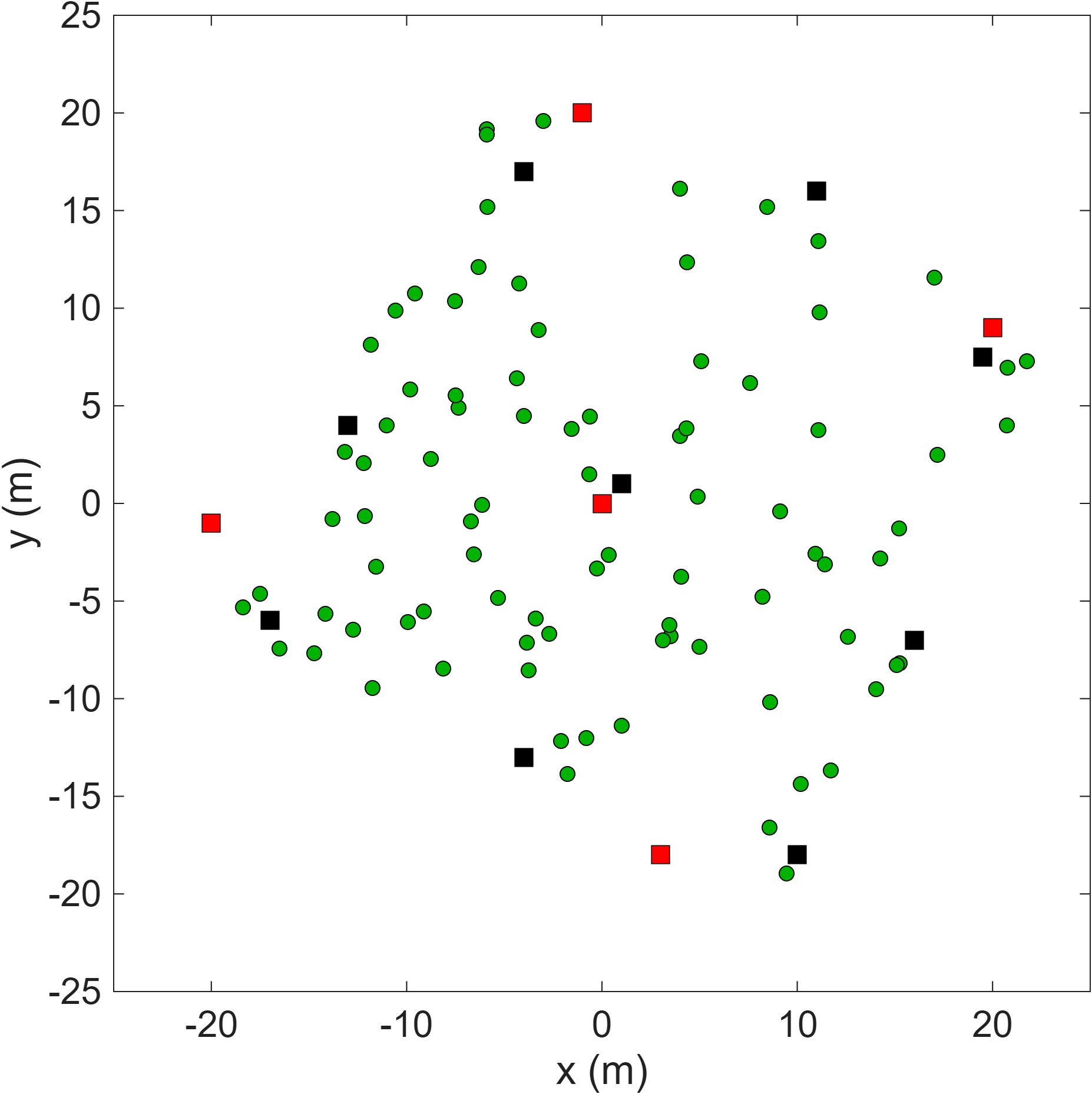}
    \end{subfigure}
    \begin{subfigure}{0.33\textwidth}
        \caption{\label{fig:trajectories2} MLS}
        \includegraphics[width=\linewidth]{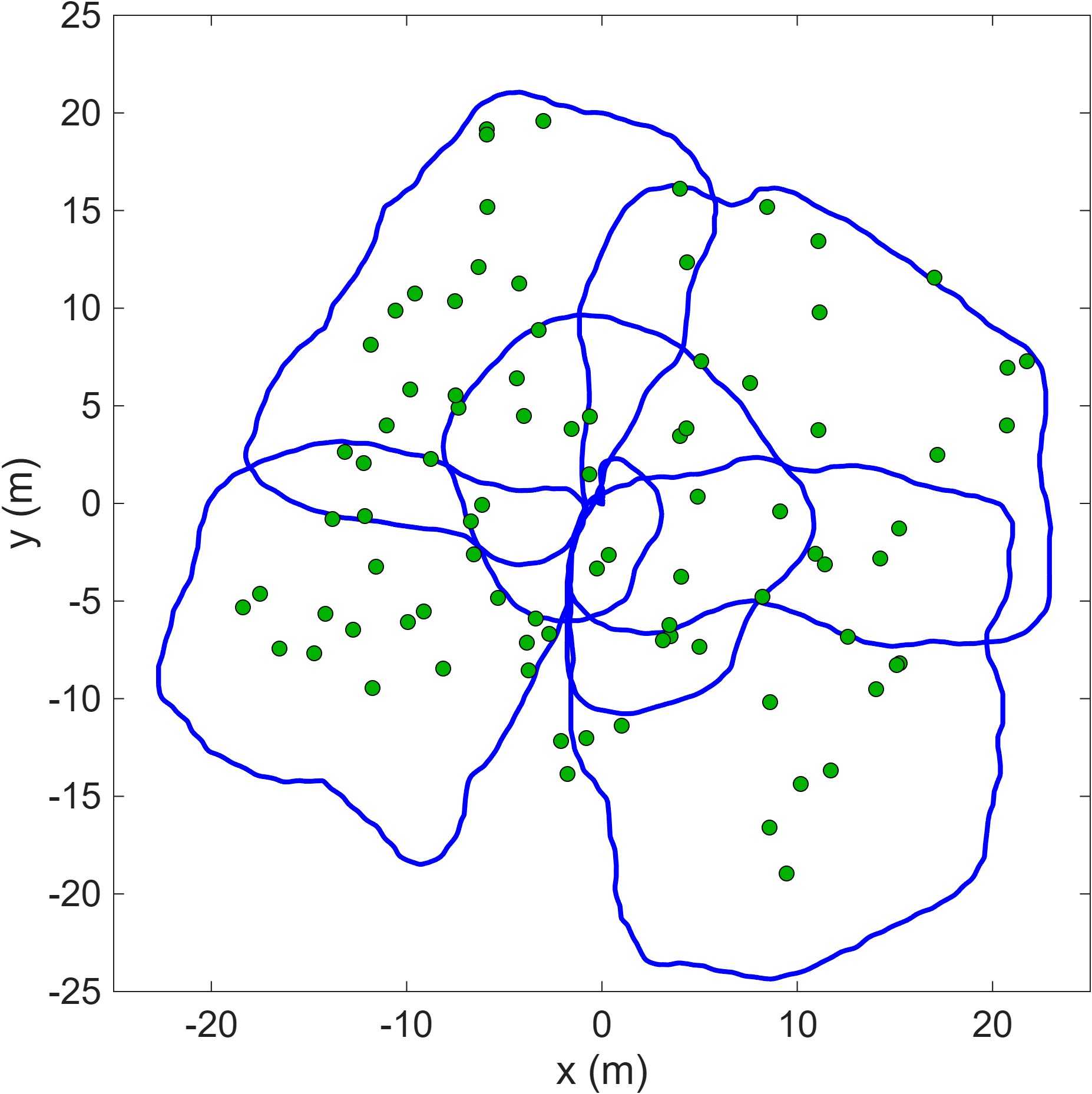}
    \end{subfigure}
    \begin{subfigure}{0.33\textwidth}
        \caption{\label{fig:trajectories3} ALS}
        \includegraphics[width=\linewidth]{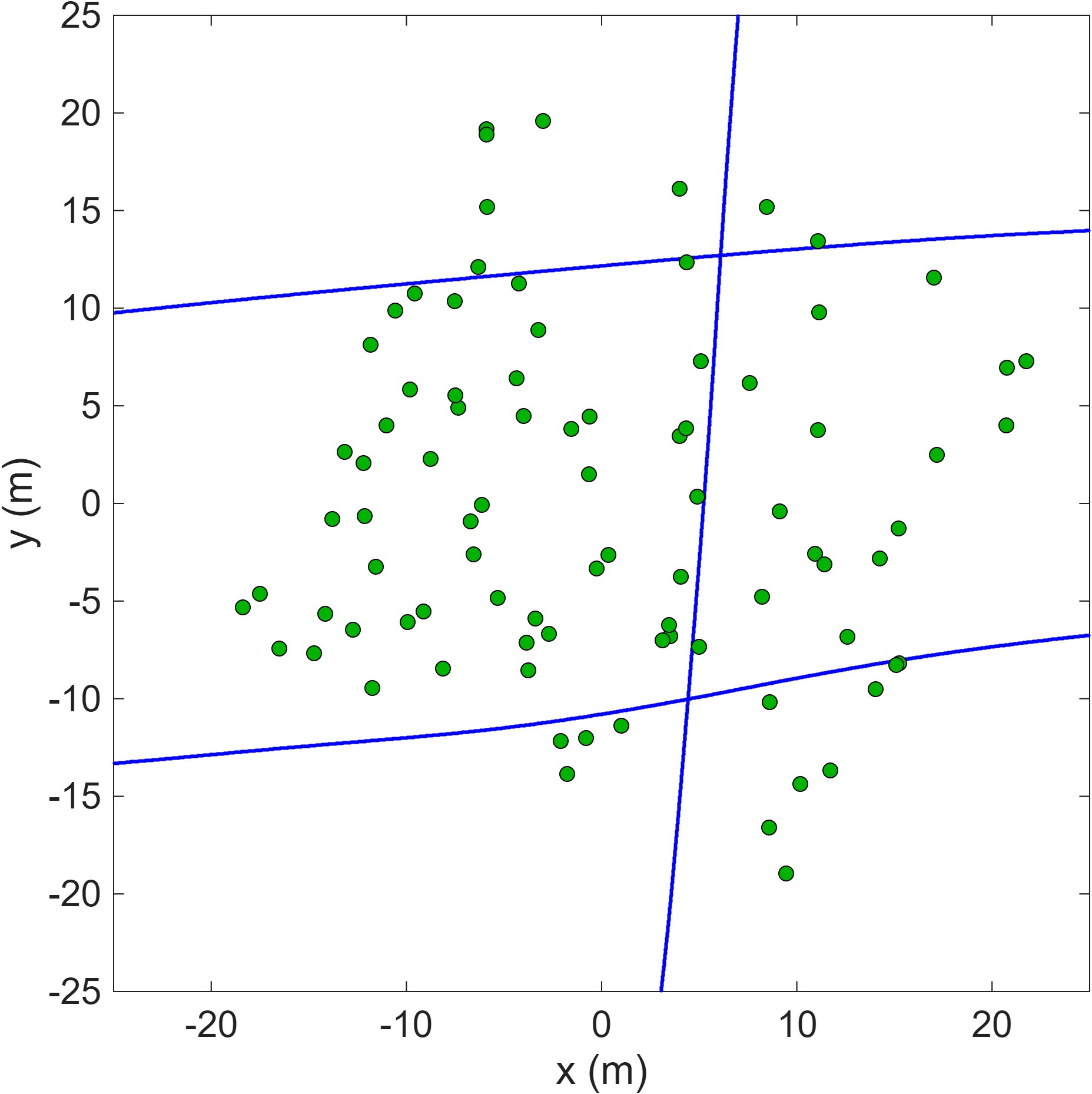}
    \end{subfigure}
    \caption{Example scanner positions and scanning trajectories for TLS, MLS, and ALS data from one of the easy plots in this study. Green marks indicate manually measured tree locations. \ref{fig:trajectories1}: TLS scanner positions from 2014 (red squares) and 2023 (black squares). The positions were hand-drawn based on real scanner positions. \ref{fig:trajectories2}: MLS scanning trajectory from the 2020 data collection showing the loops traversed during data acquisition. \ref{fig:trajectories3}: ALS scanning trajectory from the 2021 campaign. Both trajectories are real trajectories, but manually transformed to the same coordinate system.}
    \label{fig:trajectories}
\end{figure*}

For TLS acquisitions, either five or nine scans were combined into a single multi-scan point cloud using reference spheres with accurately measured locations within each forest plot. Scan positions were arranged such that one scan was taken from the plot center, with the remaining scans surrounding it to minimize occlusion effects. MLS data were collected using backpack or handheld carriers, with scanning trajectories incorporating loop closures for post-processing with (proprietary) simultaneous localization and mapping (SLAM) algorithms, and for providing a 360-degree view of the tree stems. Fig. \ref{fig:trajectories} illustrates example scanner configurations and acquisition trajectories for TLS, MLS, and ALS point clouds.

\begin{table*}[htbp!]
    \caption{\label{tab: tlsmls_sensors}Nominal specifications of the used TLS and MLS sensors.}
    \centering
    \begin{threeparttable}
    \begin{tabular*}{\textwidth}{@{\extracolsep\fill}lllll@{}}
        \toprule
        \textbf{Device/Property}& \textbf{Accuracy/precision} (mm) & \textbf{Beam divergence} (mrad) & \textbf{Wavelength} (nm) &  \textbf{Max. FOV} (\degree) \\
        \midrule
        Leica HDS6100& $\pm5/1$ \tnote{a} & 0.22& 650--690 & $360\times 310$ \\
        Leica RTC360& $\pm1.9/0.4$\tnote{b}& 0.5 & 1550& $360 \times 310$ \\
        GeoSLAM ZEB Horizon & $\pm30$/NA & $1.5 \times 3.0$ &  905 & $360 \times 270$  \\
        FARO Orbis & $\pm10$/5& $0.7 \times 1.7$ &  905 & $360 \times 290$ \\
        \bottomrule
    \end{tabular*}
    \begin{tablenotes}
    \item[a] @ < 25m
    \item[b] @ 10 m
    \end{tablenotes}
    \end{threeparttable}
\end{table*}

\subsection{Manual measurements}

Manual DBH measurements were used in this study as reference data for evaluating the accuracy of one-time DBH estimates, modeled DBH values, as well as to obtain manual growth estimates. These measurements were collected in the years 2014, 2019, 2021, and 2024 with only trees with a DBH greater than 5 cm recorded. Species information and the location of the tree were also noted. The 2014, 2019, and 2021 DBH measurements were collected using steel calipers, where the average of two measurements from perpendicular directions around the tree stem was recorded. Meanwhile, the DBH measurements in 2024 were collected with a tape measure. Using two different approaches to DBH measurement collection may introduce minor systematic differences. However, prior research has shown these differences to be small and around a few millimeters \citep{liu2011comparing, wilson2007comparison}. Heights of the trees were also measured manually in the years 2014, 2019, and 2021 using clinometers.

Tree stem volume was calculated using the standard Finnish allometric model based on DBH, height, and tree species information \citep[][Eqs. (61.6)]{Laasasenaho1982}. The Laasasenaho model provides species-specific equations for pine, spruce, and birch. Trees that were not one of these three species, in our study categorized as \textit{other}, used the equation for pine. The equations were applied in two different ways for separate analyses in this study. First, we evaluated the equations using the manually measured DBH and height from 2014, 2019, and 2021. This provided fully manual stem volume estimates for a baseline comparison between point cloud-derived estimates and manual-only estimates. Secondly, we re-evaluated the equations using height values obtained from combined ALS and MLS/TLS point clouds, while retaining the manually measured DBH values. Details of the height estimation approach are provided in Section \ref{sec:attribute_methods}. Combining ALS and MLS/TLS point clouds enabled more accurate height estimation than the use of any individual dataset alone. These volume estimates were used to compare the traditional manual method using the more accurate height data and the fully point cloud-derived volumes.

Only trees that had manual measurements available for all measurement years were included in the analysis of this study. Trees that had fallen down, died, or were otherwise missing from one of the years were excluded. Similarly, trees that were only recorded in later years after exceeding the 5 cm DBH threshold were also excluded from the overall manual results of this study.

\subsection{Data description}
\label{sec:data_description}
This section summarizes the key characteristics of the point cloud datasets used in this study, as reported in Table \ref{tab:data_characteristics}.

The ALS datasets exhibit substantial variation in point cloud density. Point density is calculated from all recorded laser echo returns, which differs from the number of emitted pulses, as the scanners can record multiple returns per emitted pulse. Multiple returns from within the laser footprint allow inference on the vertical distribution along the laser path and improve the ability of the scanner for under-canopy observations \citep{Wehr1999}. The HeliALS device (with VQ-480-U, miniVUX-1UAV, and VUX-1HA) and the UAV-mounted VUX-1UAV were flown at low altitudes, resulting in small beam footprints at tree canopy level. Consequently, laser pulses are predominantly reflected from the upper canopy layers, which limits penetration into the canopy and reduces the ability to detect under-canopy trees. This limitation is partially offset by the higher number of return echoes per emitted pulse of the VQ-480-U, VUX-1HA, and VUX-1UAV data, as well as their high pulse repetition rates, which increase overall point density. In contrast, the miniVUX-1UAV lacks both of these characteristics, resulting in comparatively lower suitability for under-canopy tree observation.

The ALS point clouds acquired in the national laser scanning campaign using the VQ-1560i scanner were collected at higher flight altitudes. In this case, the laser beam can penetrate deeper into the canopy, also supported by numerous recorded return pulses, and does not only reflect from upper canopy layers. As a result, the 2019 ALS point clouds are more suitable for under-canopy observations than what would be expected based on point density alone. The remaining ALS point clouds collected with the Optech Titan and VUX-120 combine high flight altitude with multiple returns and high point density, resulting in improved representation of tree height regardless of their location in the forest.

We subsampled the 2023 TLS point clouds to one-third of their original size to reduce processing time. Even after subsampling, the resulting point clouds retained approximately twice the point density of the densest MLS point clouds and remained comparable to the other TLS point clouds. The more recent TLS point clouds (2019, 2021, 2023) represent the densest point clouds. Meanwhile, the earlier TLS point clouds (2014 and 2019) suffer from fewer scanning positions, which increases occlusion effects. In contrast, MLS data capture tree stems from all viewing angles, which improves completeness of stem representation and is advantageous for stem curve modeling \citep{Donager2021}.

We did not correct for timing differences arising from data collection occurring in different months within the same year. Instead, all observations collected within a given year were assumed to represent the forest state for that year. Consequently, ALS, MLS/TLS, and manual measurements may reflect slightly different stages of annual tree growth. For Scots pine and Norway spruce, which constitute the majority of trees in this study, height growth is typically completed between late June and early July in southern Finland \citep{korpelainen2026uncovering, makinen2018dynamics, zhai2012variation}. Radial stem growth generally continues until mid- or late August, and may extend until early September under favorable conditions \citep{makinen2018dynamics, zhai2012variation}. As a result, measurements acquired in different months capture slightly different proportions of the annual height and diameter increments. This may therefore introduce additional variability into the produced attribute time series, as well as into one-time attribute accuracy assessments and comparisons of growth estimates.

\begin{table*}[!htb]
    \centering
    \caption{\label{tab:data_characteristics} Key characteristics of each dataset used in this study. The month of measurement is coarsed to a season of measurement if that is the resolution with which the measurement period has been recorded.}
    \begin{tabular*}{\textwidth}{@{\extracolsep\fill}llllll@{}}
         \toprule
          \textbf{Year}&\textbf{ALS}&\textbf{Period}&\textbf{Approx. altitude} (m)&\textbf{N. of returns}&  \makecell[l]{\textbf{Density}\\(points/m\textsuperscript{2})} \\
         \midrule
          2014&VQ-480-U& Dec& 75 &5& 360\\
          2016&Optech Titan&May--June& 500 &4&210 \\
          2017&VUX-1UAV& Sept&50&5&6350 \\
          2018&miniVUX-1UAV& May& 80&3&120\\
          2019&VQ-1560i& July& 1800&5&15 \\
          2021&VUX-1HA& June& 80&5&3200 \\
          2022&VUX-1HA& Sept & 80&4--5&1080\\
          2023&VUX-1HA& July& 80&5&1940 \\
          2024&VUX-120& May& 250&4--5&670 \\
          2025&VUX-1HA& June& 100&4--5&490\\
         \midrule
          \textbf{Year} & \textbf{TLS/MLS*} & \textbf{Period} & \makecell[l]{\textbf{Scanner positions/}\\\textbf{carrier}} & \makecell[l]{\textbf{Density}\\(points/m\textsuperscript{2})} & \makecell[l]{\textbf{Manual}\\\textbf{measurements}} \\
         \midrule
          2014& HDS6100&  Apr-May&center + 4 corners& $6.0\times10^4$& May--Aug\\
          2019& RTC360&  Autumn&center + 4 corners& $2.2 \times 10^5$&Autumn\\
          2020& ZEB Horizon*& Apr& handheld& $5.5 \times 10^4$&\\
          2021& RTC360&  Spring&center + 8 surrounding& $2.2 \times 10^5$&Sept--Oct\\
          2023& RTC360&  May&center + 8 surrounding& $2.2\times10^5$ &\\
          2024& FARO Orbis*&  Sept--Oct&backpack carrier& $1.4 \times 10^5$&Aug\\
          2025& FARO Orbis*&  May&backpack carrier& $1.2 \times 10^5$&\\
         \bottomrule
    \end{tabular*}
\end{table*}

\section{Methods}
\begin{figure*}[!htb]
\centering
\begin{tabular}{c|c}
    \begin{subfigure}{0.40\linewidth}
        \caption{\label{fig:workflow1}}
        \includegraphics[width=\linewidth]{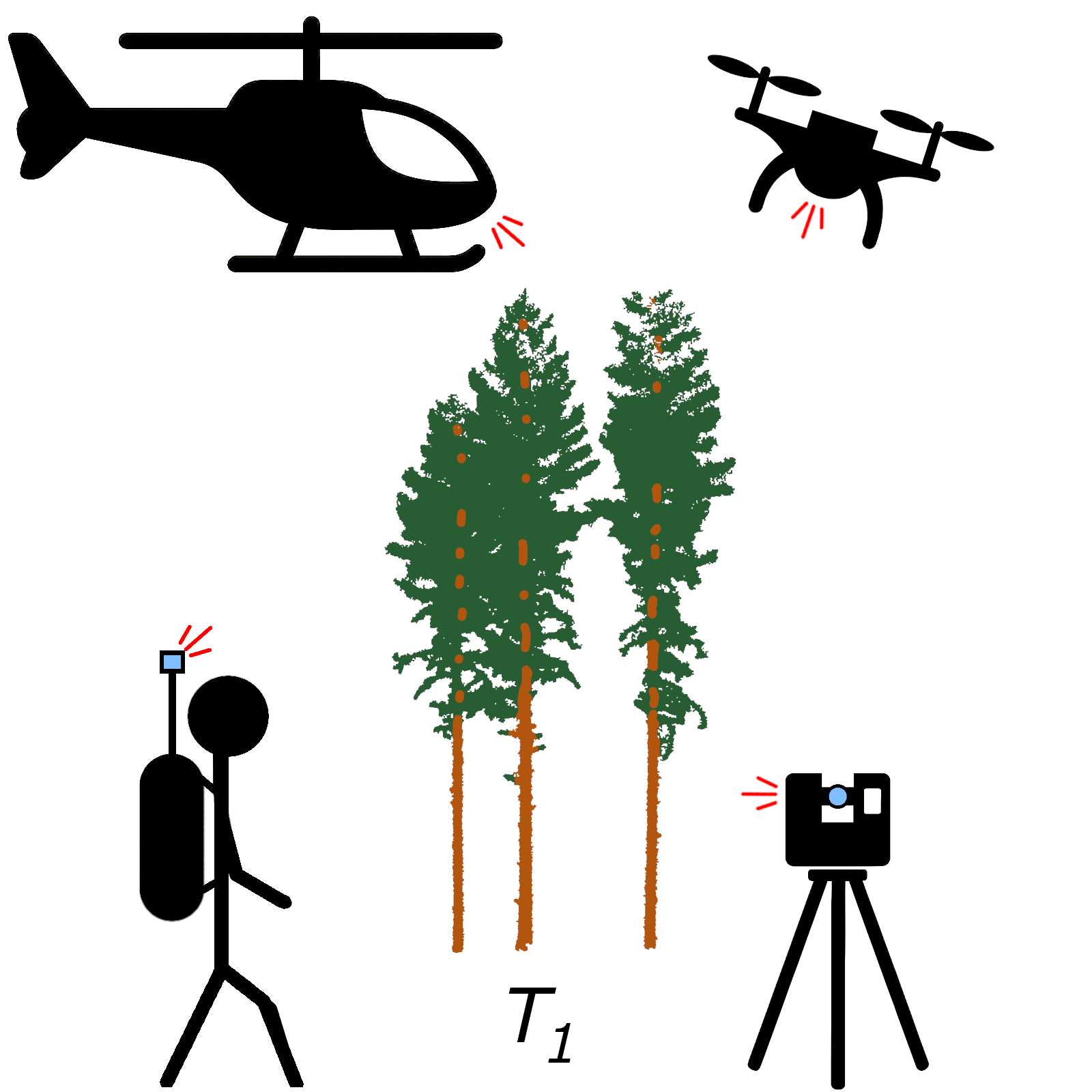}
    \end{subfigure} &
    \begin{subfigure}{0.40\linewidth}
        \caption{\label{fig:workflow2}}
        \includegraphics[width=\linewidth]{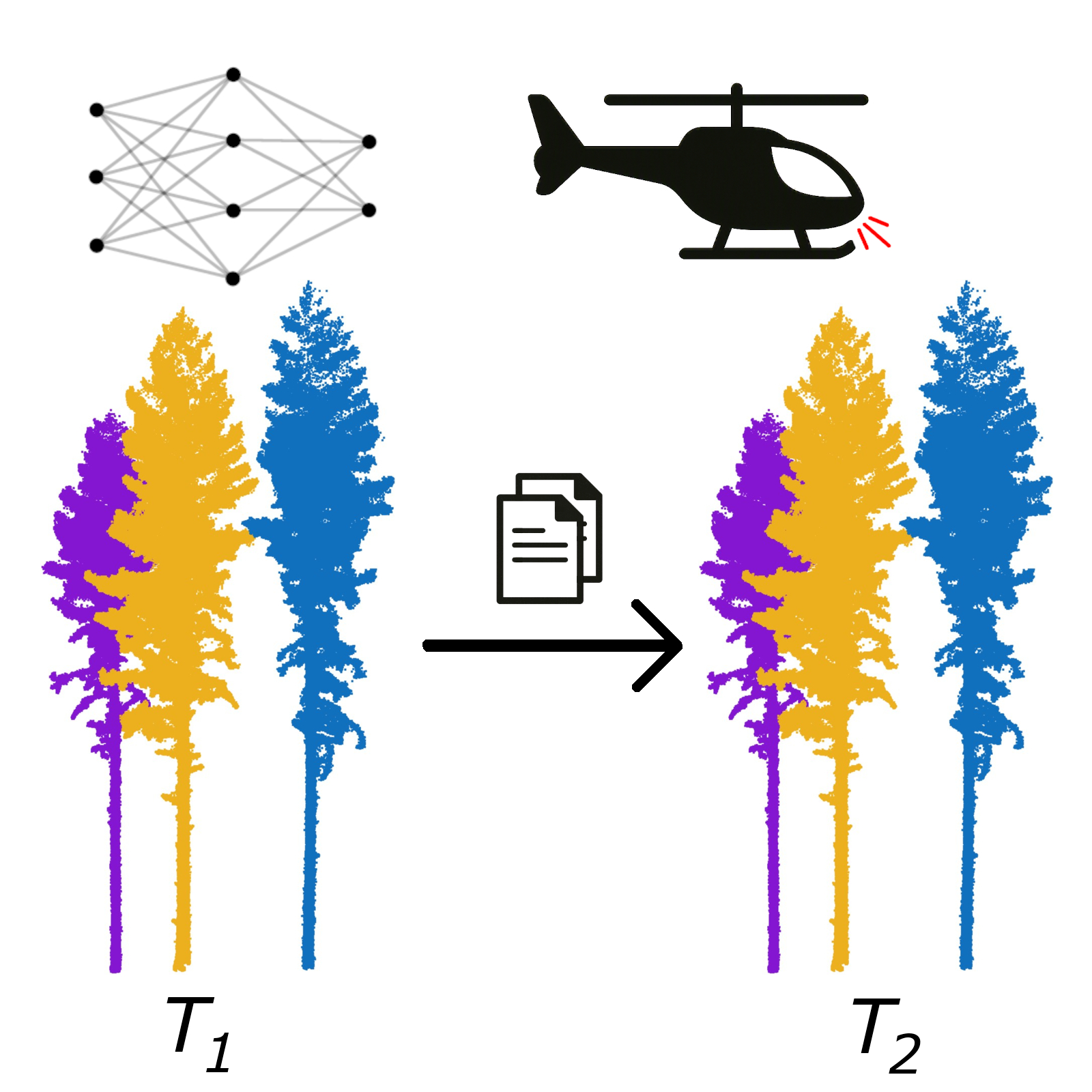}
    \end{subfigure} \\ \hline
    \begin{subfigure}{0.40\linewidth}
        \caption{\label{fig:workflow3}}
        \includegraphics[width=\linewidth]{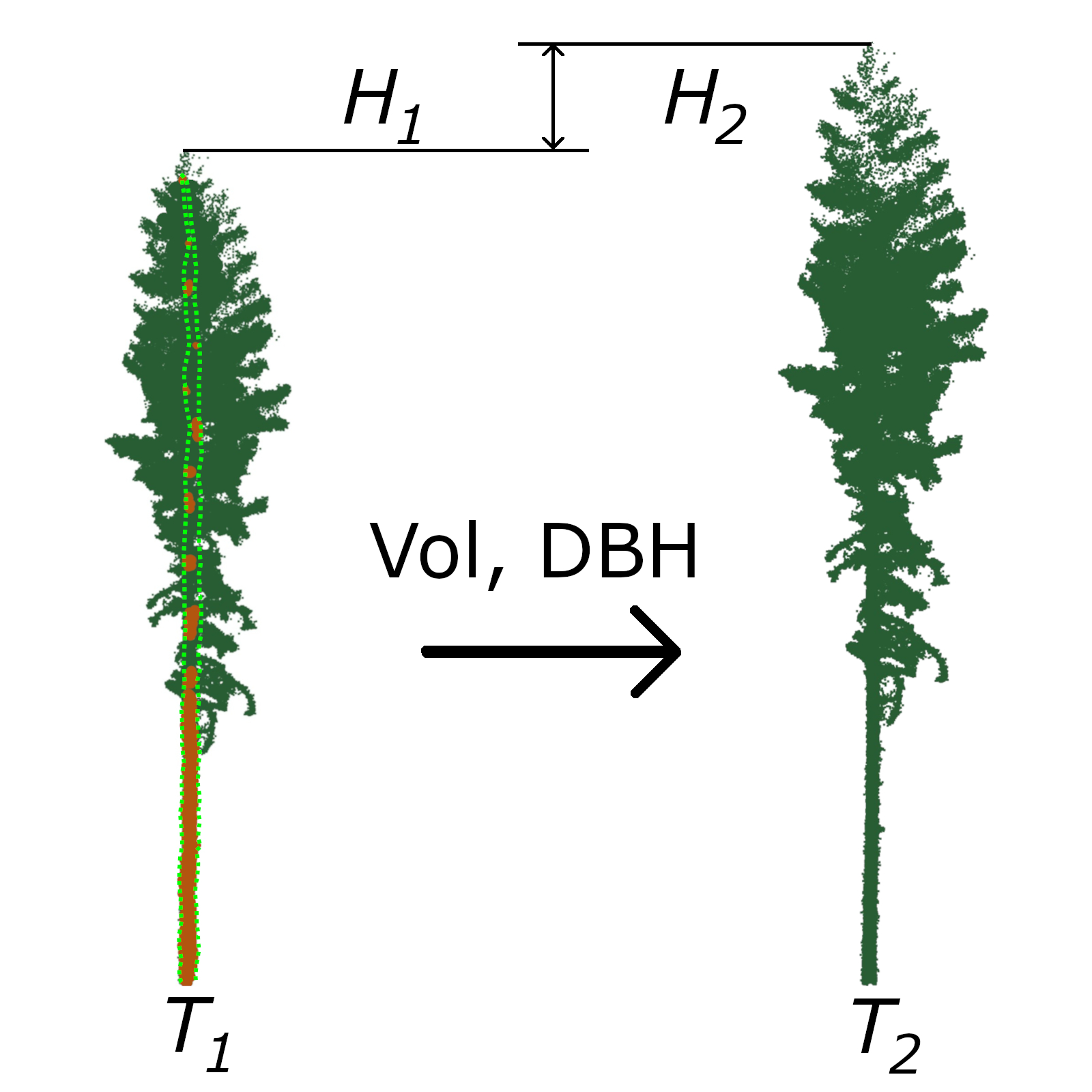}
    \end{subfigure} &
    \begin{subfigure}{0.40\linewidth}
        \caption{\label{fig:workflow4}}
        \includegraphics[width=\linewidth]{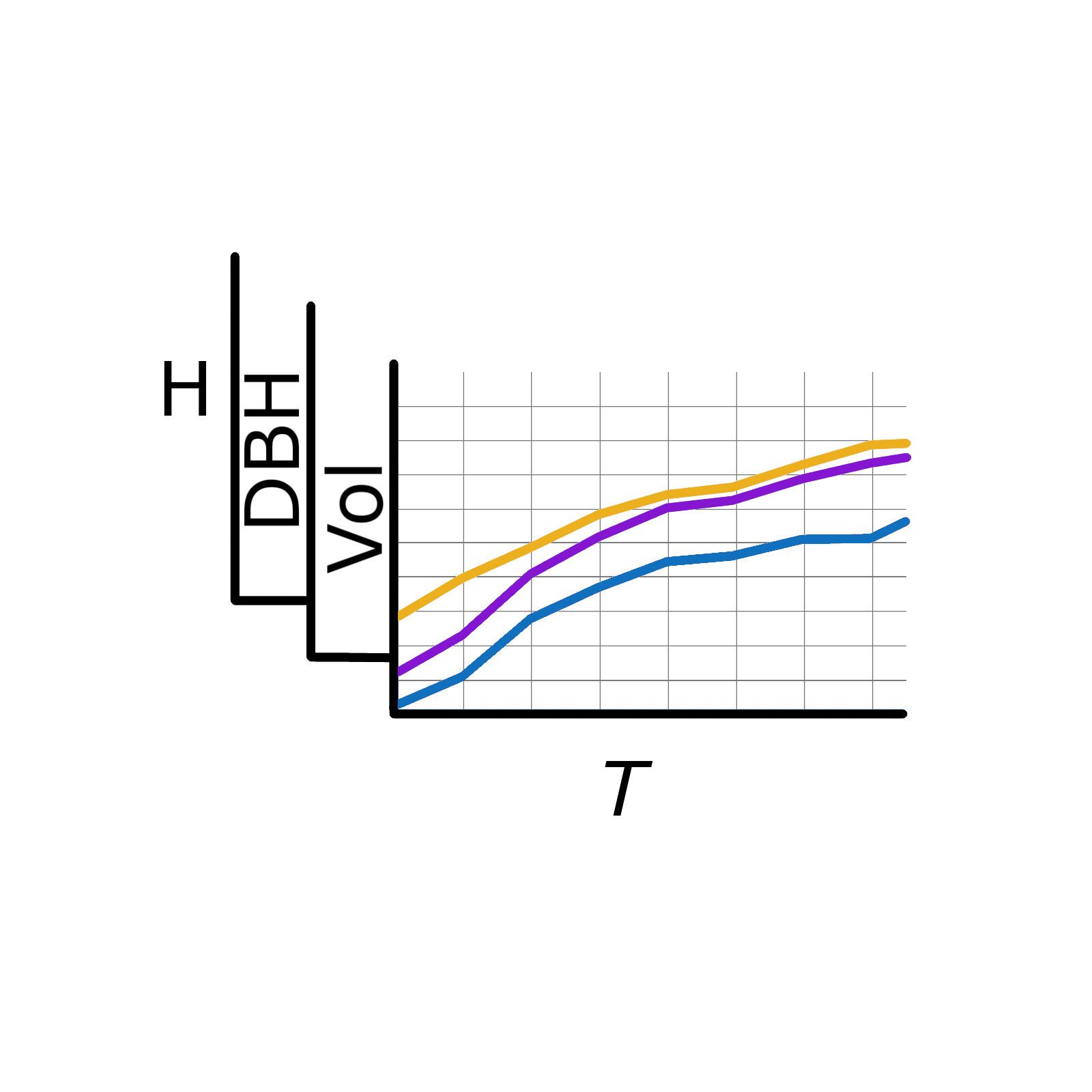}
    \end{subfigure}
\end{tabular}
\caption{\label{fig:workflow} The main steps of the proposed framework. \ref{fig:workflow1}: In the starting year ($T_1$), the forest is scanned both under the canopy using MLS (e.g. a backpack scanner) or TLS, and above the canopy using ALS (e.g. a helicopter- or UAV-based system). \ref{fig:workflow2}: Individual trees are segmented from the point cloud acquired at $T_1$ using a deep learning-based segmentation method, after which the segment labels are transferred to the ALS point cloud acquired at $T_2$. \ref{fig:workflow3}: Volume and DBH are estimated from tree stems at $T_1$. The corresponding attributes at $T_2$ are reconstructed using a height growth-based model that scales the stem proportions at $T_1$ without requiring stem observations at $T_2$. \ref{fig:workflow4}: Repeating steps b and c yields a time series of modeled DBH and stem volume together with estimated tree heights.}
\end{figure*}

This section describes the individual tree change detection framework used in this study. The overall workflow is illustrated in Fig. \ref{fig:workflow}. Point clouds acquired using ALS and MLS/TLS scanners were georeferenced and segmented into individual trees, after which DBH and stem volume were measured from tree stem points. Consistent tree correspondence across the multitemporal point clouds was established by transferring segment labels to all point clouds. DBH and stem volume growth were then estimated using a model where stem attributes were scaled according to height growth derived from ALS point clouds. Consequently, only a single under-canopy scanning campaign was required. In this study, most of the computational steps (georeferencing, segment transfer, and stem curve algorithm) were implemented on a laptop with a 20-core Intel Core i7-13850HX and 64 GBs of RAM using Matlab 2025a. The preliminary segmentation was performed on a workstation with a 12-core Intel Xeon w5-3425 CPU, $8 \times 64$ GBs of RAM, and using a NVIDIA RTX A6000 48 GB GPU.

\subsection{Planar georeferencing of the point clouds}
\label{sec:georeferencing}
The aim of coordinate georeferencing was to transform all point clouds into the same coordinate system as the manually measured tree locations. At this stage, only the $xy$ coordinates were transformed, while the $z$ coordinate was processed later using different methods. The manually measured trees existed in the global ETRS-TM35-FIN coordinate system. While many ALS point clouds were already in this global coordinate system, the MLS and TLS point clouds were in local coordinate systems and therefore required transformation to global coordinates. To determine the associated transformation, we utilized tree locations that had first been identified from the target MLS/TLS point clouds. These locations were derived either from detected tree stems using the stem curve algorithm presented in \citet{Hyypp2020} and described in Section \ref{sec:stemcurve}, or from treetops identified as the height maxima in a rasterized and smoothed image with a pixel size of $\mathrm{0.5 \,m\times 0.5\, m}$. We calculated the planar $xy$-translation and azimuthal rotation using the tree locations. For finding the transformation, we used a revised method based on the rotation- and translation-invariant coarse-to-fine registration method presented in \citet{Hyyppa2021}.

In general, the more accurate tree locations derived from the detected stems would result in better coordinate transformations than the locations derived from treetops. However, both tree detection approaches were utilized as stem-based detection alone often yielded an insufficient number of trees for a reliable transformation. The transformation was unreliable when the number of located trees in the target point cloud was low compared to the number of trees found in the manual campaigns. This issue was more common in the difficult plots, as these plots contained more understory trees that were not visible with the raster-based method and had trunks that could also not be found with the stem curve method. We solved this problem with two modifications in the georeferencing procedure.

First, only manually measured trees exceeding a specified height threshold were used as reference trees. The threshold varied depending on both the properties of the target point cloud and the forest plot density. This modification ensured that reference trees corresponded primarily to larger trees that had stems that could be detected in the target point cloud, with smaller trees that were more difficult to detect discarded. Thus, there was better correspondence in the stem count between the manual locations and point cloud-derived locations for the transformation. Second, some point clouds benefited from a coordinate transformation between the tree locations that were found from the 2020 MLS point clouds used for preliminary tree segmentation and that were georeferenced to the global coordinate system earlier. This was because the number of trees found in that point cloud and in the target point cloud corresponded better. Finally, the quality of the coordinate transformation was examined visually, and point clouds with transformations that needed fine-tuning were adjusted manually.

We selected the manually measured tree locations as the reference coordinate system because their $xy$ coordinates were already available in a well-defined global coordinate system (ETRS-TM35-FIN). However, the specific choice of reference coordinate system itself is arbitrary, provided that all datasets are accurately transformed into a common coordinate system.

\subsection{Preliminary individual tree segmentation}
\label{sec:preliminary_segmentation}

The objective of ITS was to accurately delineate as many trees as possible from each input point cloud while minimizing false positives. In general, this objective is most reliably achieved using dense point clouds with minimal occlusion. MLS data was therefore considered particularly suitable, as it typically contains fewer occlusions compared to data captured with stationary TLS systems, while also providing better coverage of understory trees than data acquired from above the canopy. Therefore, we selected the 2020 MLS point cloud for the ITS step.

Individual trees were segmented from the 2020 MLS plots using the deep learning-based ForestFormer3D model \citep{Xiang2025a}, which represents the current state of the art for individual tree segmentation in boreal forests according to a recent benchmark on high-density ALS data \citep{Ruoppa2026}. ForestFormer3D is a transformer-based panoptic segmentation framework adapted from OneFormer3D \citep{Kolodiazhnyi2024}, with several modifications specifically designed for forest data.

ForestFormer3D begins by voxelizing the input point cloud and extracting 32-dimensional feature vectors from the data using a sparse convolutional 3D U-Net. These features are subsequently split into two distinct branches: one for learning 5-dimensional instance-discriminative feature embeddings, and another for classifying voxels as foreground (tree) or background (non-tree). A set of $K_{\text{ins}}\in\mathbbm{Z}^+$ instance queries is then selected by applying farthest point sampling (FPS) in the 5-dimensional embedding space of predicted foreground voxels. These $K_{\text{ins}}$ instance queries, together with $K_{\text{sem}}\in\mathbbm{Z}^+$ semantic queries, are provided as input to a query decoder composed of six transformer layers, where the 32-dimensional U-Net features serve as keys and values. The decoder directly outputs $K_{\text{ins}}$ predicted instance masks with associated confidence scores, as well as $K_{\text{sem}}$ semantic masks. Crucially, the decoder outputs correspond directly to predicted tree instances, eliminating the reliance on user-defined hyperparameters inherent to previous deep-clustering-based ITS approaches \citep[see e.g.][]{Wielgosz2024,Henrich2024}. During training, ForestFormer3D employs one-to-many association, where each ground truth instance is matched to all queries that fall within its mask. During inference, duplicate predictions are removed based on the associated confidence scores.

Since entire forest plot point clouds are typically too large to fit into GPU memory, ForestFormer3D processes the data using an overlapping cylindrical sliding window. Predictions from individual cylinders are then combined into a single segmentation during post-processing using a novel confidence-score-based merging algorithm.

We used the official PyTorch implementation of ForestFormer3D\footnote{\url{https://github.com/SmartForest-no/ForestFormer3D}} for performing the segmentation. As only the predicted tree instances were required, semantic predictions were discarded prior to subsequent processing steps. All model hyperparameters were set to their default values, including the input cylinder radius at 16 meters. Publicly available pretrained weights trained on the FOR-InstanceV2 dataset \citep{Xiang2025a,Xiang2025b} were used, since the dataset includes forest point clouds from a wide range of forest environments and sensor modalities, including high-density TLS and MLS data, making it well suited for segmenting MLS data depicting boreal forests.

ForestFormer3D employs a fixed number of non-parametric instance queries (300), which are selected from predicted foreground points using FPS. Consequently, the model may fail to detect some instances if no adequately positioned queries are assigned to them. This issue is particularly prominent in extremely dense forests, where the number of tree instances within an input cylinder could even exceed the number of available queries. To mitigate this limitation, ForestFormer3D provides a framework for multi-iteration inference, in which the model is first applied to the full input point cloud, and subsequently to the remaining points that were not assigned to any instance in previous iterations. The predictions from all iterations are then merged into a single segmentation during post-processing.

Although the post-processing removes any obviously erroneous segments based on their geometric properties, the multi-iteration inference inadvertently introduces a certain amount of false positive segments in most cases. However, we found that in dense forest plots, the number of missed detections after a single inference pass was so high that the slight increase in false positives was an acceptable trade-off for our use case, where maximizing the number of detected trees was the primary objective. A total of two rounds of inference were performed for all plots in the 2020 MLS dataset.

\subsection{Transferring the tree segmentation across point clouds}
\label{sec:segmentation_transferring}
Since the characteristics of the point clouds used in this study differ substantially, independently segmenting each point cloud would not necessarily result in the detection of the same set of trees across all datasets. This issue is particularly pronounced for small understory trees, which are often difficult to segment from sparse ALS point clouds, even when using state-of-the-art deep learning segmentation methods. Furthermore, independently generated segmentations would not guarantee correspondence of the trees and their segment labels across point clouds, as the same tree could be assigned different segment labels in different datasets. Such inconsistencies would reduce the size of the common set of trees found in all point clouds and make identifying the same tree from all point clouds more challenging. Therefore, the segmentation obtained from the 2020 MLS dataset was transferred to all other point clouds in this study.

Prior to segmentation transfer, digital terrain models (DTMs) were created for both the source and target point clouds in order to normalize the $z$ coordinates. We used a voxel-based algorithm for producing the DTM introduced in \citet{Hyypp2020}. First, the point cloud was rasterized into pixels, after which the points inside each pixel were divided vertically into voxels. The DTM was determined as the mean height of the points in the lowest voxel containing enough points (> 0.5\% of all points within the pixel). Finally, the resulting DTM was smoothed using a Gaussian blur method. Only points classified as background (non-tree) by the segmentation were used as input to the DTM algorithm. We used a variable pixel side length between 1--2.5 m and a voxel height of 1.5 m. Larger pixel sizes helped compensate for holes in the DTM image from patchy point clouds.

Segmentation transfer was implemented using a kd-tree-based method using Matlab's (version 2025a) \verb|KDTreeSearcher| functionality. A kd-tree partition was calculated for the segmented MLS 2020 source point clouds. Next, the target point cloud was input to the kd-tree partitioner, which found the closest points in the source point cloud to the points in the target point cloud. The found closest points were then assigned the same point labels as in the source point cloud. Importantly, we used every 100\textsuperscript{th} point from the source point cloud and a leaf node size of 25 points when creating the kd-tree model, in order to decrease the partitioning model creation time.

This method of segmentation transferring relied on very accurate point cloud georeferencing. Ideally, differences in the locations of individual tree parts between datasets should only reflect tree growth and small variations caused by factors such as wind conditions, branch orientation, and laser scanning parameters \citep{Honkanen2025}. Following segmentation transfer, the transferred segmentation was examined to verify accurate ground level alignment and to identify potential segmentation errors. After transferring the segmentation to all point clouds, each tree could be identified across all point clouds by a tree ID consisting of similarly labeled points.

\subsection{Matching trees across different datasets}
Individual trees detected from the point clouds were matched across datasets using the tree IDs obtained from the tree segmentation. Because the transferred segmentation provided consistent tree delineations across all point clouds, the same tree could be identified throughout the multitemporal point clouds using the corresponding tree ID.

The manually measured trees did not share these same tree IDs, and therefore required a separate matching procedure. Trees with detected stems and stem curve information from the point clouds, as described in Section \ref{sec:stemcurve}, were matched to the manual measurements using the registration and matching approach outlined in Section \ref{sec:georeferencing}. This method both aligns the target and reference point clouds based on tree locations and establishes tree correspondences, with a 0.5 m threshold used for matching the trees. The common set of trees analyzed in this study was based on these matches.

In some cases, multiple stems were detected from a single segment and therefore shared the same tree ID. Because these stems were spatially close, the algorithm initially matched them either to the same manually measured tree or to different nearby manual trees, depending on the dataset. In these cases, if all detected stems across all years were matched to a single manual tree, that tree was retained as the reference and the stem with its center closest to the manual tree was matched. In cases where multiple manual trees had been matched, if at least one year contained only a single detected stem matched to a single manual tree, then this match was used to define the reference tree. In all other years, the stem closest to this reference manual tree was matched. In contrast, if all years contained multiple detected stems matched to more than one manual tree, each stem was matched to the nearest manual tree within the 0.5 m threshold, and multiple matches were retained.

Overall, this matching approach enabled the retention of stem-level results from segments containing multiple detected stems, thereby increasing the common set of trees included in this study. However, despite the distance-based matching approach and the maintaining of one-to-one correspondence between detected and manual reference trees, incorrect matches may still have occurred, particularly in dense plots where neighboring stems are in close proximity. These mismatches may propagate into subsequent analyses, appearing as increased errors in the modeling and growth estimation results where manually measured values are involved.

\subsection{Stem curve extraction}
\label{sec:stemcurve}
Tree stem curves were estimated using the algorithm first presented for 2D scanners in \citet{Hyypp2020}, and later adapted for 3D scanners \citep{hyyppa2021under} and ALS \citep{hyyppa2022direct}. We used the individual tree segments as inputs for the algorithm. Overall, for each identified tree segment, the algorithm searched for stem arcs that were subsequently used to generate the stem curve of the tree. These stem arcs corresponded to clusters of points that typically originate from laser hits on a tree stem.

The algorithm starts by dividing the point cloud into height intervals with bin heights of 0.2 m. From each of these intervals, stem arcs were determined using density-based clustering of applications with noise \citep[DBSCAN,][]{ester1996density}, where a core point had 9 points within a 7.5 cm neighborhood. Circles were then fitted to the detected stem arcs using a random sample consensus-based \citep[RANSAC,][]{fischler1981random} method. An arc candidate was accepted if it had at least 35 points, with 80\% of the points being within 3 cm of a circular fit to the points. An iterative arc division step was next used to remove noise points by creating sub-arcs if neighboring points had a central angle over 10\degree. The arc candidate was accepted if it met quality criteria, including that it contained over 35 points, possessed a central angle greater than 108\degree, had a diameter between 5 cm and 80 cm, as well as a standard deviation of radial residuals smaller than 1.5 cm.

To differentiate a tree from a false positive segment, center points of the accepted arcs were further clustered using DBSCAN. Here, an arc center was determined to be a core point when it had at least 5 points within a 25 cm neighborhood. If the \textit{z} coordinates of the clusters of arc center points spanned over 1 m vertically, then the points were considered a tree. Principal component analysis (PCA) was then used to find the growth direction of the tree. On a plane perpendicular to this direction of growth, the inliers of the arc points had a circle re-fitted to them according to an algebraic fit by \citet{al2009error}. Finally, the stem curves were found by using a cubic smoothing spline fitted to the mean stem diameter estimates, with the stem arcs of the tree split into 40 cm intervals along the \textit{z} direction.

\subsubsection{Tree attribute estimation from the stem curve}
\label{sec:attribute_methods}
Tree heights were computed as the vertical difference between the DTM and the highest points within each segment. The DTM was generated using the same procedure as described for segmentation transfer in Section \ref{sec:segmentation_transferring}. Even when using a relatively large pixel size, some holes in the DTM image were still observable after the DTM creation. These affected height estimates of trees that were located where the holes in the DTM were. To address this, we fixed the affected DTMs using Matlab's \verb|imfill| operation \citep{Soille1999}.

Heights were estimated for the subset of trees for which stems were detected in each year from MLS/TLS data, using both ALS-only point clouds and a combination of ALS and under-canopy MLS/TLS point clouds. We removed trees that contained points assigned to multiple different segments (see Appendix \ref{app:errors_in_segmentation_and_stem_curve_extraction} for a detailed description), ensuring that the height of each tree was determined using points from only one segment. In cases where the tree was split across segments, choosing the segment that would correctly define the tree height was unclear. While generally this could be possible by using the stem coordinates as a tool to decide which segments the points belong to, this approach was not adopted in order to maintain a robust and unambiguous processing workflow. The remaining subset of trees after this removal formed the common set of trees analyzed in this study.

The heights of the remaining trees that belonged to a single segment were derived using a voxel-based approach. Using only points belonging to the specific tree ID, the tree segment was divided into intervals of 0.5 m in the $z$ direction. A cylinder was formed around the tree $xy$ coordinate that was obtained from the stem location and points within 1 m of the stem center line were used. Tree height was calculated as the average of the three highest points within the highest voxel containing at least five points. When incorporating the MLS/TLS data, this stem location was obtained from the stem center of the detected tree stem in the segment. However, when estimating the heights from the ALS-only point clouds, the segments did not contain any directly located stems. Therefore, the centers of the tree stems from the 2021 TLS data were used for the ALS-only height estimation in all years. This choice was made in order to use more precise tree center locations compared to what could be obtained from only the tree segments of the ALS data. It should be noted that tree locations from any of the years or point clouds with located stems could have been used, since the main motivation was simply to use more accurate stem locations. Additionally, using accurate stem locations enabled the estimation of height for the cases where more than one stem was found from a single tree segment. In these instances, despite the points of the tree segment used for height estimation being the same for both stems, the accurate stem center location enabled the estimation of the height in the correct location around the tree.

Because the MLS/TLS point clouds collected inside the forest generally underestimate tree height \citep{Wang2019}, which is a critical measurement when determining tree volume, heights were also estimated by augmenting the MLS/TLS data with the ALS data from the same year in order to achieve more accurate height determination. However, only the highest 10\% of ALS points were included. We were therefore able to capture the tree height more accurately with the added ALS points, but did not interfere with the stem curve estimation that used the MLS/TLS point clouds by adding potentially noisy or slightly offset points from ALS, which could still be a possibility to some extent even after accurate coordinate transformation.

From the stem curves of the trees, the DBH was found by interpolating from the spline fit at a 1.3 m height. The DBH was extrapolated for trees that contained measurements only above this height, according to the method outlined in \citet{Hyypp2020}. A linear fit was used for stems where diameter measurements spanned more than 3 m, whereas a square root function was used when the diameters spanned a smaller range.

The stem curves and estimated heights were then used to find the stem volume of the tree. A parabolic function of height $z$ was fitted to the estimates of the stem radii of the detected tree using the least squares-method
\begin{equation}\label{eq:parabola}
    R_1(z) = a_1(h-z)^2 + a_2(h-z), \\
\end{equation}
as well as a square root function
\begin{equation}\label{eq:sqrt}
    R_2(z) = b_1\sqrt{h-z},
\end{equation}
with $h$ referring to the tree height, and $a_1$, $a_2$, and $b_1$ referring to parameters from the fit. The stem volume was then calculated by integrating the Eqs. \eqref{eq:parabola} and \eqref{eq:sqrt} as a solid of revolution, taking an average,
\begin{equation}\label{eq:volume}
    V = \frac{\pi}{2} \left( \int_0^h R_1^2(z) \text{d}z + \int_0^h R_2^2(z) \text{d}z \right) = \pi \int_0^hR_\textit{eff}^2(z)\text{d}z.
\end{equation}

\subsection{Scaling model for DBH and stem volume change}
\label{sec:change_model}
The estimation of DBH and stem volume change was based on modeled stem attributes derived from tree height change. We scaled the source tree diameter $D_\mathit{src}$ at height $z$ using the height growth factor $b$ to obtain the target diameter $D_\mathit{tgt}$. Mathematically, the model is defined as
\begin{equation}
    \label{eq: scaled_diameter_without_form}
    D_\mathit{tgt}(z) = b \cdot D_\mathit{src}(z). 
\end{equation}
Importantly, the argument $z$ took values in a half-open interval $z \in [0, \min(h_\mathit{src},h_\mathit{tgt}))$ because the diameter at $h_\mathit{src}$ or $h_\mathit{tgt}$ equaled zero. The factor $b$ depends on the source and target points-in-time ($t_\mathit{src}$ and $t_\mathit{tgt}$, respectively) and is defined as 
\begin{align}
    b &= \min(1, h_\mathit{tgt}/h_\mathit{src}),\,  t_\mathit{src} > t_\mathit{tgt} \\
    b &= \max(1, h_\mathit{tgt}/h_\mathit{src}),\,  t_\mathit{src} < t_\mathit{tgt} \\
    b &= 1,\, t_\mathit{src} = t_\mathit{tgt}.
\end{align}
The minimum and maximum functions ensured that the tree trunk did not expand when hindcasting backwards in time or shrink when forecasting forward in time, which would be against our assumptions of tree growth. The scaled DBH was calculated directly from Eq. \eqref{eq: scaled_diameter_without_form} at $z = 1.3$ m. Stem volume was calculated by fitting the Eqs. \eqref{eq:sqrt} and \eqref{eq:parabola} to an extended set of data points $\{(D_\mathit{tgt}^{(i)}/2, z^{(i)})\}_{i=1}^n \cup (0, h_\mathit{tgt})$. 

Previous studies have shown that stem taper changes as a tree grows \citep{luoma2019examining, luoma2021revealing}. To account for this effect, we also calculated a form factor $f$ which worked similarly to the growth factor $b$, but was defined using the stem taper functions fitted to Finnish forest conditions by \citet{Laasasenaho1982}. The functions are based on the relative tree diameter 
\begin{equation}
    \frac{D(z)}{D_{0.2h}} = f_s(z) + f_t(z\,|\,\text{DBH}, h) = f_c(z\,|\,\text{DBH}, h),     
\end{equation}
where $D_{0.2h}$ is the diameter at 20\% of the tree height, $f_s$ is the species-wise form function, and $f_t$ is the tree-wise form function that depends on the tree species as well as DBH and height. The species-wise function $f_s$ is a polynomial with powers defined by the Fibonacci series up to a degree of 34 (\citet{Laasasenaho1982}, Eqs. (33.1) and (41.1)), and the tree-wise function $f_t$ is a cubic spline fitted to points defined by Eqs. 41.2 in \citet{Laasasenaho1982}, or to capped absolute values of 0.1, as per the definition. We fixed an error in the second equation for spruce by changing the term $d^2(h-1.3)^2$ to $d^2/(h-1.3)^2$ since the error caused unnatural stem tapers.

\begin{figure}[htb!]
    \centering
    \includegraphics[width=\linewidth]{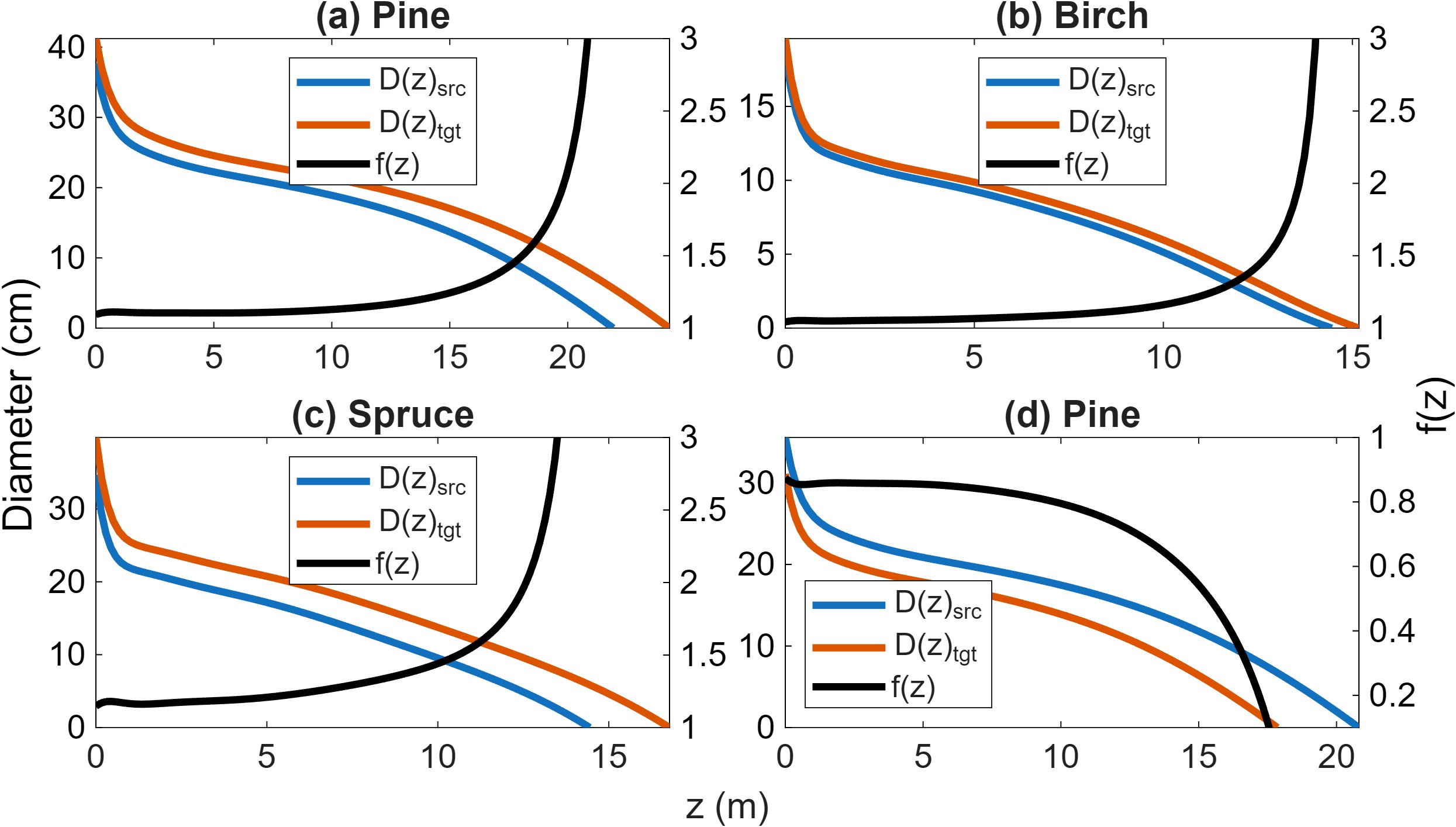}
    \caption{Examples of the prototype taper curves (Eq. \eqref{eq: scaled_diameter_with_form}) and the weighting function $f$ (Eq. \eqref{eq:f}). The target taper curve is obtained using the scaled DBH as the input for the stem taper equation (nominator in Eq. \eqref{eq:f}). In Figs. 3a--c, the source time is 2014 and target time is 2025. In Fig. 3d, the source time is 2025 and the target time is 2014.}
    \label{fig:example_stem_curves_laasasenaho}
\end{figure}

The form factor $f$ was solved using the following procedure. First, the $D_{0.2h}$ for the source time $t_\mathit{src}$ was estimated using the procedure suggested by \citet{Laasasenaho1982} 
\begin{equation}
    \hat{D}_{0.2h,\, \mathit{src}} = \frac{\text{DBH}_\mathit{src}}{f_c(z=\mathrm{1.3\,m}\,|\, \text{DBH}_\mathit{src}, h_\mathit{src})}.
\end{equation}
The procedure was then repeated for the target time $t_\mathit{tgt}$ using the growth factor $b$ as an initial guess for the DBH change
\begin{equation}
    \hat{D}_{0.2h,\, \mathit{tgt}} = \frac{b \cdot \text{DBH}_\mathit{src}}{f_c(z =\mathrm{1.3\, m}\,|\,b \cdot \text{DBH}_\mathit{src}, h_\mathit{tgt})}. 
\end{equation}
The form factor $f$ was then calculated as the ratio
\begin{equation}
    \label{eq:f}
    f(z) = \frac{D_\mathit{tgt}(z)}{D_\mathit{src}(z)} = \frac{\hat{D}_{0.2h,\, \mathit{tgt}} \cdot f_c(z\,|\,b \cdot \text{DBH}_\mathit{src}, h_\mathit{tgt})}{\hat{D}_{0.2h,\, \mathit{src}} \cdot f_c(z\,|\,\text{DBH}_\mathit{src}, h_\mathit{src})}.
\end{equation}
This enabled us to model the change in stem taper induced by tree growth with the model
\begin{equation}
    \label{eq: scaled_diameter_with_form}
    D_\mathit{tgt}(z) = f(z) \cdot D_\mathit{src}(z), \quad z \in [0, \min(h_\mathit{src},h_\mathit{tgt})).
\end{equation}
The model is only valid for pine, spruce, and birch, for which the equations were originally developed. Note, however, that $f(z = \mathrm{1.3\, m}) = b$, meaning that the form model does not apply to DBH modeling. Stem volume was then calculated as before by fitting Eqs. \eqref{eq:sqrt} and \eqref{eq:parabola} to the new radii estimates. Fig. \ref{fig:example_stem_curves_laasasenaho} shows examples of the taper curves calculated using the Laasasenaho models and the weighting function in Eq. \eqref{eq:f}. 

\subsection{Change estimation}

In this study, we set the starting year for one-time measurements to 2014 when forecasting, whereas the starting year was 2025 when hindcasting one-time values. For the growth analyses, we used 2024 as the starting year to hindcast the DBH and stem volumes of individual trees to obtain growth over 10 years (2014--2024) and 5 years (2019--2024).

In the model-based approach, growth was estimated using the tree attributes obtained from the 2024 point clouds as well as the modeled values in 2019 and 2014 to estimate the growth as
\begin{equation}
    \Delta \hat{a}^{(i)}_\text{model} = a_{24}^{(i)} - \hat{a}^{(i)}_{14/19},\quad i\in [1, n_\text{trees}],\,a \in \{ \text{DBH},\, \text{stem volume} \}.
\end{equation}
The direct approach used one-time DBH and stem volume estimates determined independently for each year from combined ALS and MLS/TLS point clouds. These point clouds were used to obtain tree height, stem curve, and DBH, with these directly derived attributes then used to calculate the change as
\begin{equation}
    \Delta a^{(i)} = a_{24}^{(i)} - a^{(i)}_{14/19}.
\end{equation}
Unlike the model-based framework, the direct approach assumes no prior information about the forest under study and requires both over-canopy and under-canopy measurements at every observation time.

The manual approach to tree attribute change estimated the change similarly to the direct approach. DBH was measured manually, while stem volume was estimated using the Laasasenaho allometric model with the manually measured DBH, species, and tree height estimated from the combined ALS and MLS/TLS point clouds as inputs. Heights derived from the combined point cloud data were used, as they were assumed to provide the most accurate height estimates. Comparing the model-based, direct, and manual approaches enabled assessment of the reliability of the different growth estimation methods.

\subsection{Segmentation accuracy}
We counted the segments outputted by the segmentation method and the trees with a detected stem curve in all MLS/TLS point clouds in order to analyze the success of the preliminary segmentation and the final number of trees included in the analysis, referred to as the common set of trees. The reference stem count was obtained by counting the manually measured trees. The trees in all categories were divided into height classes to evaluate the segmentation performance across different height intervals. We used the manual measurements of height from 2019 as the height value for the manually measured trees and the trees in the common set of trees.

The segments outputted by the deep learning segmentation were counted differently from the manually measured trees and the trees in the common set of trees. For the subset of segments belonging to the common set of trees, we used the associated 2019 manual measurements for the heights so that the segmented and reference counts were based on the same height values. Meanwhile, the remaining segmented trees without a detected stem were counted as follows. First, the point cloud was normalized by subtracting the DTM as described previously. Then, segment positions were computed as the mean \textit{xy} coordinates of the corresponding segments. Plot boundaries were calculated using Matlab's \verb|boundary| function with the tightness parameter set to 0, resulting in a convex hull. A buffer of 0.5 m was added to the boundary. Finally, segments whose locations fell within this boundary were retained, and their heights were determined as the height of their highest point. False positives segments were identified heuristically as segments that spanned $\leq1$ m in the \textit{z} direction ($z_\textit{max} - z_\textit{min} \leq \mathrm{1\,m}$) while exceeding 1 m in height ($z_\textit{max} > \mathrm{1\,m}$).

We fitted the Näslund equation \citep[see e.g.][]{Mehttalo2015, naslund1936skogsforsoksanstaltens}  
\begin{equation}
    \label{eq:naslund}
    h(\text{DBH}) = 1.3 + \frac{\text{DBH}^2}{(a\cdot \text{DBH} + b)^2}
\end{equation}
with fitting parameters $a$ and $b$ to manually measured DBH values and ALS + TLS height values of 2021. We used the fitted relationship for finding the threshold height corresponding to a DBH of 5 cm which was used as an exclusion principle of manually measured trees during the campaign. The threshold height was 7.64 m.

\subsection{Error metrics}
The statistical error analysis was evaluated using the RMSE, mean absolute error (MAE), and bias, together with their relative counterparts defined as
\begin{align}
    \text{RMSE} &= \sqrt{\frac{1}{n}\sum_{i=1}^n(y^{(i)}_\mathit{pred} - y^{(i)}_\mathit{ref})^2} \\
    \label{eq:rel_rmse}
    \text{RMSE\%} &= \frac{\text{RMSE}}{\frac{1}{n}\sum_{i=1}^n y^{(i)}_\mathit{ref}} \times 100 \% 
\end{align}

\begin{align}
    \text{MAE} &= \frac{1}{n}\sum_{i=1}^n |y^{(i)}_\mathit{pred} - y^{(i)}_\mathit{ref}| \\
    \text{MAE\%} &= \frac{\text{MAE}}{\frac{1}{n}\sum_{i=1}^n y^{(i)}_\mathit{ref}} \times 100 \%
\end{align}

\begin{align}
    \text{bias} &= \frac{1}{n}\sum_{i=1}^n(y^{(i)}_\mathit{pred} - y^{(i)}_\mathit{ref}) \\
    \text{bias\%} &= \frac{\text{bias}}{\frac{1}{n}\sum_{i=1}^n y^{(i)}_\mathit{ref}} \times 100 \%.
\end{align}
The relative metrics are especially useful for comparing results across different years, as they account for the growth of trees by using the yearly mean attribute value as the denominator. MAE measures the residual error similarly to RMSE, but is more robust in the presence of outliers \citep{Hodson2022}. The Pearson correlation coefficient (R) and coefficient of determination ($R^2$) were additionally computed between the different DBH and stem volume growth estimation approaches in order to determine the agreement between these approaches.

\section{Results}
\subsection{Stem detection rate}
\begin{figure}[htbp!]
    \centering
    \begin{subfigure}{\linewidth}
        \caption{\label{fig:stemcount1} Easy plots}
        \includegraphics[width=\linewidth]{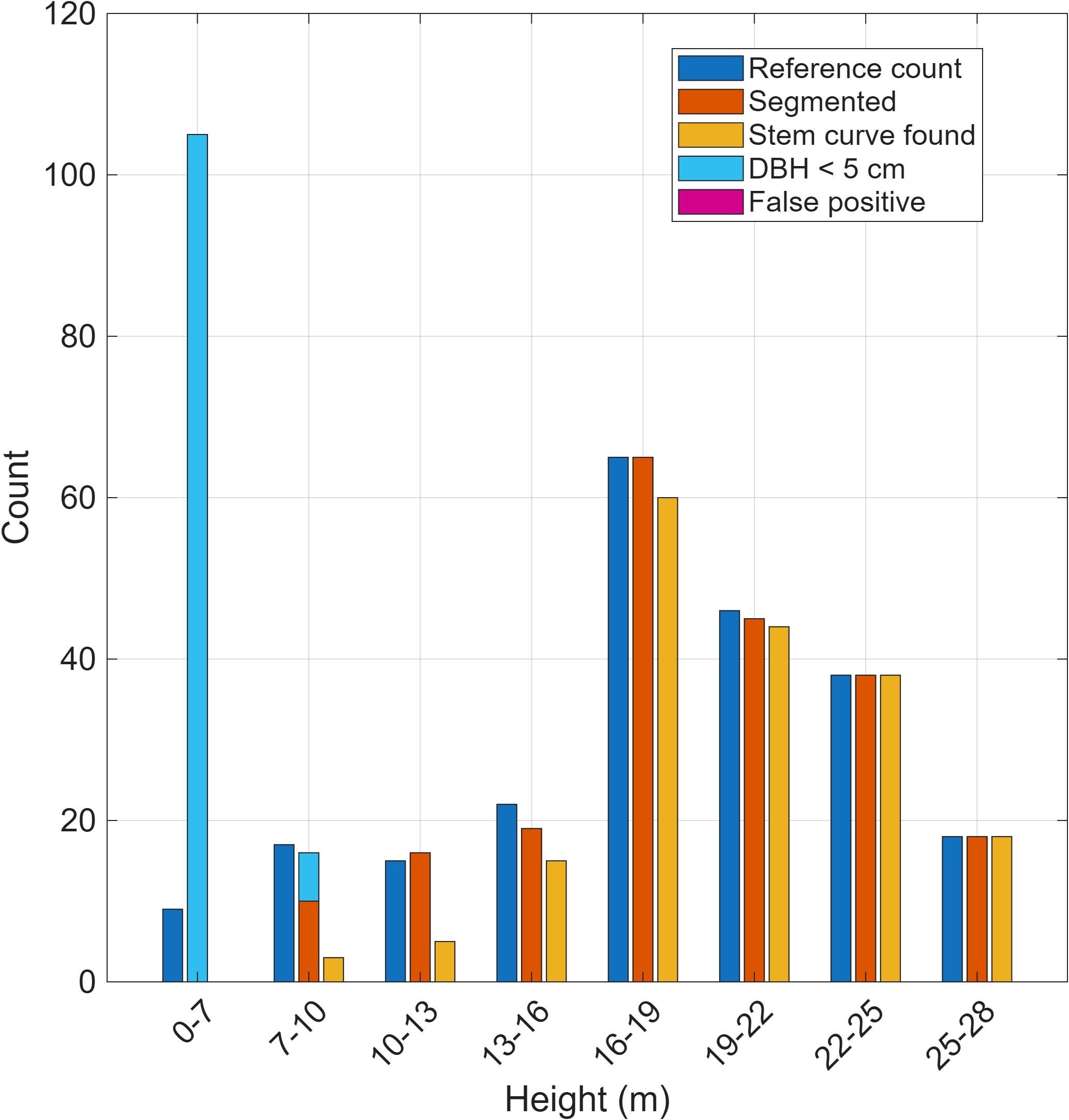}
    \end{subfigure} \\
    \begin{subfigure}{\linewidth}
        \caption{\label{fig:stemcount2} Difficult plots}
        \includegraphics[width=\linewidth]{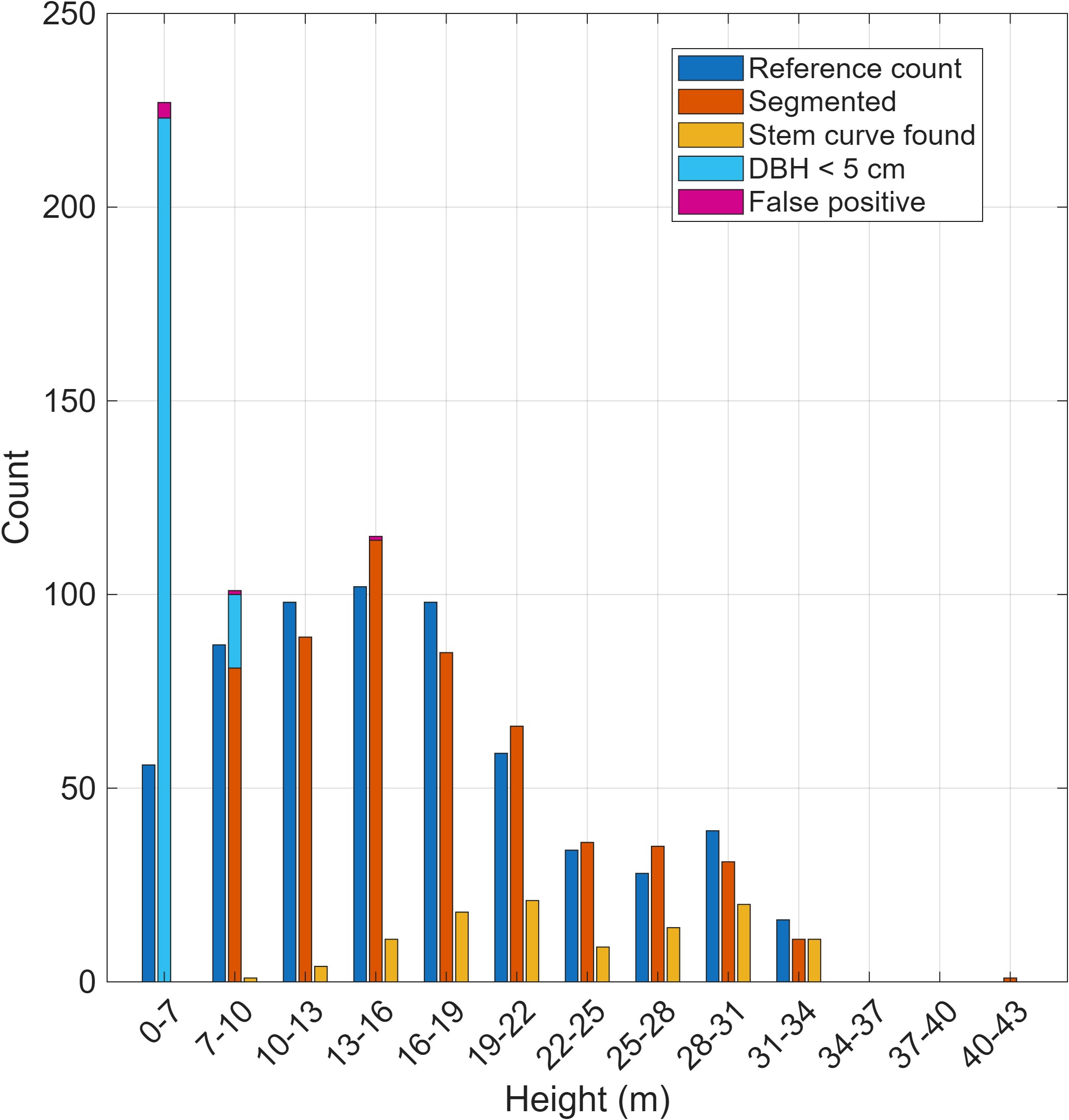}
    \end{subfigure}
    \caption{Tree counts by height interval. \textit{Reference count} corresponds to the manually measured trees. The bars labeled as \textit{segmented} and \textit{stem curve found} depict the output of the deep learning segmentation and the common set of trees in this study, for which stem curves were successfully detected, respectively. The bars labeled \textit{DBH < 5 cm} represent segments with a height below the threshold obtained by evaluating Eq. \eqref{eq:naslund}. The trees with height between 0--7 m are grouped into a single interval because the estimated threshold height for 5 cm DBH was 7.64 m. \ref{fig:stemcount1}: The distribution of detected trees across height intervals in easy plots. \ref{fig:stemcount2}: The same for difficult plots.}
    \label{fig:stemcount}
\end{figure}

The stem counts by height for both plot difficulty levels are presented in Fig. \ref{fig:stemcount} and the stem detection rates in Table \ref{tab:stem_detection_rate}. The resulting common set of trees totaled 183 trees in the easy plots and 109 trees in the difficult plots. During the subsequent analyses of this study, two trees were removed as described in Sec. \ref{sec:time_series_of_error_} and Appendix \ref{app:removed_trees_dbh_vol}. This decreased the number of analyzed trees in the difficult plots to 107.

In the easy plots in Fig. \ref{fig:stemcount1}, all height intervals above 7 m are almost completely segmented and stem curves are detected reliably from trees with a height over 13 m. There were some manually measured trees with a DBH over 5 cm and height less than 7 m, but there are no detected stem curves. The segmentation rates and stem curve finding rates are high, ranging between 80\%--98\%. No false positive segments were identified in the easy plots by the applied heuristic rule. In the difficult plots, as shown in Fig. \ref{fig:stemcount2}, the number of trees with a detected stem curve is low compared to the number of segmented trees, with 18\% stem curve detection rate. However, the reference and segmented counts show good agreement for trees taller than 7 m, and the segmentation rates are similar to the easy plots. The lowest height interval again contained more segmented trees than reference trees, and there are no found stem curves. A small number of false positive segments was also observed.

The results presented here should be regarded as approximate, as height estimation from both manual and MLS measurements produces errors, particularly with tall trees, which may shift trees between some adjacent intervals. Furthermore, the Näslund equation (Eq. \eqref{eq:naslund}) introduces some error affecting the number of segmented trees with height close to the threshold height of 7.64 m. The one-year difference between the manual and MLS height measurements is unlikely to introduce a significant error, because both acquisitions took place during the non-growing season between the measurement dates (see Table \ref{tab:data_characteristics}).

\begin{table}[htb!]
\centering
\caption{\label{tab:stem_detection_rate} Segmentation, stem detection, and false positive rates. The following notation and references to colors in Fig. \ref{fig:stemcount} are used: $n$ = total number of reference trees (dark blue), $\tilde{n}$ = number of reference trees with a height exceeding the 5 cm DBH threshold height (Eq. \eqref{eq:naslund}), seg = number of segmented trees exceeding the 5 cm DBH threshold height (red), sc = number of trees with a successfully detected stem curve (orange), fp = number of false positives (magenta).}
\begin{tabular*}{\linewidth}{@{\extracolsep\fill}lllll@{}}
    \toprule
              & seg/$n$ (\%) & seg/$\tilde{n}$ (\%) & sc/$n$ (\%) & fp/$n$ (\%) \\ 
    \midrule
    Easy      & 92           & 98                   & 80          & 0 \\
    Difficult & 89           & 101                  & 18          & 1 \\
    \bottomrule
\end{tabular*}
\end{table}

\subsection{Segmentation transferring and measurement time series}
\begin{figure*}[htbp!]
    \centering
    \begin{subfigure}{0.95\textwidth}
        \caption{\label{fig:example_tree1} Easy}
        \includegraphics[width=\linewidth]{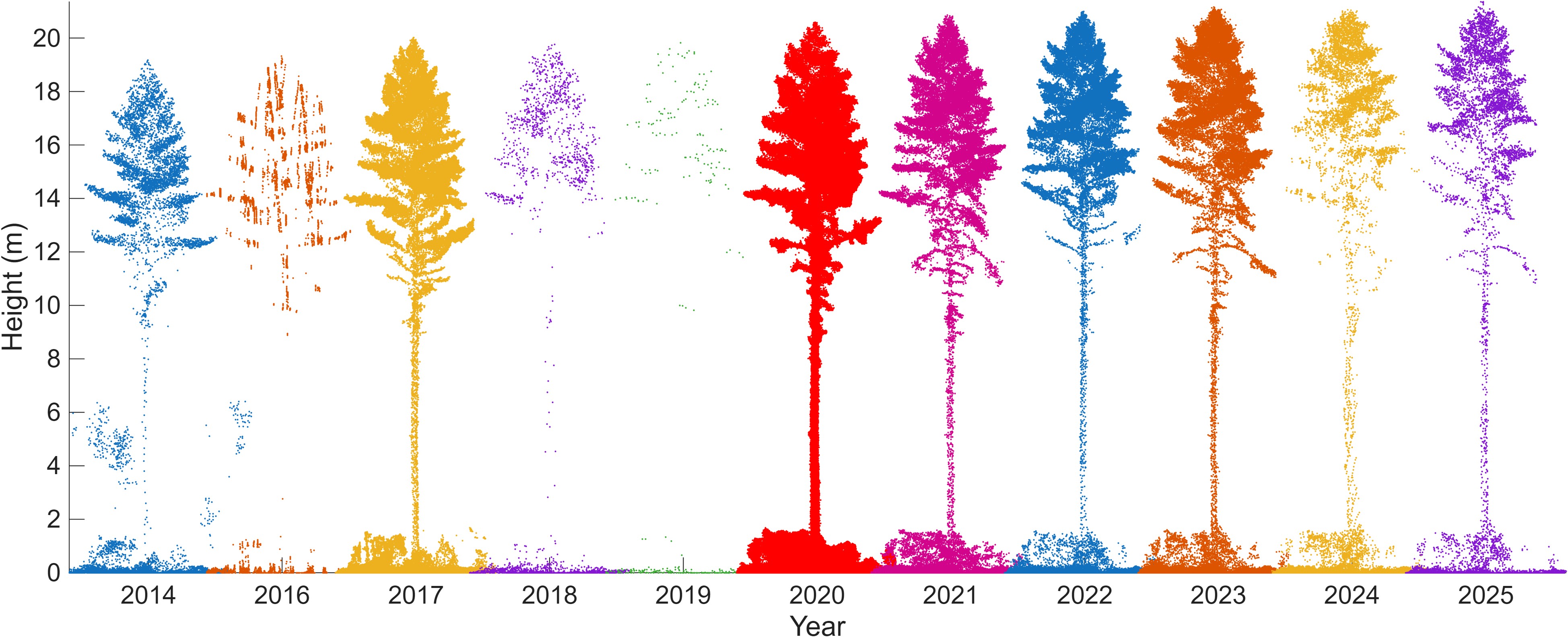}
    \end{subfigure} \\
    \begin{subfigure}{0.95\textwidth}
        \caption{\label{fig:example_tree2} Difficult}
        \includegraphics[width=\linewidth]{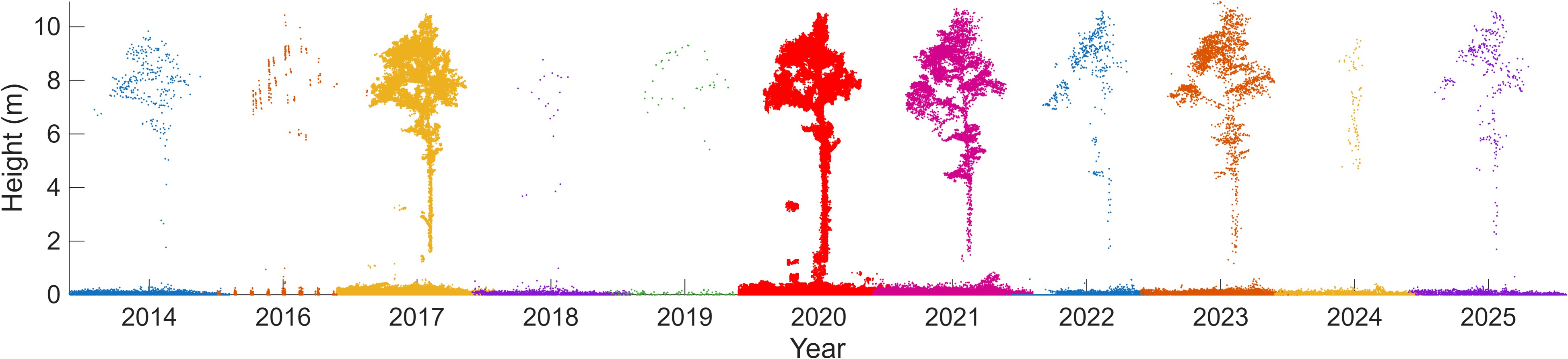}
    \end{subfigure} \\
    \begin{subfigure}{0.35\textwidth}
        \caption{\label{fig:example_tree3} }
        \includegraphics[width=\linewidth]{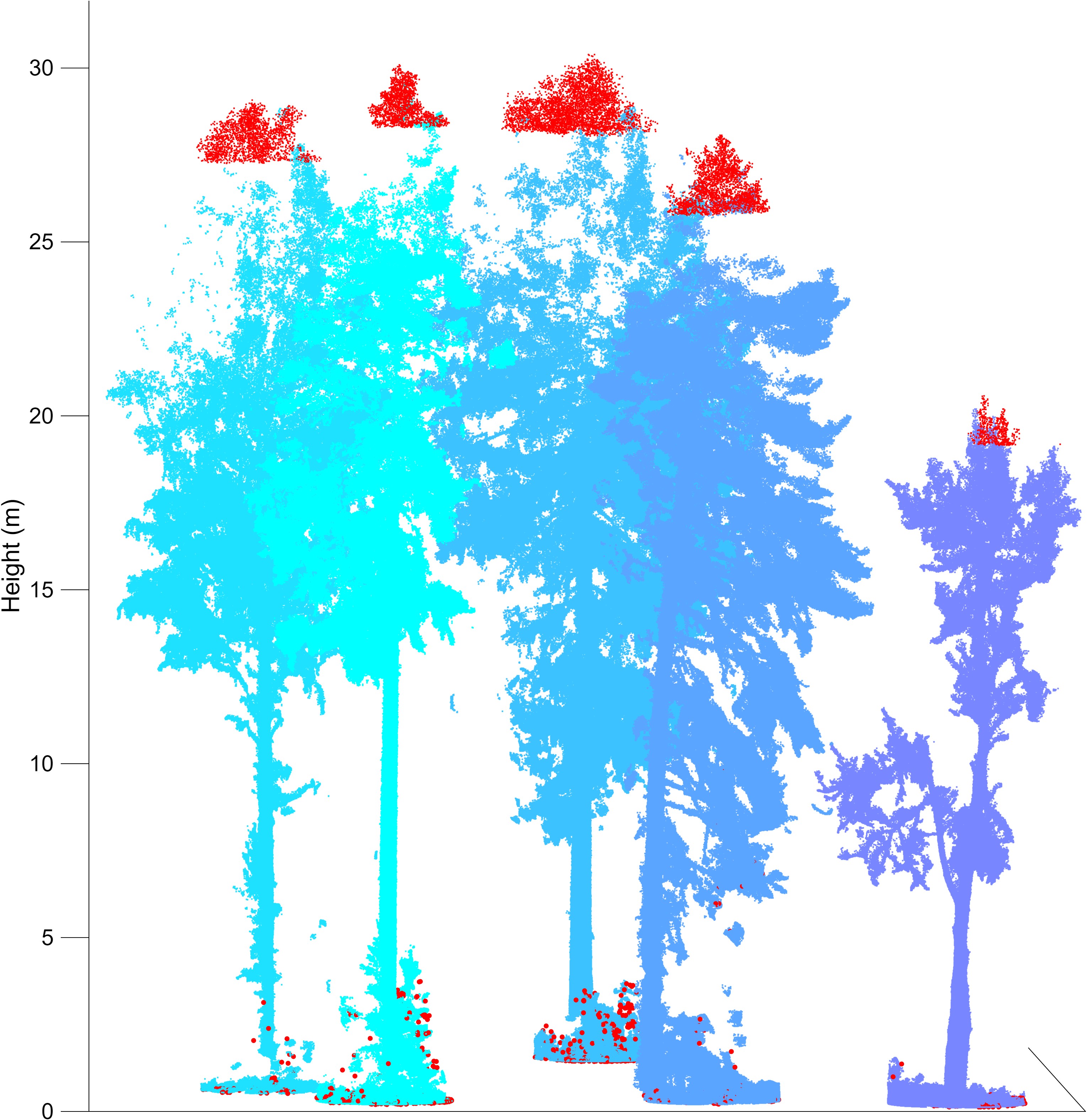}
    \end{subfigure}
    \caption{\label{fig:example_tree} An example of how the segment transferring appears in \ref{fig:example_tree1}: easy plots and \ref{fig:example_tree2}: difficult plots. All point clouds were acquired using ALS scanners, apart from the 2020 point cloud (red), which is the originally segmented 2020 MLS point cloud. The tree point clouds also include some points from the background (non-tree) class, which mainly correspond to ground hits. The tree in the easy plot was visible from above, whereas the tree in the difficult plot was partially covered by neighboring trees. \ref{fig:example_tree3}: An example of how combining ALS and MLS point clouds completed canopy representation and improved height estimation from five trees in a difficult plot in 2025. Blue points are from an MLS sensor, whereas the red points correspond to the highest 10\% of points in the ALS point cloud, which was the setup used for inputting the point clouds into the stem curve algorithm. Ground points are also shown.}
\end{figure*}

\begin{figure*}[htbp!]
    \centering
    \begin{subfigure}{0.43\textwidth}
        \caption{\label{fig:time_series1.0} Height (ALS only), easy}
        \includegraphics[width=\linewidth]{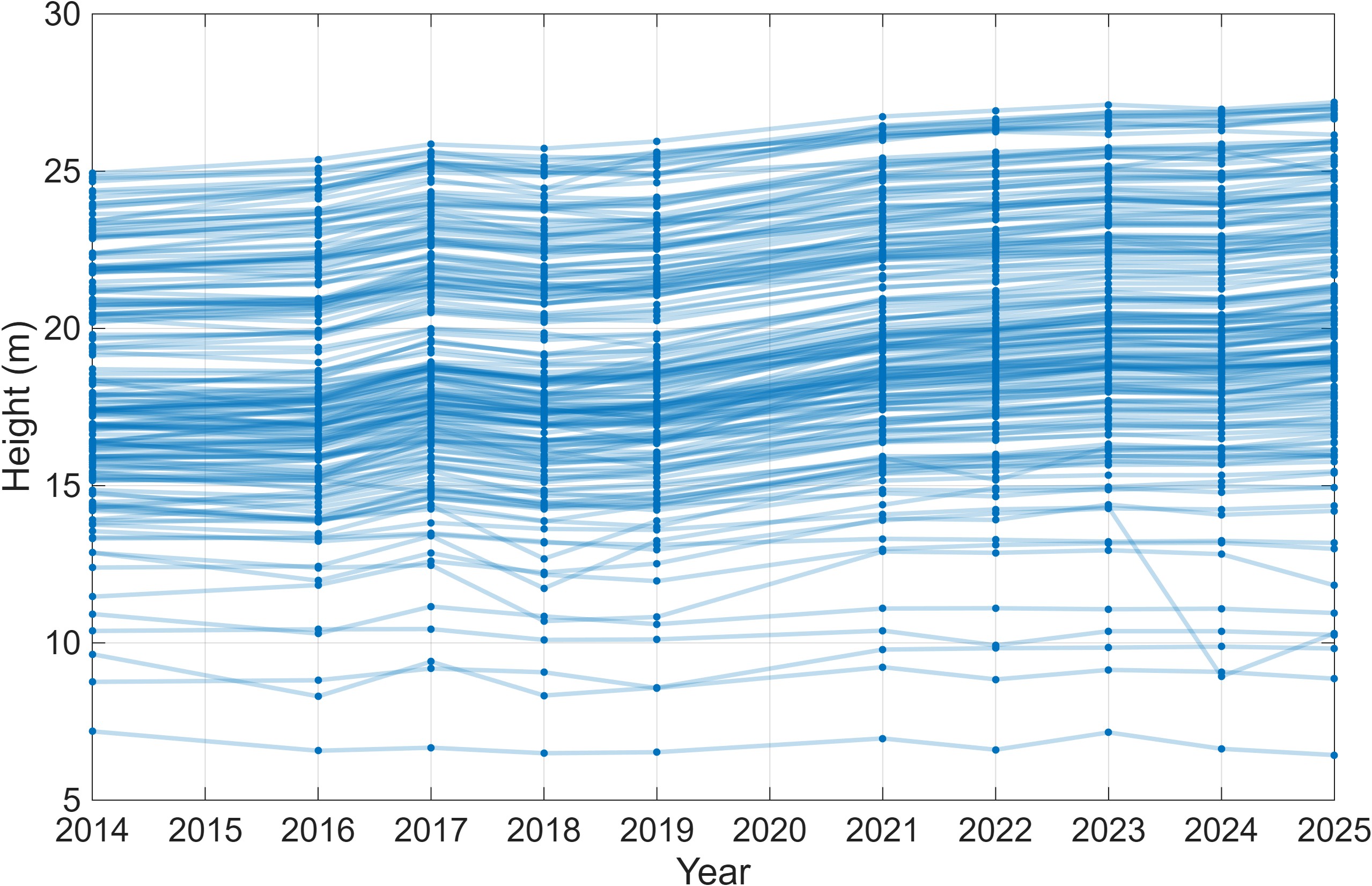}
    \end{subfigure}
    \begin{subfigure}{0.43\textwidth}
        \caption{\label{fig:time_series2.0} Height (ALS only), difficult}
        \includegraphics[width=\linewidth]{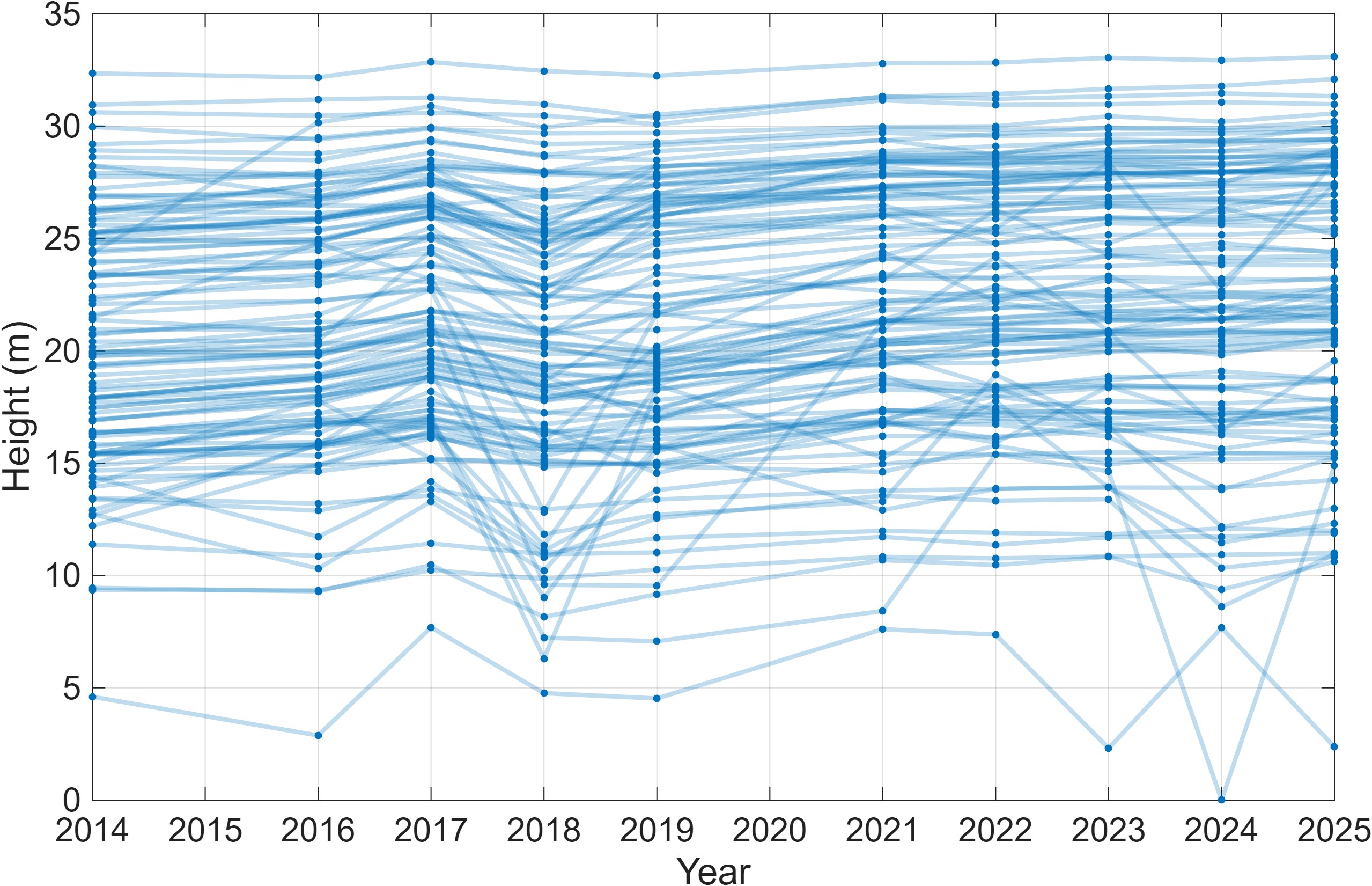}
    \end{subfigure}
    \begin{subfigure}{0.43\textwidth}
        \caption{\label{fig:time_series1.1} Height (ALS + TLS/MLS), easy}
        \includegraphics[width=\linewidth]{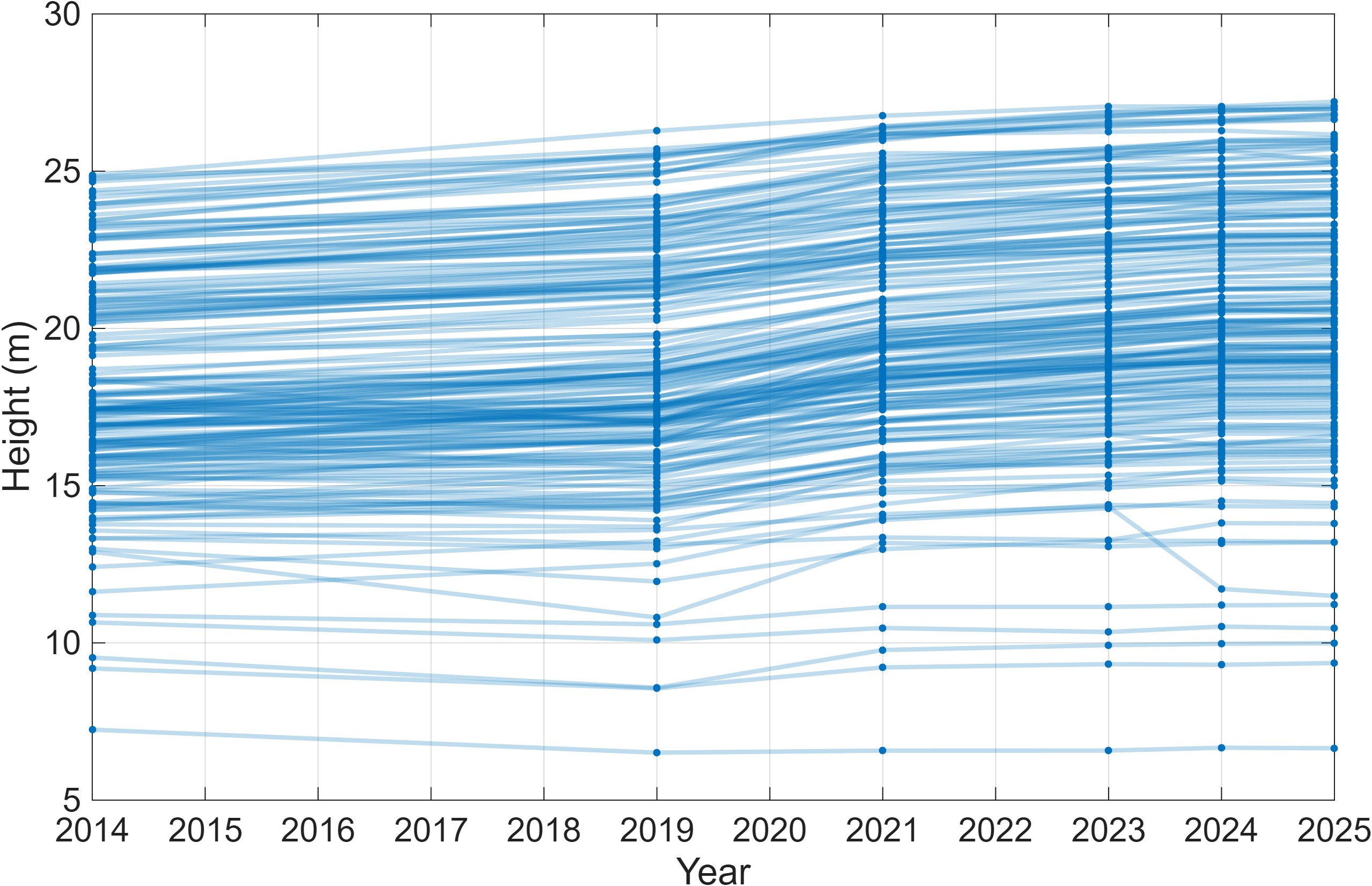}
    \end{subfigure}
    \begin{subfigure}{0.43\textwidth}
        \caption{\label{fig:time_series2.1} Height (ALS + TLS/MLS), difficult}
        \includegraphics[width=\linewidth]{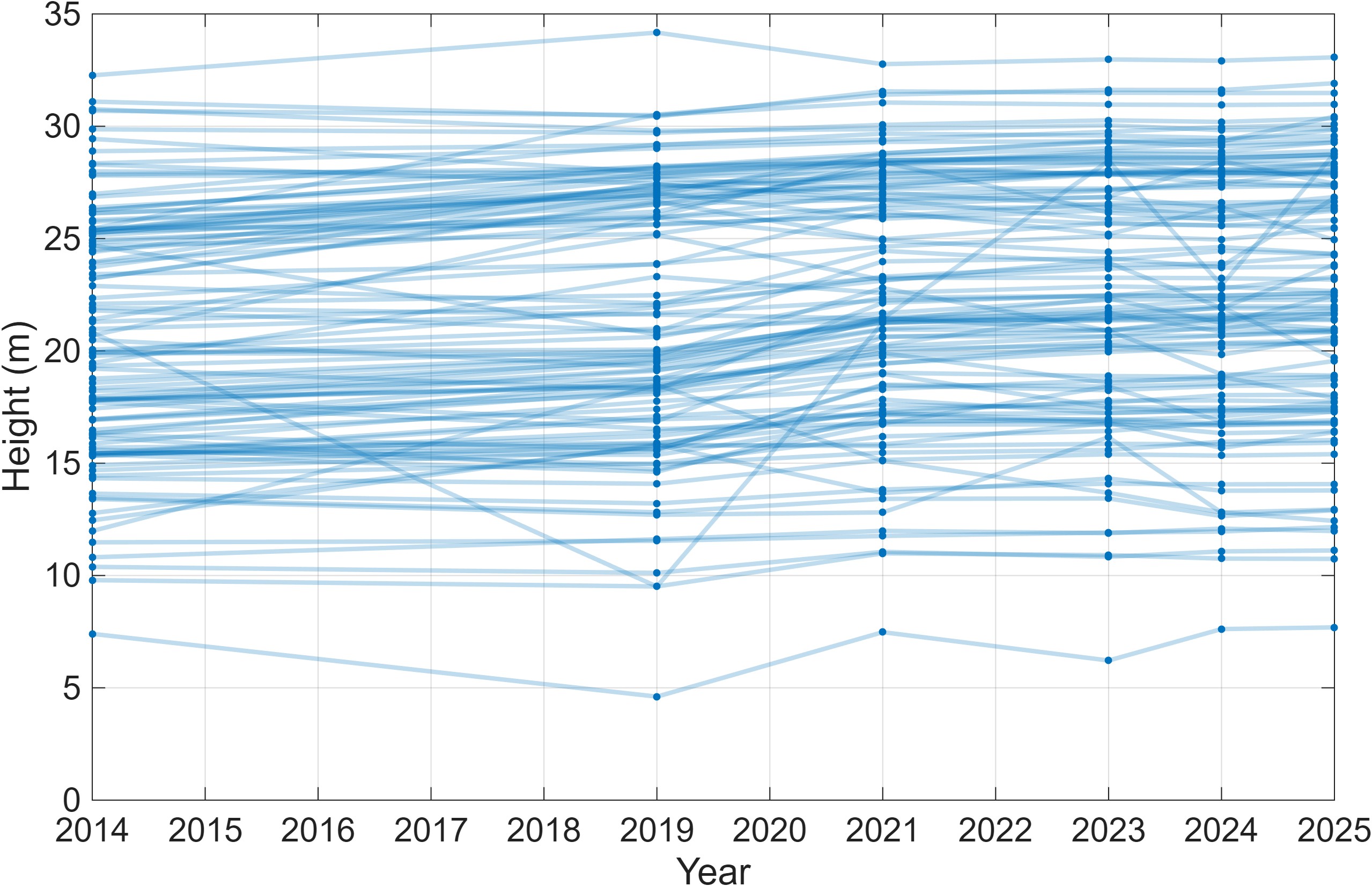}
    \end{subfigure} \\
    \begin{subfigure}{0.43\textwidth}
        \caption{\label{fig:time_series3}DBH, easy}
        \includegraphics[width=\linewidth]{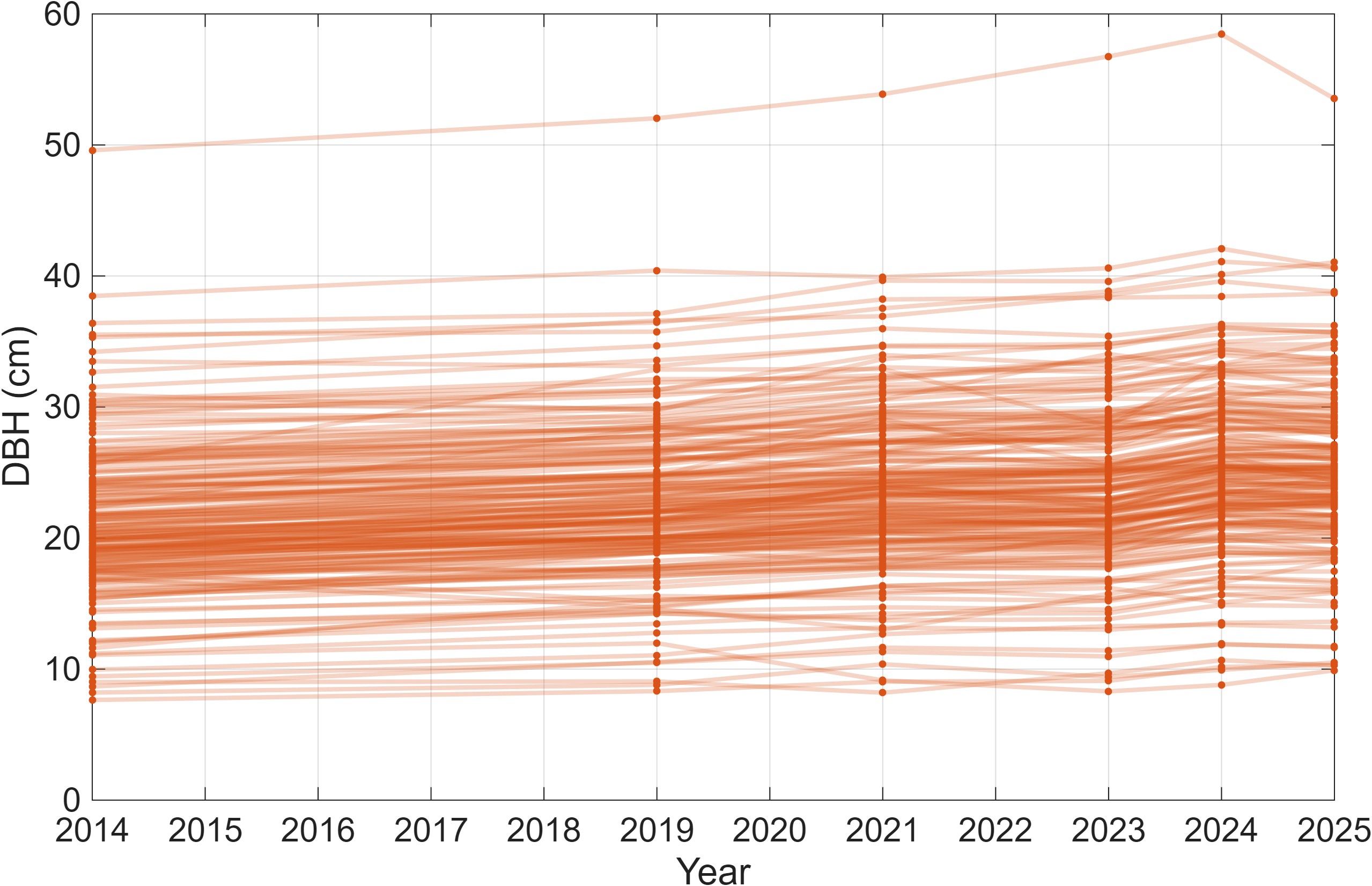}
    \end{subfigure}
    \begin{subfigure}{0.43\textwidth}
        \caption{\label{fig:time_series4} DBH, difficult}
        \includegraphics[width=\linewidth]{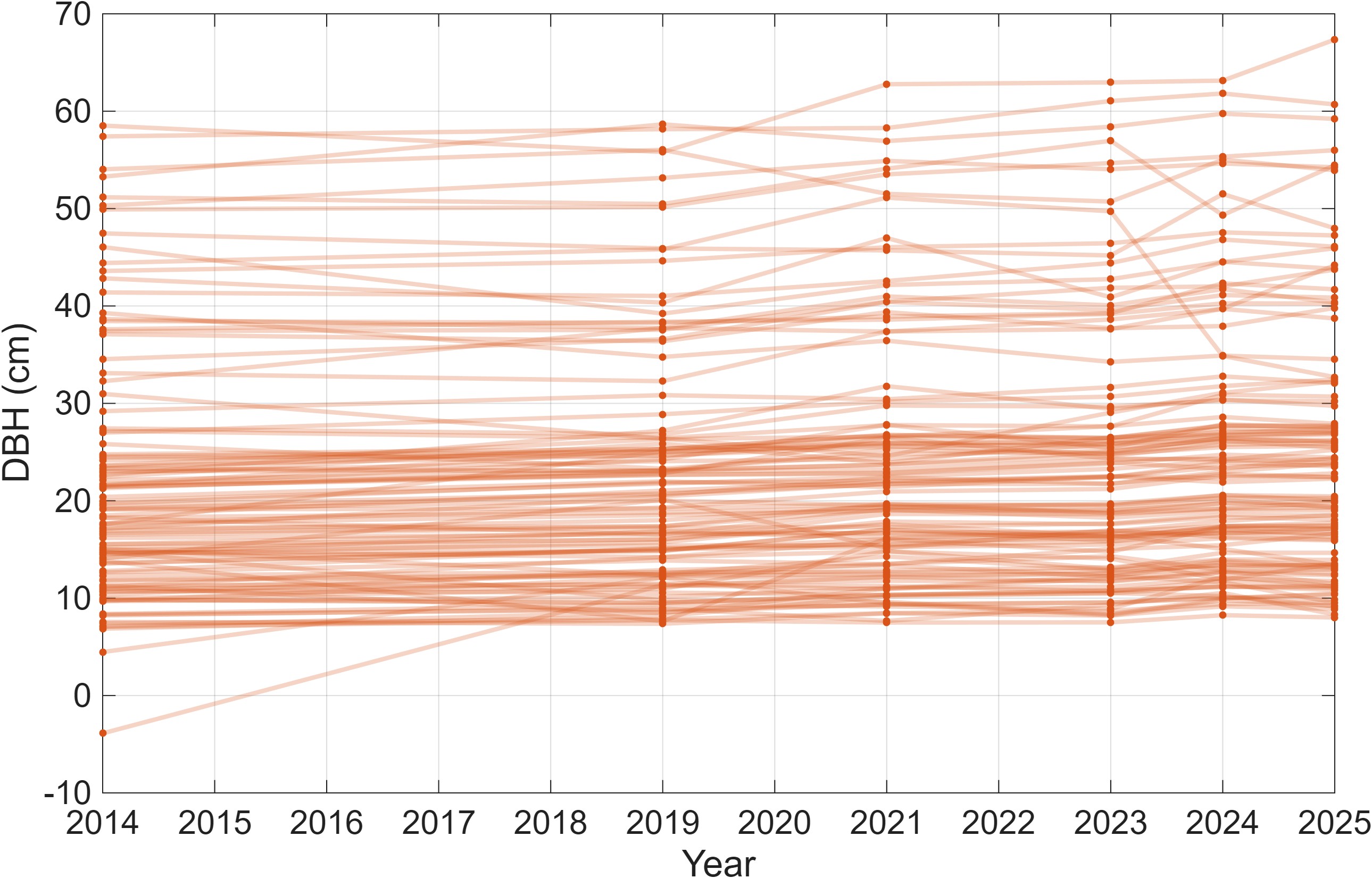}
    \end{subfigure} \\
    \begin{subfigure}{0.43\textwidth}
        \caption{\label{fig:time_series5} Stem volume, easy}
        \includegraphics[width=\linewidth]{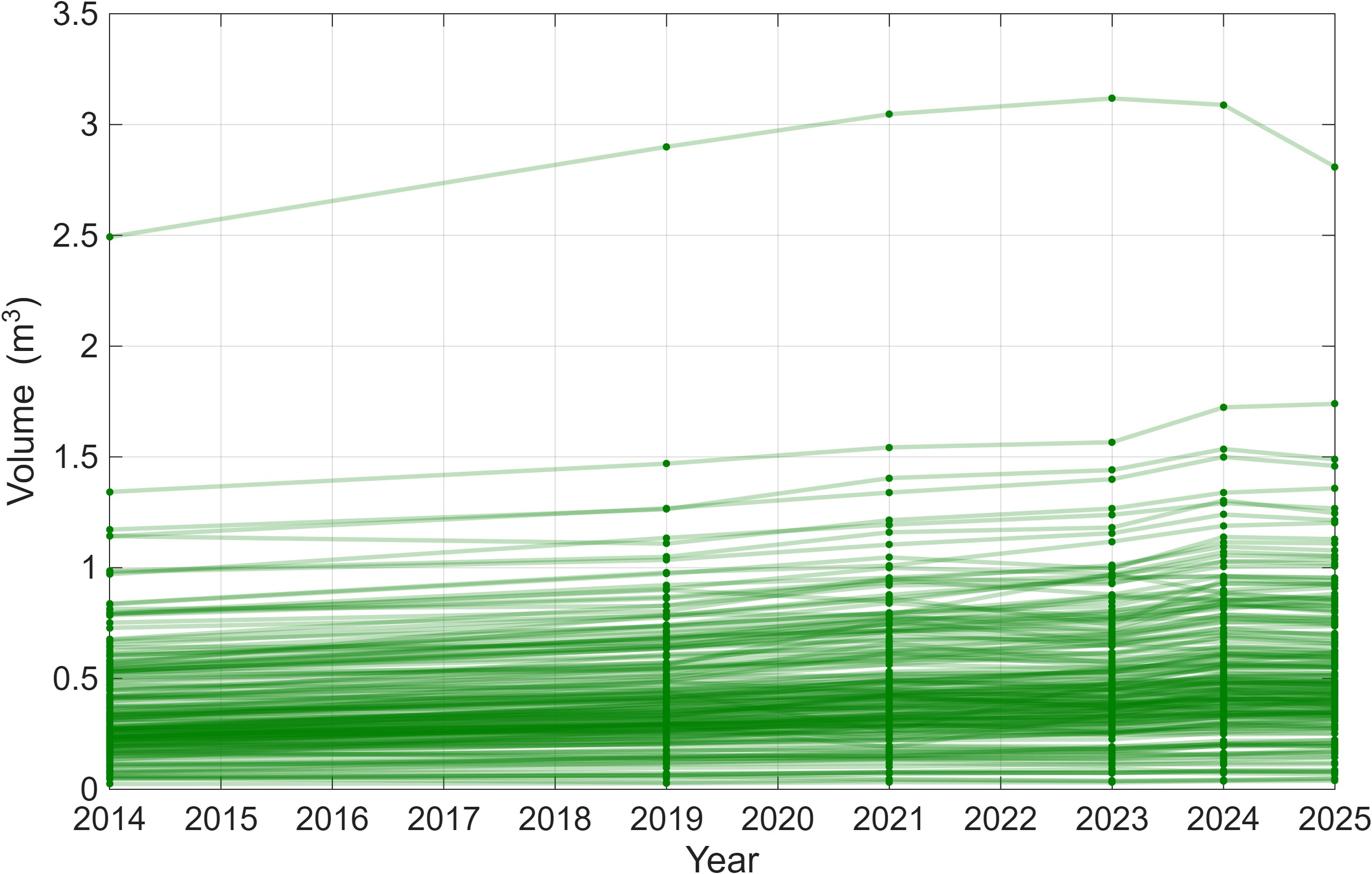}
    \end{subfigure}
    \begin{subfigure}{0.43\textwidth}
        \caption{\label{fig:time_series6} Stem volume, difficult}
        \includegraphics[width=\linewidth]{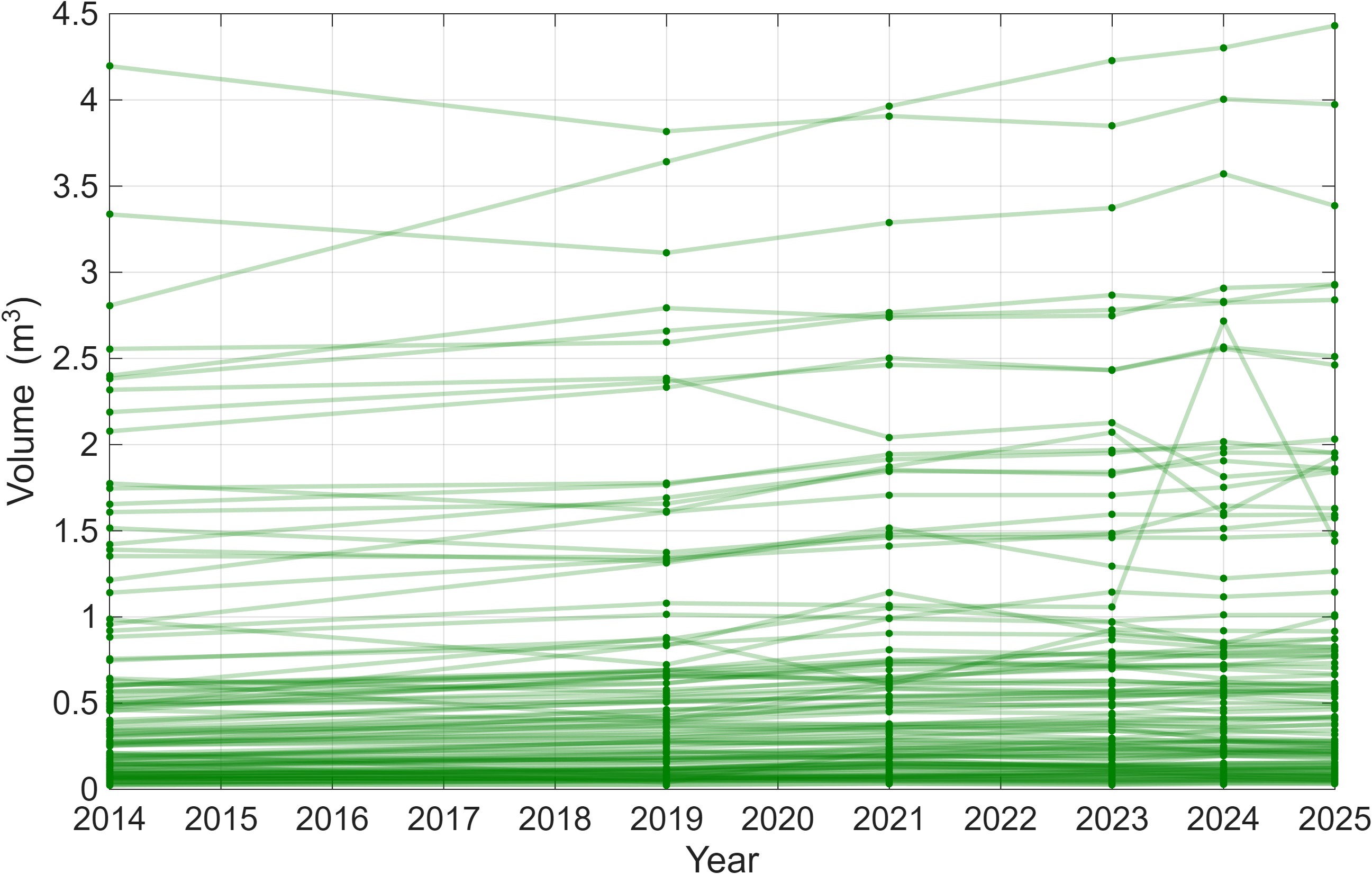}
    \end{subfigure} \\
    \caption{\label{fig:time_series} The measurement time series of height, DBH, and stem volume. Each line represents the growth of a tree attribute in the common set of trees and each dot is a measurement year. The measurements are not modeled but represent the output of the stem curve algorithm. Note that the years 2016--2018 and 2022 were scanned only using ALS.}
\end{figure*}

Segmentation transfer is a key step in the growth analysis, because tree-wise attributes are derived for each individual segment assumed to represent the same tree across all point clouds. In this study, we used the observed growth curves in Figs. \ref{fig:time_series1.0}--\ref{fig:time_series6} as a proxy indicator for the success of segment transferring. Errors in segment transferring would appear as discontinuities with random orientations in the trend lines. It should be noted that all errors are not caused by errors in segment transferring, but also, for example, by the limited ability of the point cloud data to represent complete tree objects or differences in the DTMs. Fig. \ref{fig:example_tree} shows examples of how the segment transferring appears across the point clouds.

A direct comparison of the preliminary segmentation and the segment transferring execution times can be made based on the number of points in the point clouds, despite differences in the computers used for executing the work and differences in areas covered by the point clouds. Segmenting the original 2020 MLS point clouds took approximately 1--1.5 h per plot. The maximum number of points in the 2020 MLS point clouds was approximately 130 million points, while the minimum number was 55 million points. In comparison, segment transfer to the larger MLS/TLS point clouds took only 20--200 seconds, when excluding the load times of the point clouds and the creation of DTMs, but including the kd-tree model creation time, the actual nearest neighbor search, and saving the segment labels. The point cloud sizes varied between 50--390 million points. These results demonstrate the computational savings gained by using preliminarily segmented point clouds for segment transferring.

In easy plots, tree height was estimated consistently over time (Fig. \ref{fig:time_series1.0}), resulting in smooth growth curves. A consistent increase in height is observed in 2017, which is related to the increased point density relative to the preceding and succeeding years (see Table \ref{tab:data_characteristics} and Fig. \ref{fig:example_tree}). Due to the increased density, the tops of trees were captured more accurately, leading to higher estimated tree heights. Further contributing to this sharper increase in heights is the longer time between the data collection compared to the preceding year, with the 2016 data collected in May/June and the 2017 data collected in September. The mean difference between the height captured using only ALS and ALS combined with MLS/TLS (Fig. \ref{fig:time_series1.1}) is at the millimeter scale in easy plots. Only in 2024 and 2025 do differences reach centimeter scale, corresponding to $-25$ cm and $-6$ cm, respectively. This indicates that ALS-based height estimation is generally accurate in these conditions, and that incorporating point clouds collected under-canopy (MLS/TLS) does not substantially improve the accuracy.

In difficult plots, height estimation is challenging using only ALS (Fig. \ref{fig:time_series2.0}). Short trees in particular exhibit increased variability in height estimates. The 2018 ALS data seem to severely underestimate tree height, despite these point clouds not having the lowest point density. This may be related to the low number of returns captured by the scanner, which was the lowest among the ALS scanners used in this study, and low flight altitude, as further outlined in Sec. \ref{sec:data_description}. Overall, the heights estimated from ALS-only data were lower than those derived from the combined ALS and MLS/TLS data. Per applicable years, the average differences were between $-68$ cm and $-4$ cm. Only in 2023 and 2025 do ALS-only estimates exceed the estimates from the combined data, with mean differences of 0.5 cm and 8 cm, respectively. The generally negative difference is likely explained by the improved under-canopy representation provided by the MLS and TLS sensors.

The DBH time series shows DBH measured consistently in easy plots (Fig. \ref{fig:time_series3}), as smooth growth trajectories are observed without abrupt changes in the growth patterns. The growth in difficult plots (Fig. \ref{fig:time_series4}) is also captured well, although occasional changes in the trend lines indicate increased DBH measurement uncertainty. Similarly, stem volume is measured more consistently in the easy plots compared to the difficult plots (Figs. \ref{fig:time_series5} and \ref{fig:time_series6}). The stem volume was computed as the average of two fitted equations (Eq. \eqref{eq:volume}), which introduces a smoothing effect. This makes the trend lines rather smooth also in the difficult plots.

We also compared the raw point cloud-derived estimates of height, DBH, and stem volume to their manual-only counterparts for applicable years, with the results presented in Table \ref{tab:error_raw_measurements}. It should be noted that manual measurements are not the most accurate measurements of height \citep{Wang2019}. In addition, the manual stem volume estimates are based on the Laasasenaho allometric equations \citep{Laasasenaho1982}, which represent a generalized tree in Finland and do not account for individual tree taper variation of the stem curve apart from incorporating height and DBH information. Manual DBH measurements, however, are considered the most accurate manual measurement. The error metrics between the manual-only measurements and point cloud-derived estimates give comparable results to other studies. According to the results in Table \ref{tab:error_raw_measurements}, both RMSE and bias are consistently lower in the easy plots than in difficult plots. There are no major differences in RMSE between years, whereas there is more variation in the bias values when calculated relative to other years.

\begin{table}[htb!]
\centering
\caption{\label{tab:error_raw_measurements}The RMSE and bias values between manual measurements and values derived from point clouds of individual tree height, DBH, and stem volume in easy and difficult plots. The error metrics are the baseline error metrics between the stem curve algorithm output and manual measurements. Height and stem volume were not manually measured in 2024. The height refers to ALS-only height.}
\begin{tabular*}{\linewidth}{@{\extracolsep\fill}lll@{}}
\toprule
\textbf{Easy plots}                     & \textbf{RMSE}     & \textbf{Bias} \\
\midrule
Height 2014 (m)                         & {0.89 (4.9\%)}    & {0.10 (0.5\%)} \\
Height 2019 (m)                         & {1.24 (6.6\%)}    & {$-0.90$ ($-4.8$\%)} \\
Height 2021 (m)                         & {0.83 (4.2\%)}    & {$-0.32$ ($-1.6$\%)} \\
Total (m)                               & {1.00 (5.2\%)}    & {$-0.37$ ($-1.9$\%)} \\
\hline
DBH 2014 (cm)                           & {1.6 (7.1\%)}    & {$-0.8$ ($-3.5$\%)} \\
DBH 2019 (cm)                           & {1.5 (6.4\%)}    & {$-0.6$ ($-2.5$\%)} \\
DBH 2021 (cm)                           & {1.4 (6.0\%)}    & {$-0.3$ ($-1.1$\%)} \\
DBH 2024 (cm)                           & {1.2 (5.0\%)}    & {0.3 (1.4\%)} \\
Total (cm)                              & {1.4 (6.1\%)}    & {$-0.3$ ($-1.4$\%)} \\
\hline
Stem volume 2014 (m\textsuperscript{3}) & {0.074 (19.2\%)}  & {$-0.003$ ($-0.9$\%)} \\
Stem volume 2019 (m\textsuperscript{3}) & {0.081 (17.6\%)}  & {$-0.004$ ($-0.8$\%)} \\
Stem volume 2021 (m\textsuperscript{3}) & {0.083 (16.6\%)}  & {0.003 (0.6\%)} \\
Total (m\textsuperscript{3})            & {0.071 (15.8\%)}  & {$-0.005$ ($-1.1$\%)} \\
\midrule
\textbf{Difficult plots}                & {\textbf{RMSE}}   & {\textbf{Bias}} \\
\midrule
Height 2014 (m)                         & {2.51 (11.8\%)}   & {$-0.79$ ($-3.7$\%)} \\
Height 2019 (m)                         & {2.80 (12.3\%)}   & {$-1.60$ ($-7.1$\%)} \\
Height 2021 (m)                         & {1.84 (8.0\%)}    & {$-0.65$ ($-2.8$\%)} \\
Total (m)                               & {2.42 (10.8\%)}   & {$-1.01$ ($-4.5$\%)} \\
\hline
DBH 2014 (cm)                           & {2.8 (12.2\%)}   & {$-0.9$ ($-4.0$\%)} \\
DBH 2019 (cm)                           & {3.2 (13.4\%)}   & {$-1.2$ ($-4.9$\%)} \\
DBH 2021 (cm)                           & {2.2 (9.2\%)}    & {$-0.5$ ($-2.0$\%)} \\
DBH 2024 (cm)                           & {3.2 (12.6\%)}   & {$-0.5$ ($-1.9$\%)} \\
Total (cm)                              & {2.9 (11.9\%)}   & {$-0.8$ ($-3.2$\%)} \\
\hline
Stem volume 2014 (m\textsuperscript{3}) & {0.166 (27.3\%)}  & {$-0.015$ ($-2.5$\%)} \\
Stem volume 2019 (m\textsuperscript{3}) & {0.212 (29.8\%)}  & {$-0.056$ ($-7.9$\%)} \\
Stem volume 2021 (m\textsuperscript{3}) & {0.202 (27.3\%)}  & {$-0.032$ ($-4.4$\%)} \\
Total (m\textsuperscript{3})            & {0.194 (28.3\%)}  & {$-0.035$ ($-5.0$\%)} \\
\bottomrule
\end{tabular*}
\end{table}

\subsection{Time series of error of the modeled tree attributes}
\label{sec:time_series_of_error_}
Based on the estimated attributes and the resulting growth trends, we discarded two trees from the subsequent analyses. One tree with a seemingly negative DBH in the year 2014 in one of the difficult plots (Fig. \ref{fig:time_series4}) was discarded from the analysis, as negative DBH cannot be measured in reality. Additionally, the tree with an estimated stem volume that jumps suddenly from approximately 1 m\textsuperscript{3} to 2.5 m\textsuperscript{3} in 2024 (Fig. \ref{fig:time_series6}) was deleted as the error inflated the error metrics and made analysis difficult. Both discarded trees are analyzed in Appendix \ref{app:errors_in_segmentation_and_stem_curve_extraction}. Hereafter, the results are presented after removing these two cases. The manual measurements of stem volume were obtained by using the height estimates from the combined ALS and MLS/TLS point clouds with the allometric volume equations that used manually measured DBH and species information.

\begin{figure*}[!htbp]
    \centering
    \begin{subfigure}{0.49\textwidth}
        \caption{\label{fig:one_time_scaling_dbh1} Starting year 2014 (forecasted)}
        \includegraphics[width=\textwidth]{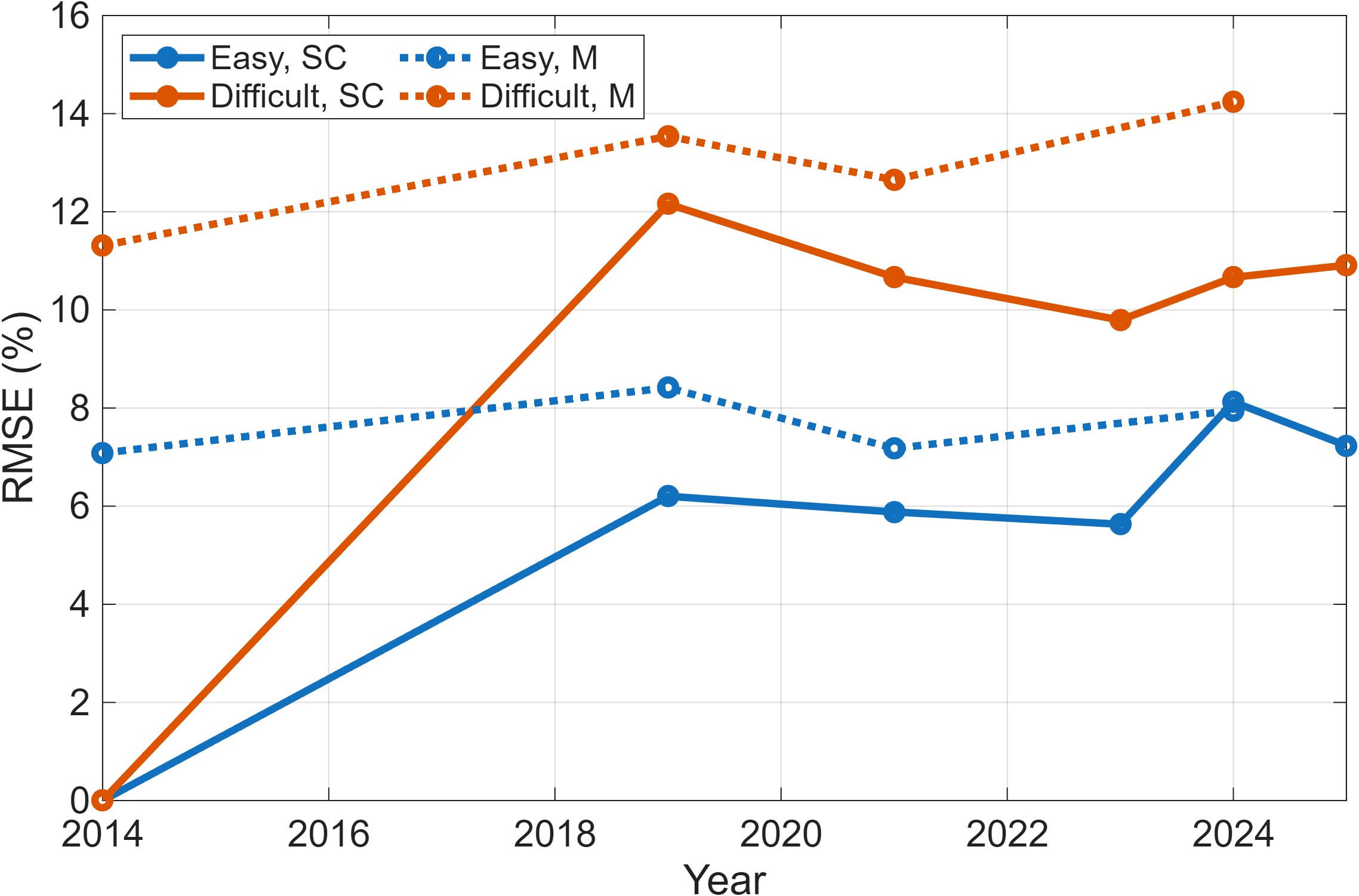}
    \end{subfigure}
    \begin{subfigure}{0.49\textwidth}
        \caption{\label{fig:one_time_scaling_dbh2} Starting year 2025 (hindcasted)}
        \includegraphics[width=\textwidth]{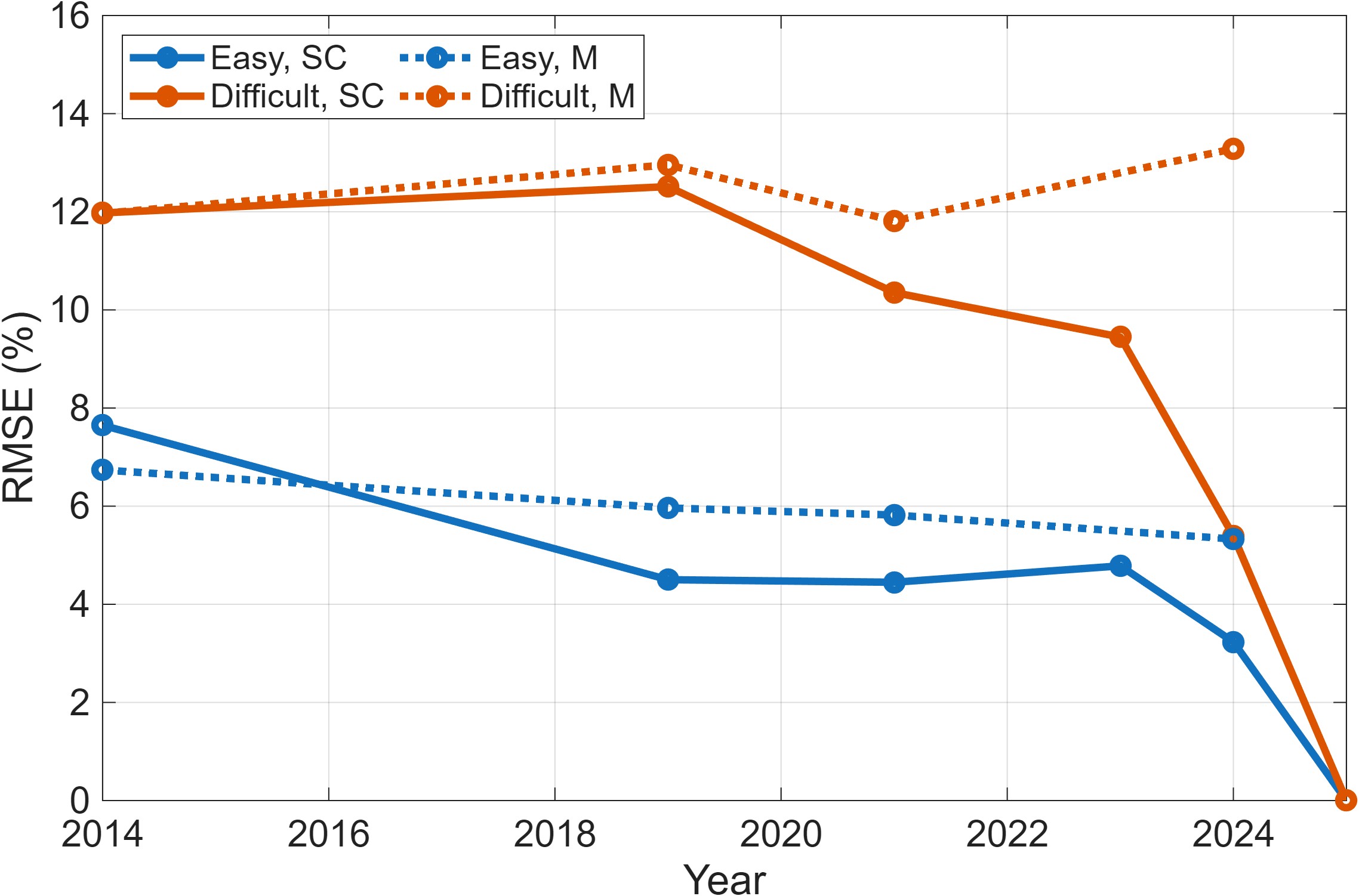}
    \end{subfigure} \\
    \begin{subfigure}{0.49\textwidth}
        \caption{\label{fig:one_time_scaling_dbh3} Starting year 2014 (forecasted)}
        \includegraphics[width=\textwidth]{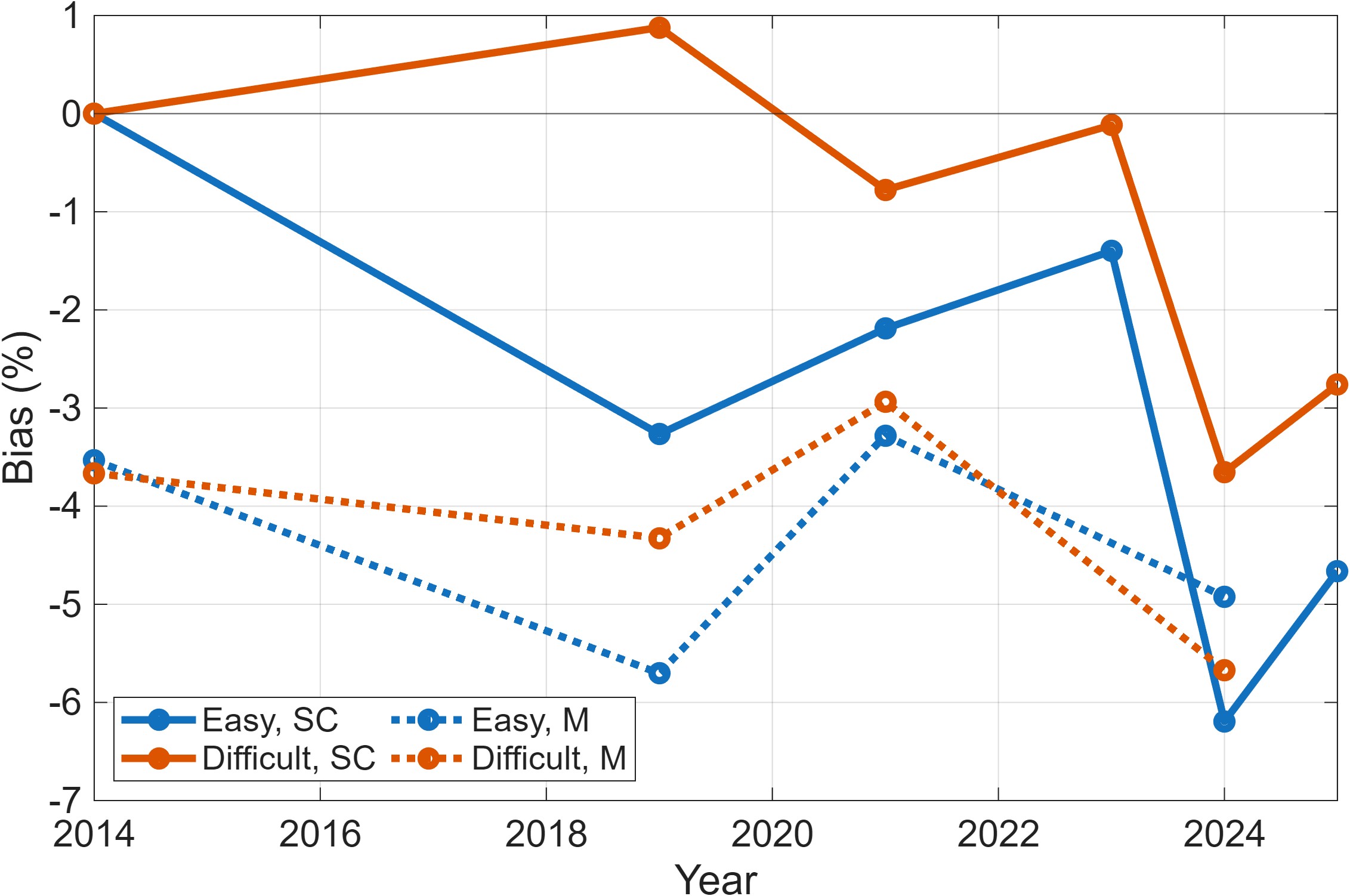}
    \end{subfigure}
    \begin{subfigure}{0.49\textwidth}
        \caption{\label{fig:one_time_scaling_dbh4} Starting year 2025 (hindcasted)}
        \includegraphics[width=\textwidth]{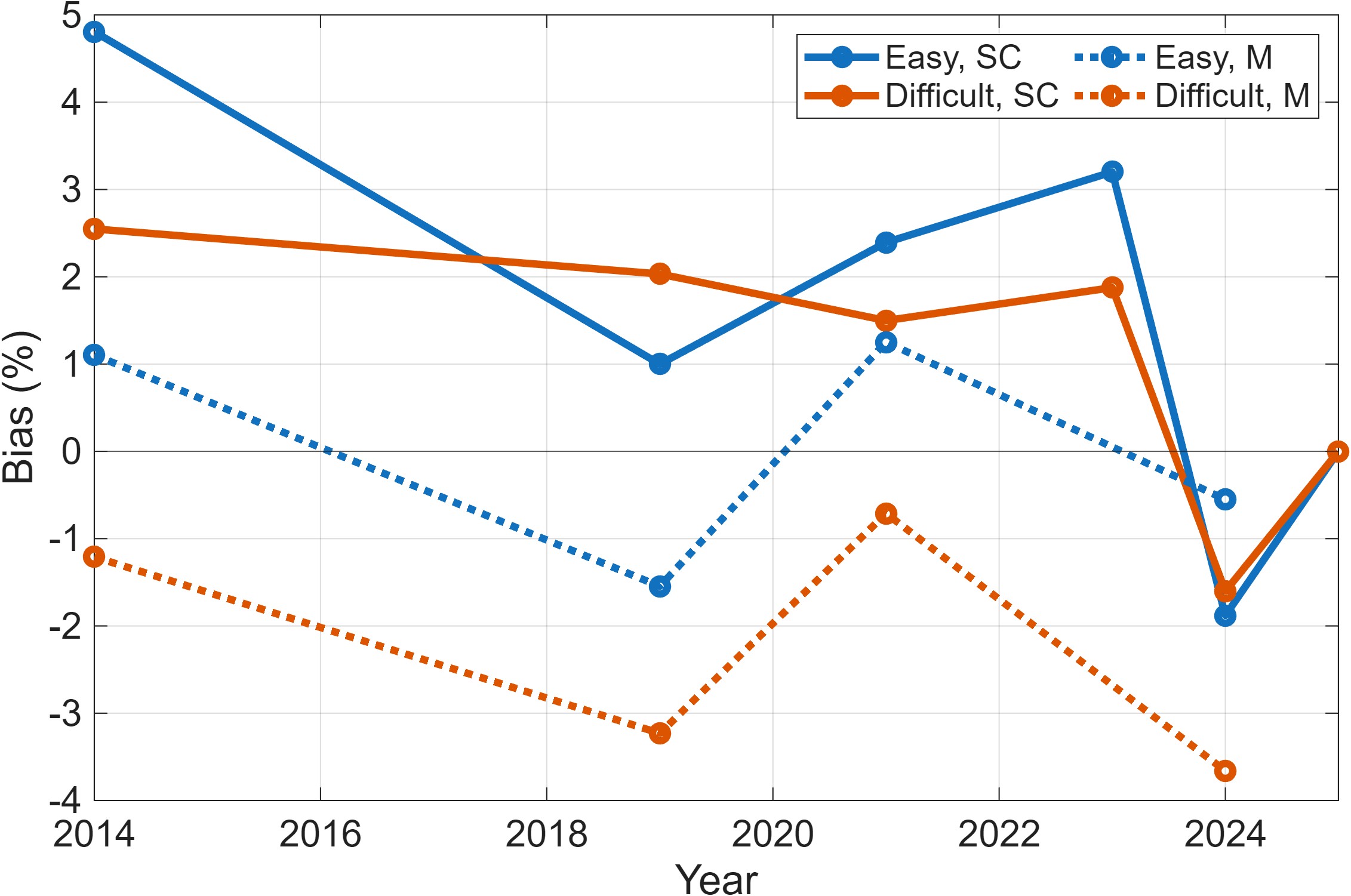}
    \end{subfigure}
    \caption{The growth in error of the forecasted (starting year 2014) and hindcasted (starting year 2025) results for DBH. The continuous line, labeled as SC, shows the difference between the direct stem curve measurements and the modeled DBH values. The dotted line, labeled as M, is the difference between the manual measurements and the modeled DBH values. The forecasted results are read from left to right, and the hindcasted results from right to left.}
    \label{fig:one_time_scaling_dbh}
\end{figure*}

Fig. \ref{fig:one_time_scaling_dbh} shows how the RMSE and bias of the modeled DBH values change as the time difference grows compared to both the direct measurements of stem curve and manual measurements. For the forecasted values, the RMSE indicates that most of the increase in error between the direct stem curve measurements and the modeled values occurs between 2014 and 2019, although details of the change cannot be examined in greater detail due to a lack of intermediate observations. In contrast, the RMSE of the hindcasted DBH values shows that most of the increase in RMSE occurs between 2025 and 2023 in easy plots, and 2025 and 2019 in difficult plots. The RMSE against the manually measured DBH indicates that the stem curve scaling approach is a rather robust method for change estimation, as the error does not substantially increase when either forecasting or hindcasting. The RMSE between the manual measurements and modeled DBH values is almost always larger than the RMSE between the direct stem curve measurements and modeled values.

The bias of the forecasted and hindcasted DBH values exhibit more inconsistent behavior than the RMSE values. In the forecasted results, the bias between the modeled and direct measurements decreases over time. In contrast, the hindcasted bias shows an increasing trend (2025\textrightarrow2014). The trend between the manual measurements and the modeled values is much smaller. Additionally, it appears that the modeled values in the easy plots are often more biased than the values in the difficult plots, with some of the bias values being rather large in magnitude.

\begin{figure*}[!htbp]
   \centering
   \begin{subfigure}{0.49\textwidth}
        \caption{\label{one_time_scaling_stemvolume1} Starting year 2014 (forecasted)}
        \includegraphics[width=\linewidth]{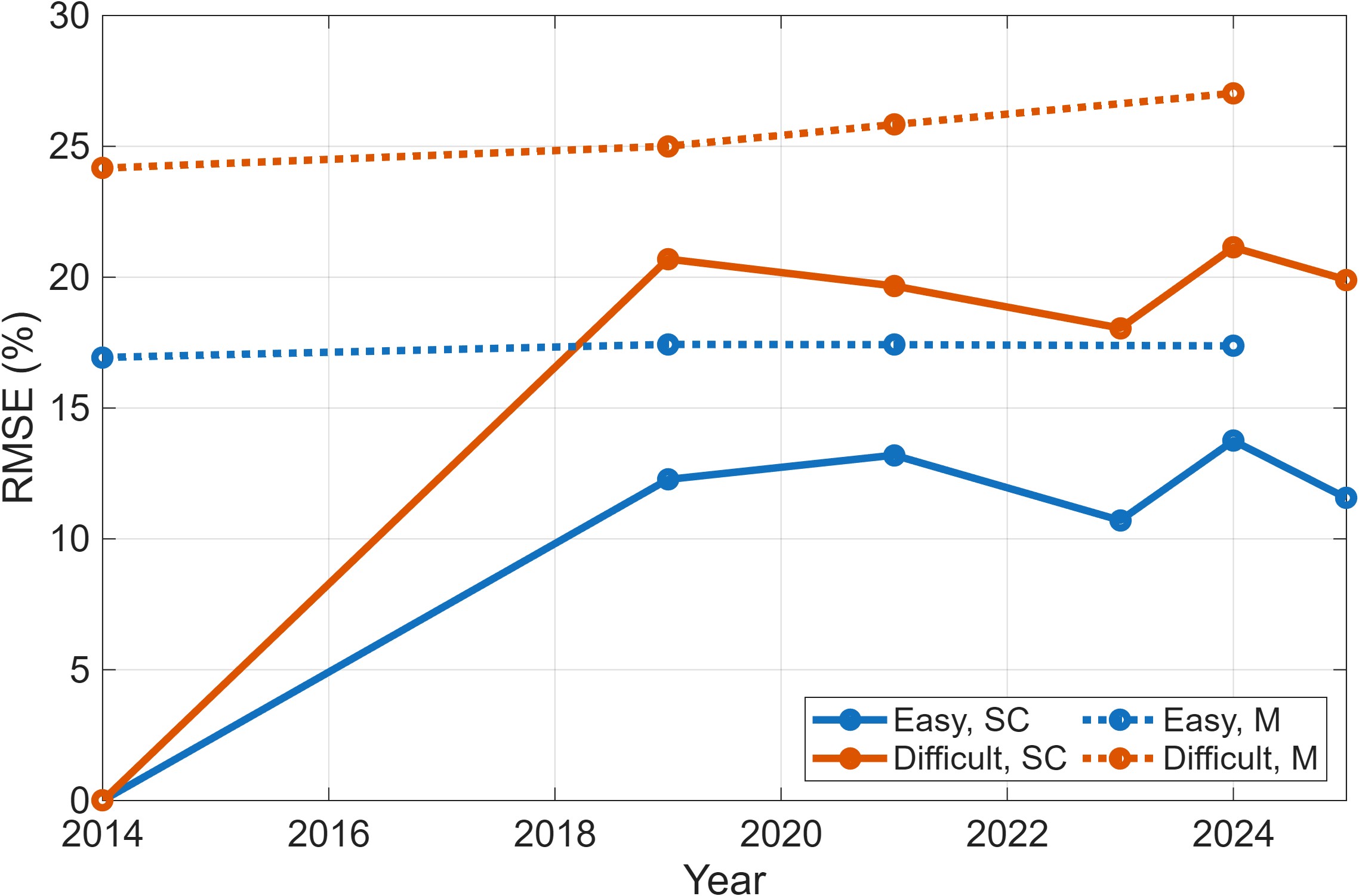}
   \end{subfigure} 
    \begin{subfigure}{0.49\textwidth}
        \caption{\label{one_time_scaling_stemvolume2} Starting year 2025 (hindcasted)}
        \includegraphics[width=\linewidth]{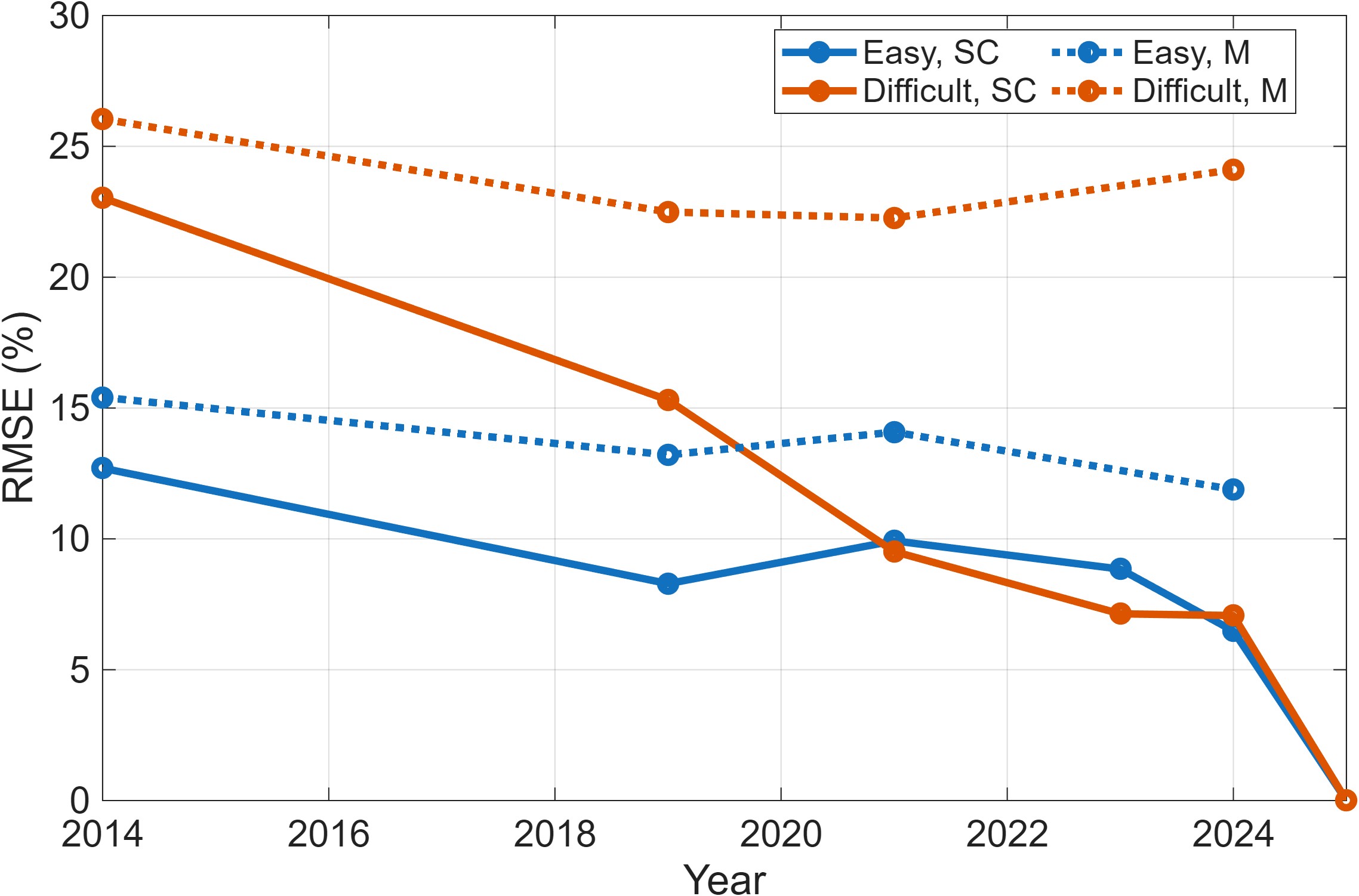}
   \end{subfigure}
   \begin{subfigure}{0.49\textwidth}
        \caption{\label{one_time_scaling_stemvolume3} Starting year 2014 (forecasted)}
        \includegraphics[width=\linewidth]{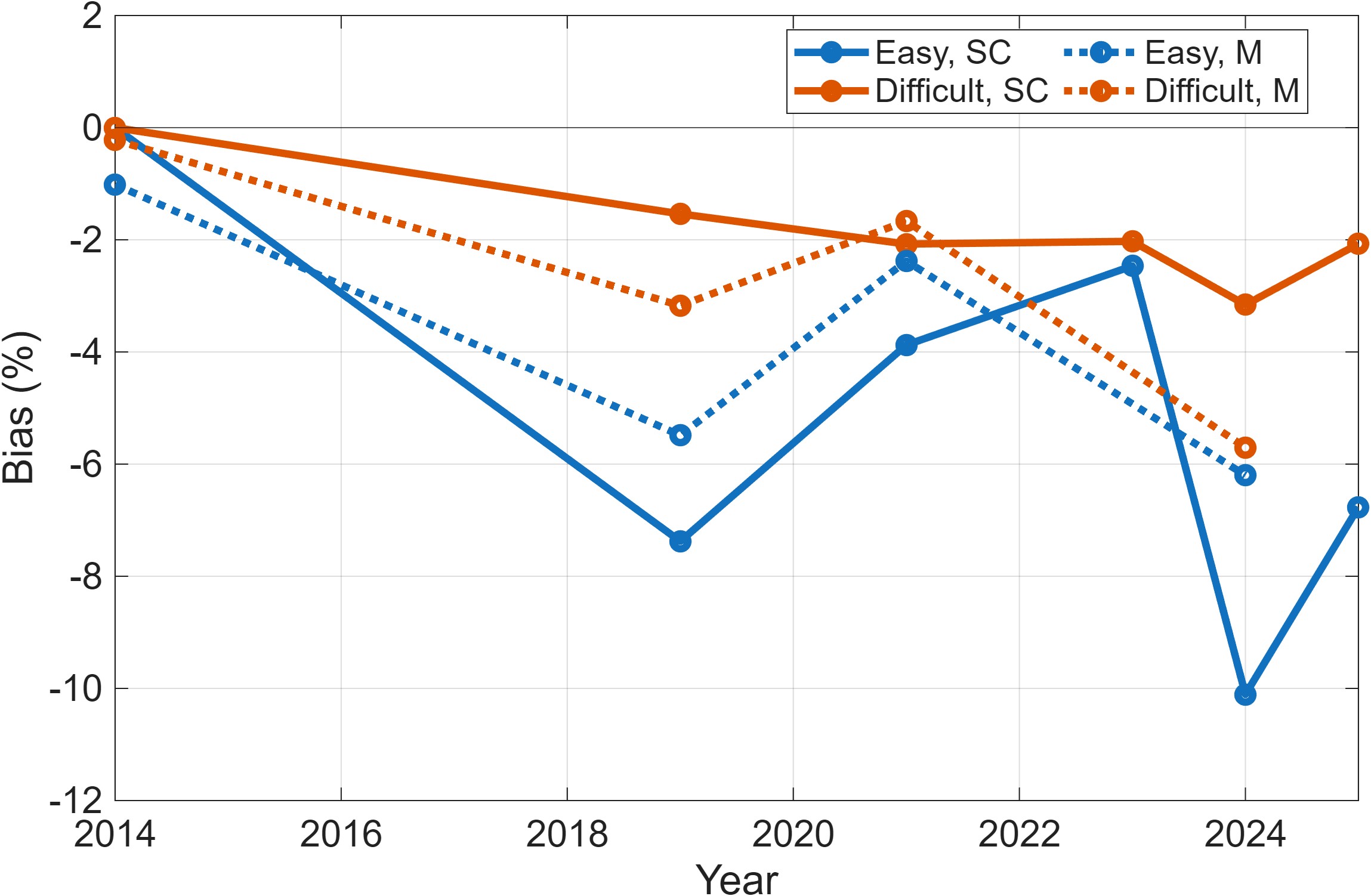}
   \end{subfigure} 
    \begin{subfigure}{0.49\textwidth}
        \caption{\label{one_time_scaling_stemvolume4} Starting year 2025 (hindcasted)}
        \includegraphics[width=\linewidth]{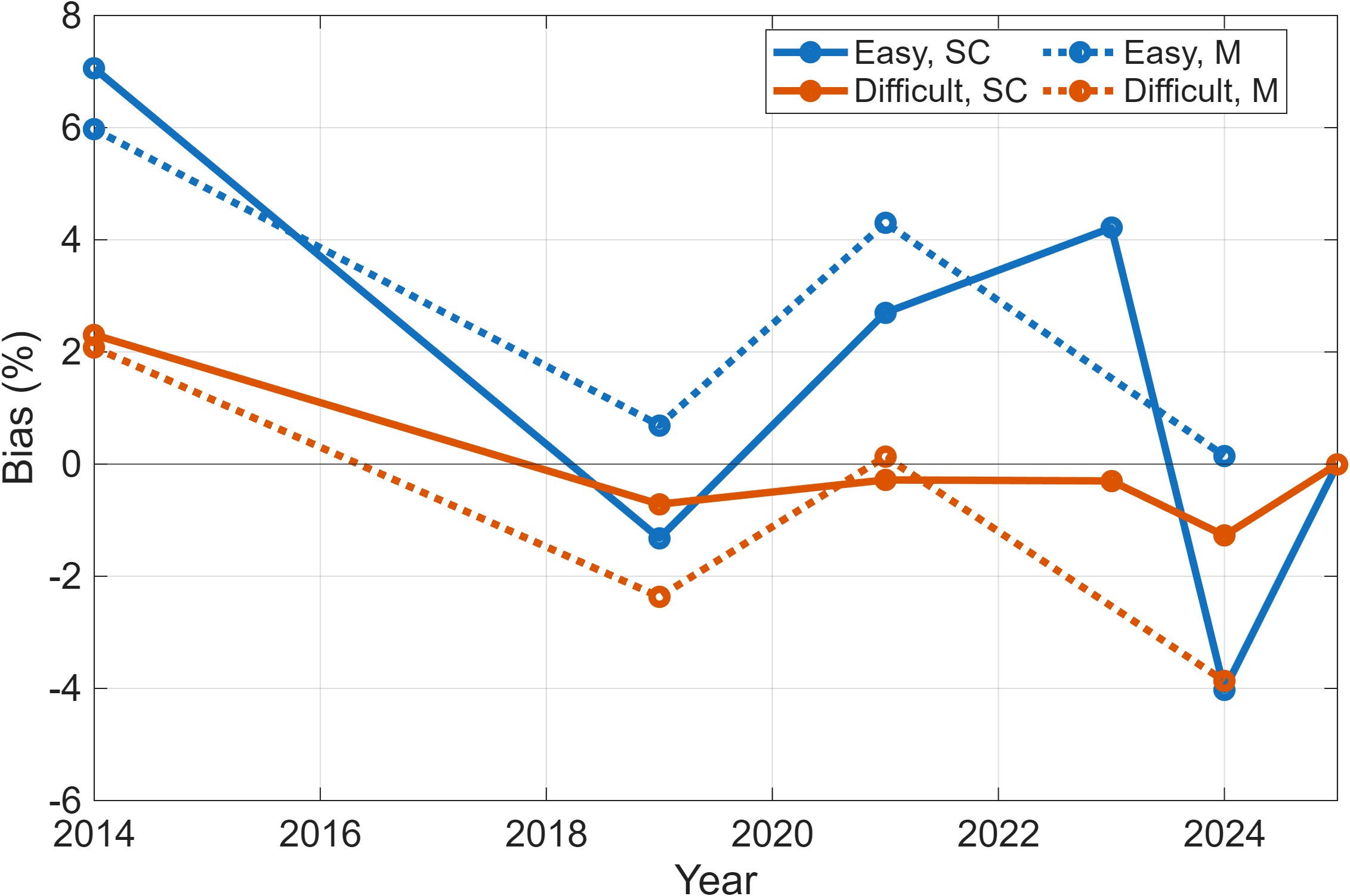}
   \end{subfigure}
   \caption{The growth in error of the forecasted (starting year 2014) and hindcasted (starting year 2025) results for stem volume. The continuous line, labeled as SC, shows the difference between the direct stem curve measurements and the modeled stem volume values. The dotted line, labeled as M, is the difference between the manual measurements and the modeled stem volume values. The forecasted results are read from left to right, and the hindcasted results from right to left. }
   \label{fig:one_time_scaling_stemvolume}
\end{figure*}

The growth in RMSE and bias of the modeled stem volume values is shown in Fig. \ref{fig:one_time_scaling_stemvolume}. For forecasted stem volume, the RMSE values show that the error stabilizes after five years between the modeled values and the direct stem curve measurements. In the hindcasted results, the RMSE in the easy plots stabilizes by 2023, but begins to slowly increase again in 2019. The RMSE of the difficult plots increases continuously, and grows linearly after the initial increase.

The stem volume bias exhibits a negative trend in the forecasted results, both when compared with manual measurements and with direct stem curve estimates. Conversely, bias grows in the hindcasted values. Importantly, some years show very large bias, with these values occurring in the easy plots. For example, the forecasted bias in 2019 is between $-5\%$ and $-7\%$ and in 2024, between $-6\%$ and $-10\%$. Overall, the forecasted values appear to be more biased than the hindcasted values.

The height measurements in 2019 originate from the point cloud acquired during the national laser scanning campaign. Here, we compare the 2019 results with those of the surrounding years to assess the suitability of height measurements derived from sparse, large-area ALS point clouds for scaling-based change detection. Because 2019 lies in the middle of the 2014--2025 time series, its errors are not expected to be the smallest. However, if the national laser scanning point clouds are suitable for this application, the error metrics should remain comparable to those of the surrounding years and should not exhibit a pronounced spike.

In the easy plots, the forecasted and hindcasted DBH RMSE in 2019 does not deviate from the surrounding years when compared with the direct stem curve measurements (Figs. \ref{fig:one_time_scaling_dbh1} and \ref{fig:one_time_scaling_dbh2}). In difficult plots, the RMSE is the largest in 2019, although only by a 2 \%-point margin to the smallest (non-zero) RMSE. When compared with the manual measurements, the 2019 results do not produce particularly erroneous values. In stem volume forecasting and hindcasting, the RMSE in 2019 is neither better nor worse than that of the surrounding years when compared to both manual measurements and direct stem curve measurements. The bias values in 2019 appear to be dependent on the scanner used in the starting year. Forecasted DBH and stem volume are generally underestimated, even by a considerable margin, whereas the bias of the hindcasted DBH and stem volume estimates is not particularly large in 2019. In fact, the bias of the hindcasted stem volume estimates in 2019 is among the smallest observed across the evaluated years.

\subsection{Modeling stem taper change}
\begin{figure*}[!htbp]
    \centering
    \begin{subfigure}{0.49\textwidth}
        \caption{\label{fig:difference_in_error1} RMSE}
        \includegraphics[width=\linewidth]{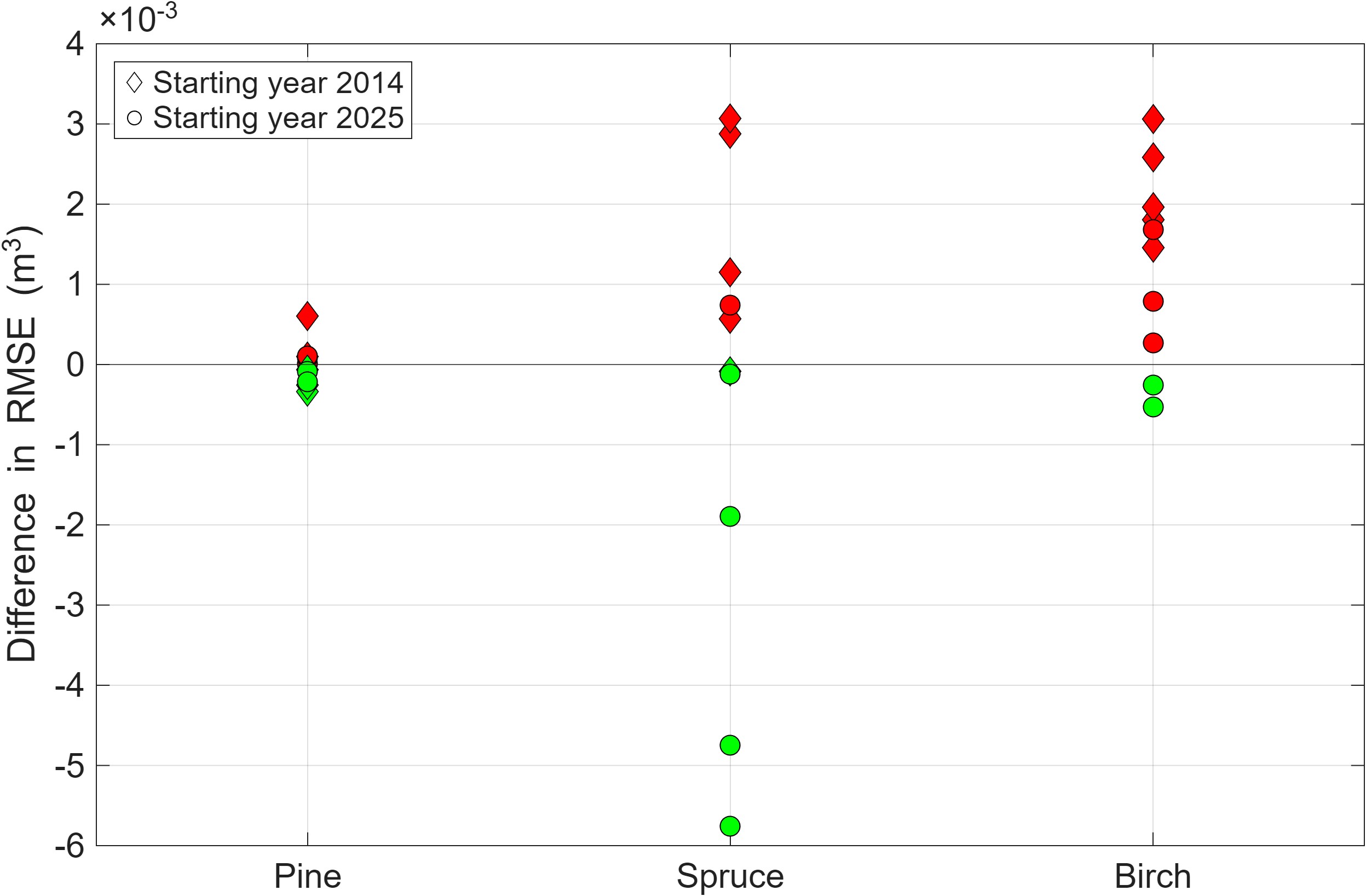}
    \end{subfigure}
    \begin{subfigure}{0.49\textwidth}
        \caption{\label{fig:difference_in_error2} Bias}
        \includegraphics[width=\linewidth]{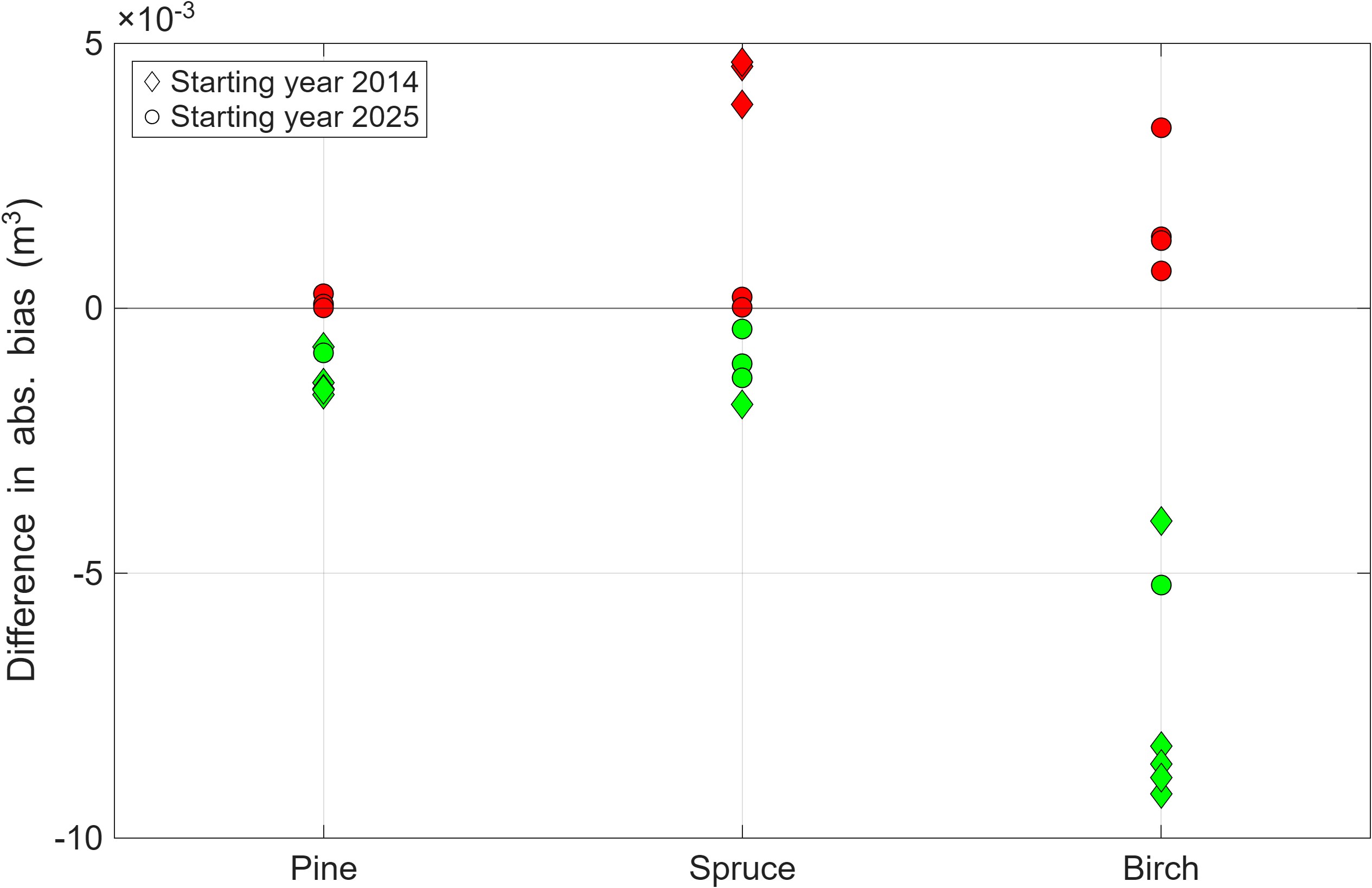}     
    \end{subfigure}
    \caption{The difference in \ref{fig:difference_in_error1}: RMSE and \ref{fig:difference_in_error2}: absolute bias for models that either include or exclude the stem taper change information when scaling the stem proportions in time. Two starting points are shown: 2014 with diamond markers and 2025 with round markers. Each marker represents a result from a year that is not the starting year (which has a zero error), but the years are not differentiated otherwise. Thus, there are five markers per species per starting year.}
    \label{fig:difference_in_error}
\end{figure*}

The effect of including stem taper information in the model was studied by examining the difference in the errors produced by the two modeling approaches. For RMSE, the difference is calculated as RMSE\textsubscript{\textit{f}} $-$ RMSE\textsubscript{\textit{b}}, where $f$ refers to results obtained with taper change information (Eq. \eqref{eq: scaled_diameter_with_form}) and $b$ to results obtained without taper change information (Eq. \eqref{eq: scaled_diameter_without_form}). Both RMSE\textsubscript{\textit{f}} and RMSE\textsubscript{\textit{b}}, and the associated bias values, are calculated between the modeled values and the direct tree stem measurements. Because bias is a signed value, it is more reasonable to study the difference in absolute biases. Hence, the results are calculated as |bias\textsubscript{\textit{f}}| $-$ |bias\textsubscript{\textit{b}}|. Negative results indicate smaller error obtained by including the stem taper information, and positive results indicate the opposite. The results are shown in Fig. \ref{fig:difference_in_error}.

The main observation is that the differences both in RMSE and bias between the results obtained without stem taper change information and with the information are minuscule. The mean stem curve-based volume in 2014 is 0.460 m\textsuperscript{3}. The maximum improvements are approximately 0.006 m\textsuperscript{3} in RMSE and 0.01 m\textsuperscript{3} in bias. The improvements are thus only approximately 1--2\% at most. Furthermore, some results are slightly worse, such as the RMSE of birch, or basically unaffected, such as pine.

\subsection{Estimation of DBH and stem volume growth}
\label{sec:growth_results}

\begin{figure*}[htbp!] 
	\centering
	\begin{subfigure}[t!]{0.49\textwidth}
        \caption{\label{fig:dbh_change_scatter_1}DBH change 2014--2024 in easy plots.}
        \includegraphics[width = \textwidth]{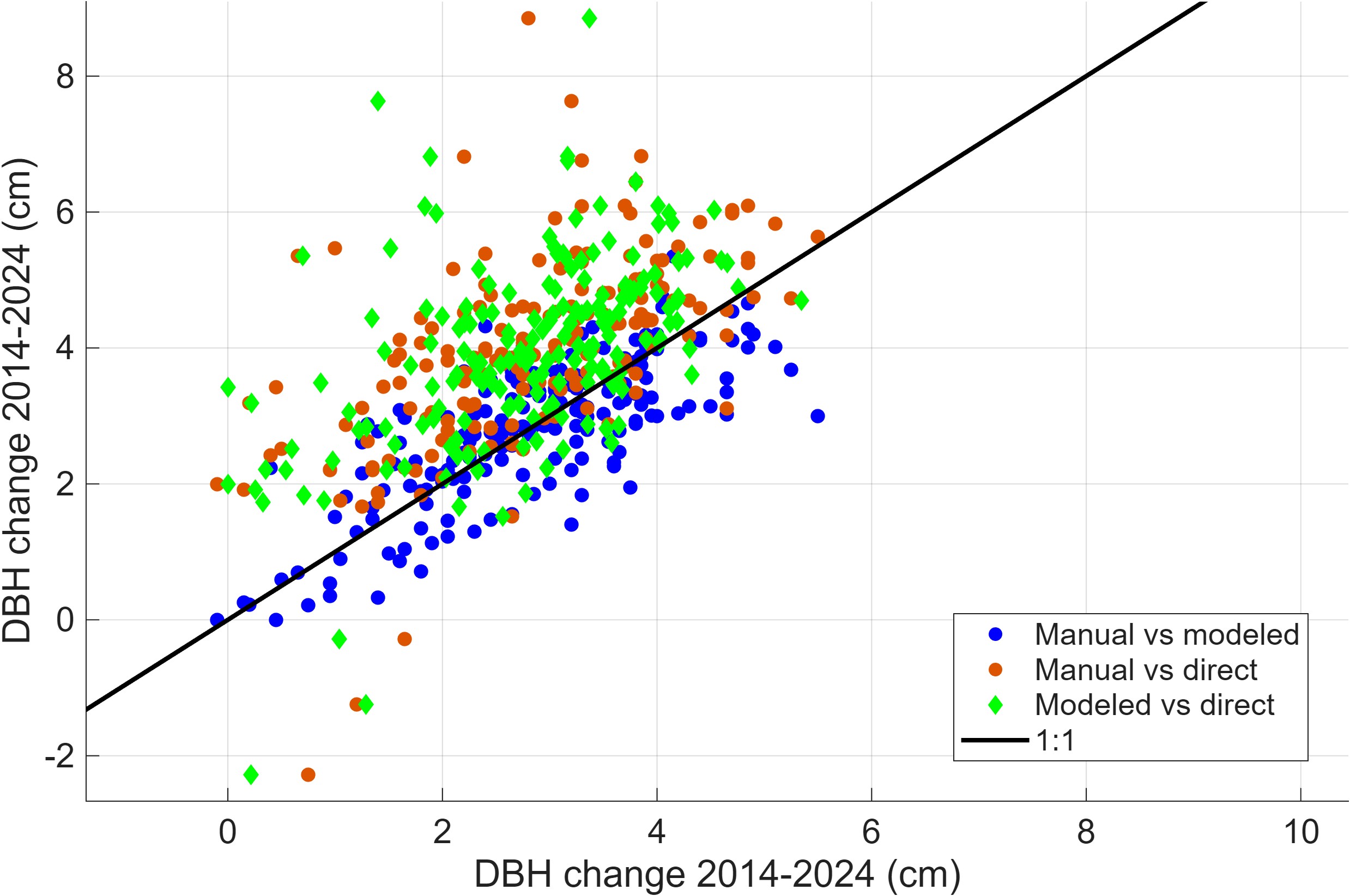}
    \end{subfigure}
    \begin{subfigure}[t!]{0.49\textwidth}
        \caption{\label{fig:dbh_change_scatter_2}DBH change 2014--2024 in difficult plots.}
        \includegraphics[width = \textwidth]{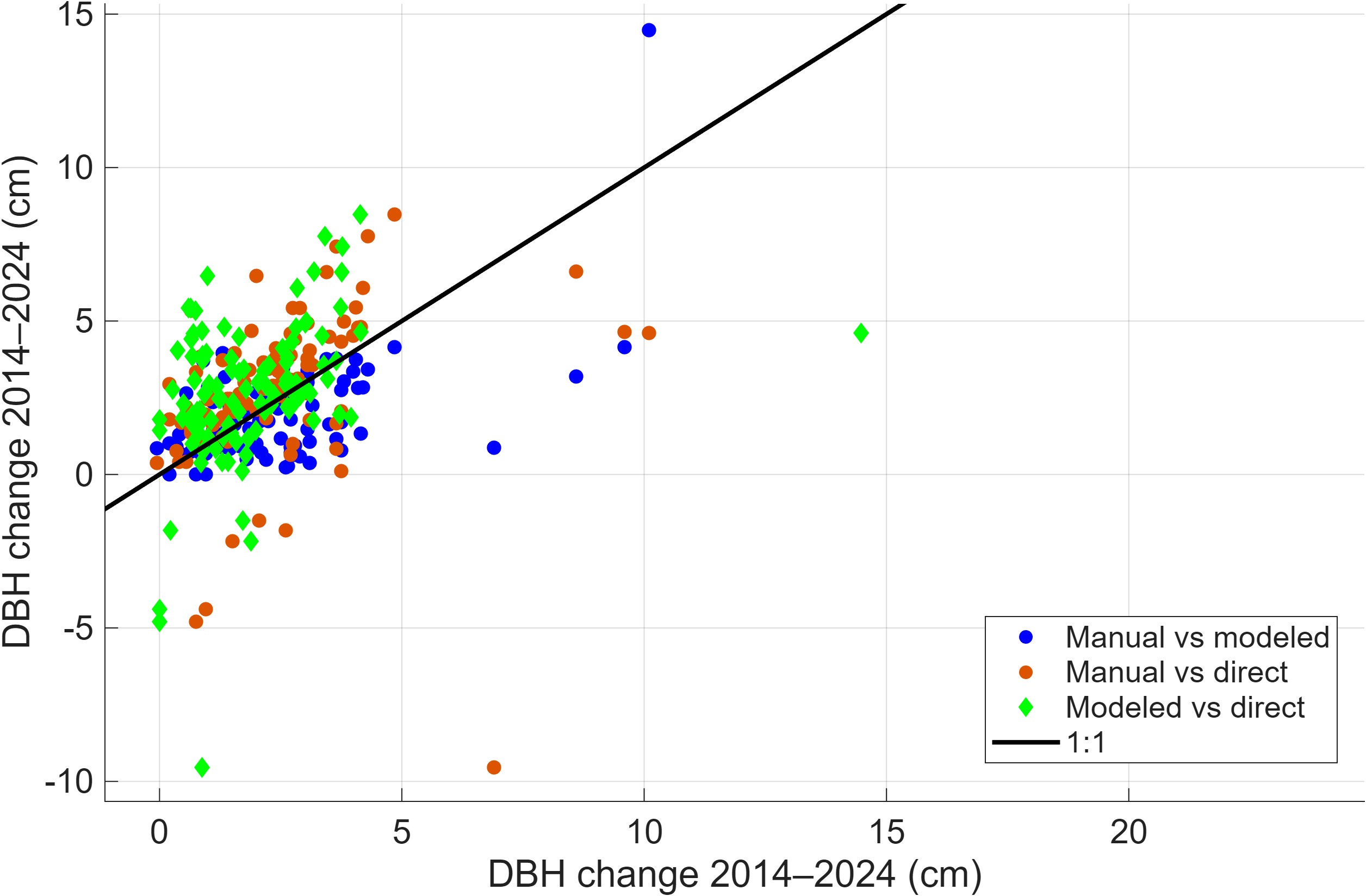} 
    \end{subfigure} \\
    \begin{subfigure}[t!]{0.49\textwidth}
        \caption{\label{fig:volume_change_scatter_1}Volume change 2014--2024 in easy plots.}
        \includegraphics[width = \textwidth]{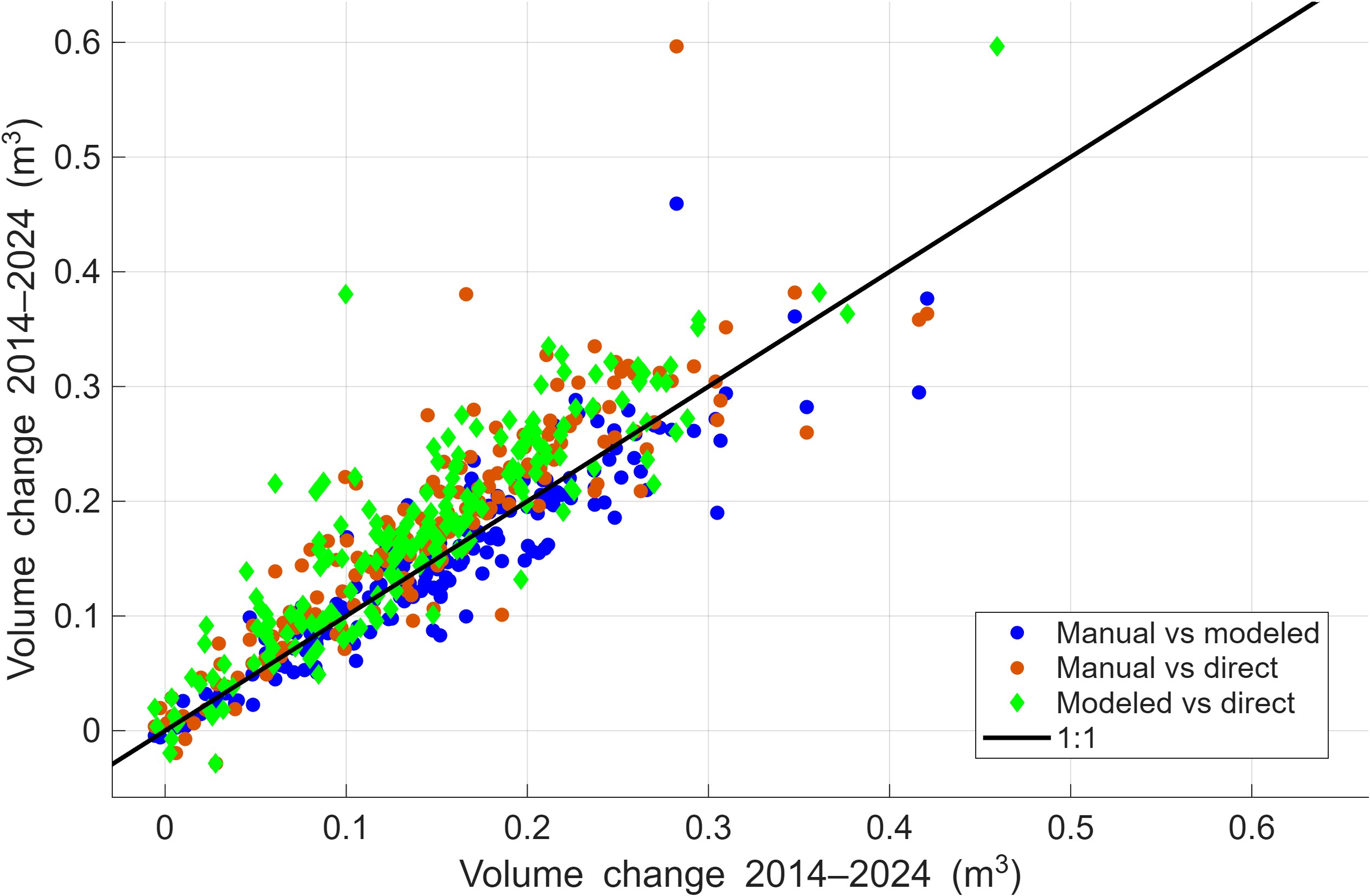}
    \end{subfigure}
    \begin{subfigure}[t!]{0.49\textwidth}
        \caption{\label{fig:volume_change_scatter_2}Volume change 2014--2024 in difficult plots.}
        \includegraphics[width = \textwidth]{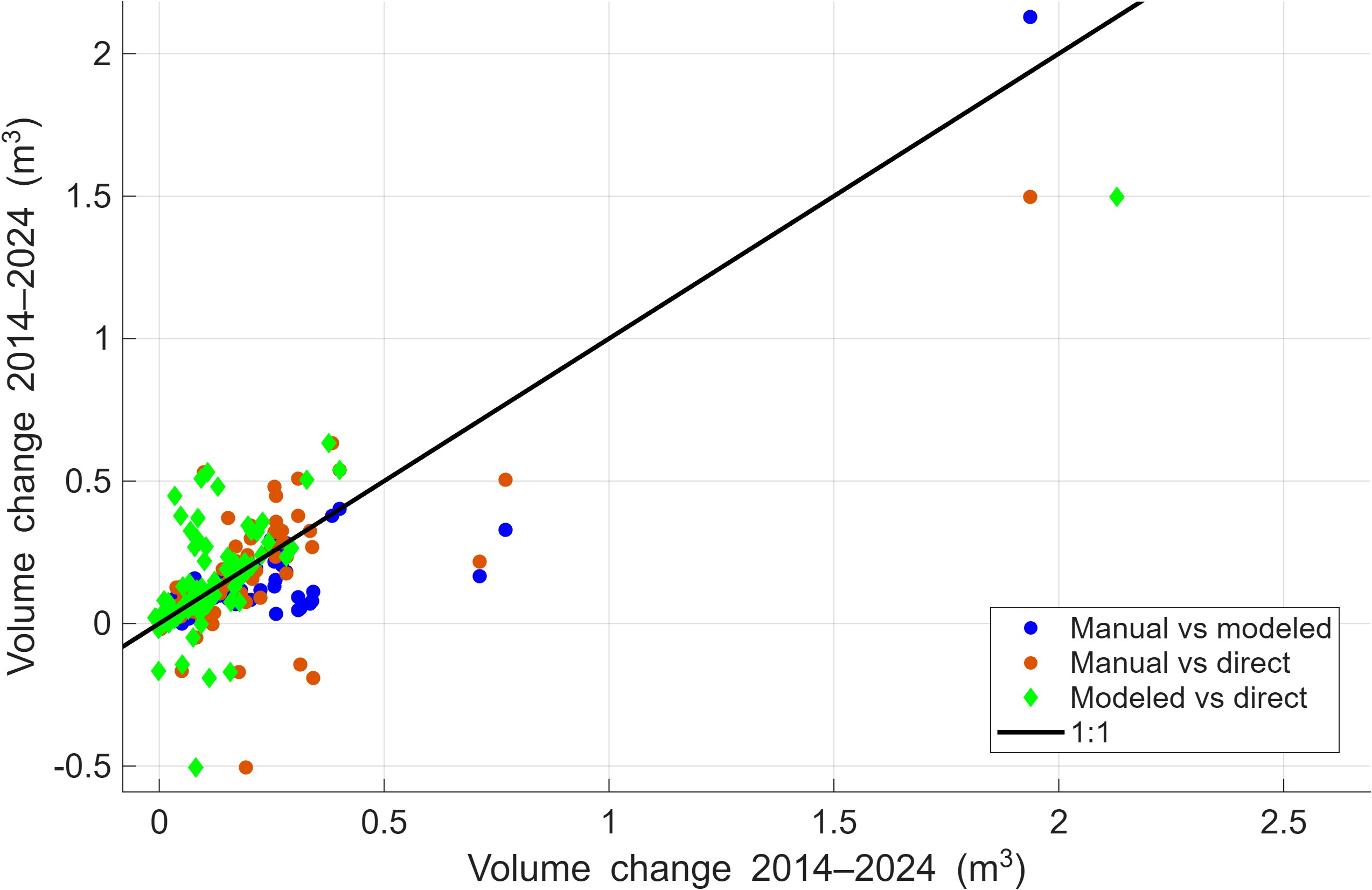} 
    \end{subfigure}
    \caption{Scatterplots illustrating the relationships of the estimated DBH change between 2014--2024 using the manual, model-based, and direct approaches for \ref{fig:dbh_change_scatter_1}: easy plots and \ref{fig:dbh_change_scatter_2}: difficult plots. \ref{fig:volume_change_scatter_1} and \ref{fig:volume_change_scatter_2}: The same for stem volume.}\label{fig:change_scatter}
\end{figure*}

DBH and stem volume growth estimation was compared between the direct approach, the model-based approach, and the manual approach. Fig. \ref{fig:change_scatter} presents scatter plots illustrating the relationships between these approaches for estimating 2014--2024 DBH and volume change. The scatter plots indicate that, for both attributes, the model-based and manual approaches produced the most similar growth estimates in both easy and difficult plots. In contrast, comparisons involving the direct approach (direct-manual and model-based-direct) show greater dispersion, indicating lower agreement. The direct approach also produced more negative growth estimates. For DBH change, negative growth was observed for one tree in both easy and difficult plots using the manual approach, compared to three and six trees, respectively, using the direct approach. Meanwhile, the model-based approach did not obtain any negative growth estimates due to its inherent constraints.

Similarly for volume growth, the manual approach measured two trees with negative growth in easy plots and none in difficult plots, while the direct approach obtained three and eight negative estimates, respectively. The model-based approach produced two negative estimates in easy test sites and four in difficult sites. Notably, the magnitude of negative growth estimates was substantially larger for the direct approach compared to the other methods.

Tables \ref{tab:dbh_growth_accuracies} and \ref{tab:vol_growth_accuracies} present error metrics for both 10-year (2014--2024) and 5-year (2019--2024) growth, calculated as differences between the three approaches. For DBH change, the model-based approach exhibited the highest agreement with the manual estimates across both plot difficulties and time spans. This is reflected in consistently lower RMSE and MAE values compared to the other approaches. For 10-year growth, RMSE values for the model-based-manual comparison ranged from 0.7 cm (26\%) in easy plots to 1.6 cm (67\%) in difficult plots. Comparatively, comparisons involving the direct approach obtained RMSEs of 1.6--1.7 cm (57--62\%) in easy plots and 2.5 cm (106--129\%) in difficult plots. Consistent with these results, higher $R^2$ values were generally observed for the model-based-manual comparison, at 0.59 and 0.23 in easy plots for 10- and 5-year growth, and 0.32 for 10-year growth in difficult plots. All comparisons found $R^2$ below 0.1 for 5-year growth in difficult plots. Meanwhile, the direct-manual comparison found $R^2$ values of 0.35 and 0.20 for 10- and 5-year growth in easy plots, respectively, and $R^2$ values of less than 0.1 in difficult test sites. The direct approach also consistently overestimated DBH growth relative to both the manual and model-based approaches, with biases of 0.3--1.2 cm (13--43\%) for 10-year growth and 0.4--0.9 cm (20--62\%) for 5-year growth. In comparison, the model-based approach shows some underestimation of 10-year DBH growth, at $-0.1$ cm ($-2\%$) to $-0.4$ cm ($-18\%$), and overestimation of 5-year growth, at 0.1--0.5 cm (6--36\%), relative to manual estimates.

For stem volume change, all three approaches showed strong agreement in easy plots across both time spans, with $R^2$ values ranging from 0.70 to 0.86. In difficult plots, the $R^2$ remains high for 10-year growth between the model-based and manual approaches, at 0.82. Meanwhile, other comparisons show reduced, though still moderate, $R^2$ values at around 0.38--0.59. As with DBH change, the lowest RMSE and MAE values were obtained from the model-based-manual comparison. The RMSE was around 0.03 $\mathrm{m}^3$ (21--31\%) for 10- and 5-year growth in easy plots, and around 0.10 $\mathrm{m}^3$ (67\%) and 0.07 $\mathrm{m}^3$ (87\%) in difficult plots, respectively. The comparisons with the direct approach obtained RMSEs in easy plots of around 0.05 $\mathrm{m}^3$ (34--37\%) for 10-year growth and 0.03 $\mathrm{m}^3$ (32--36\%) for 5-year growth. In difficult plots, the corresponding RMSEs were around 0.15 $\mathrm{m}^3$ (96--128\%) and 0.09 $\mathrm{m}^3$ (109--134\%).

\begin{figure*}[hbp] 
\centering
\begin{subfigure}{0.49\textwidth}
    \caption{\label{fig:error_boxplots_dbh} DBH change 2014--2024}
    \includegraphics[width = \textwidth]{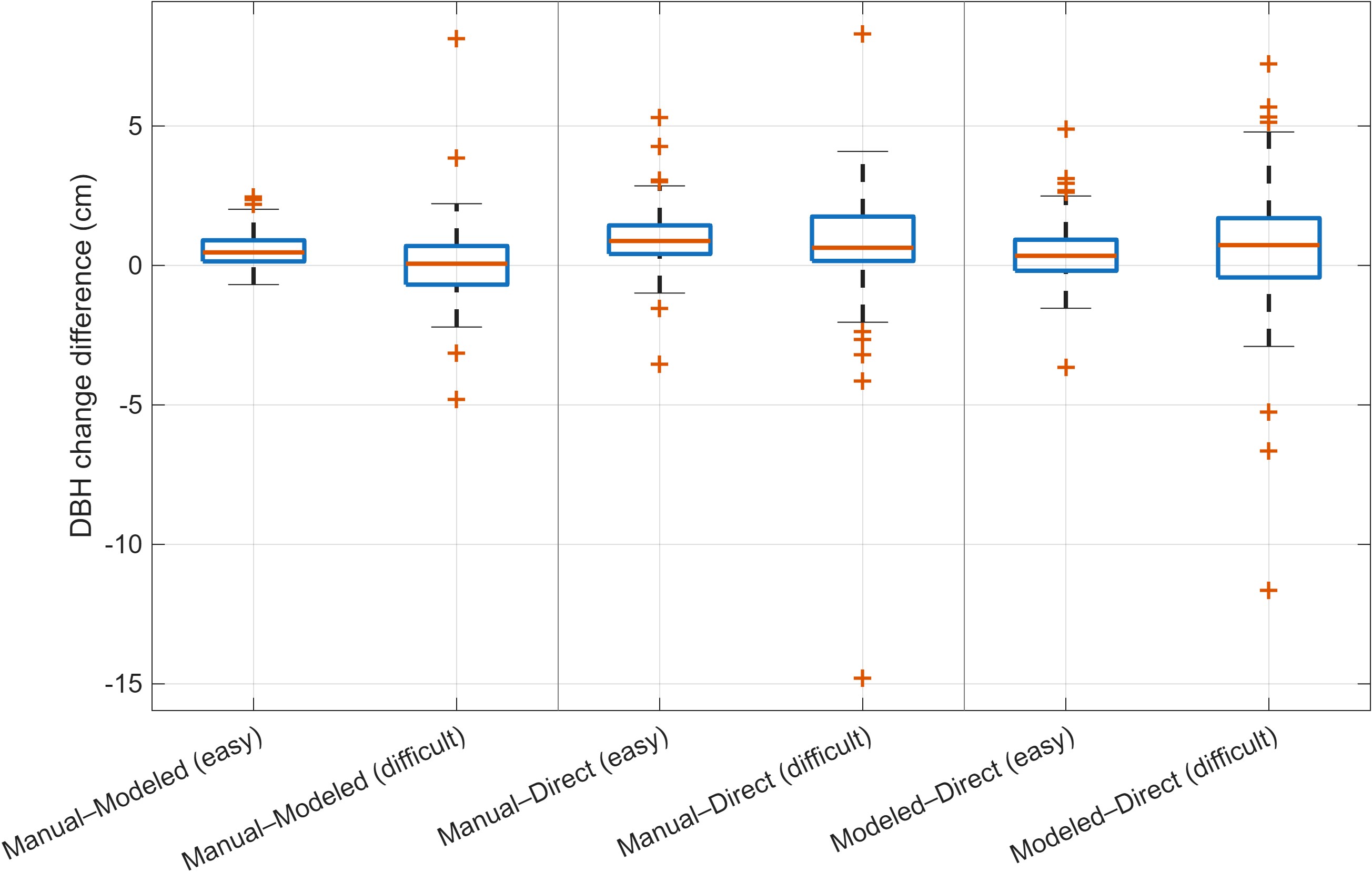}
\end{subfigure} 
\begin{subfigure}{0.49\textwidth}
    \caption{\label{fig:error_boxplots_vol} Volume change 2014--2024}
    \includegraphics[width = \textwidth]{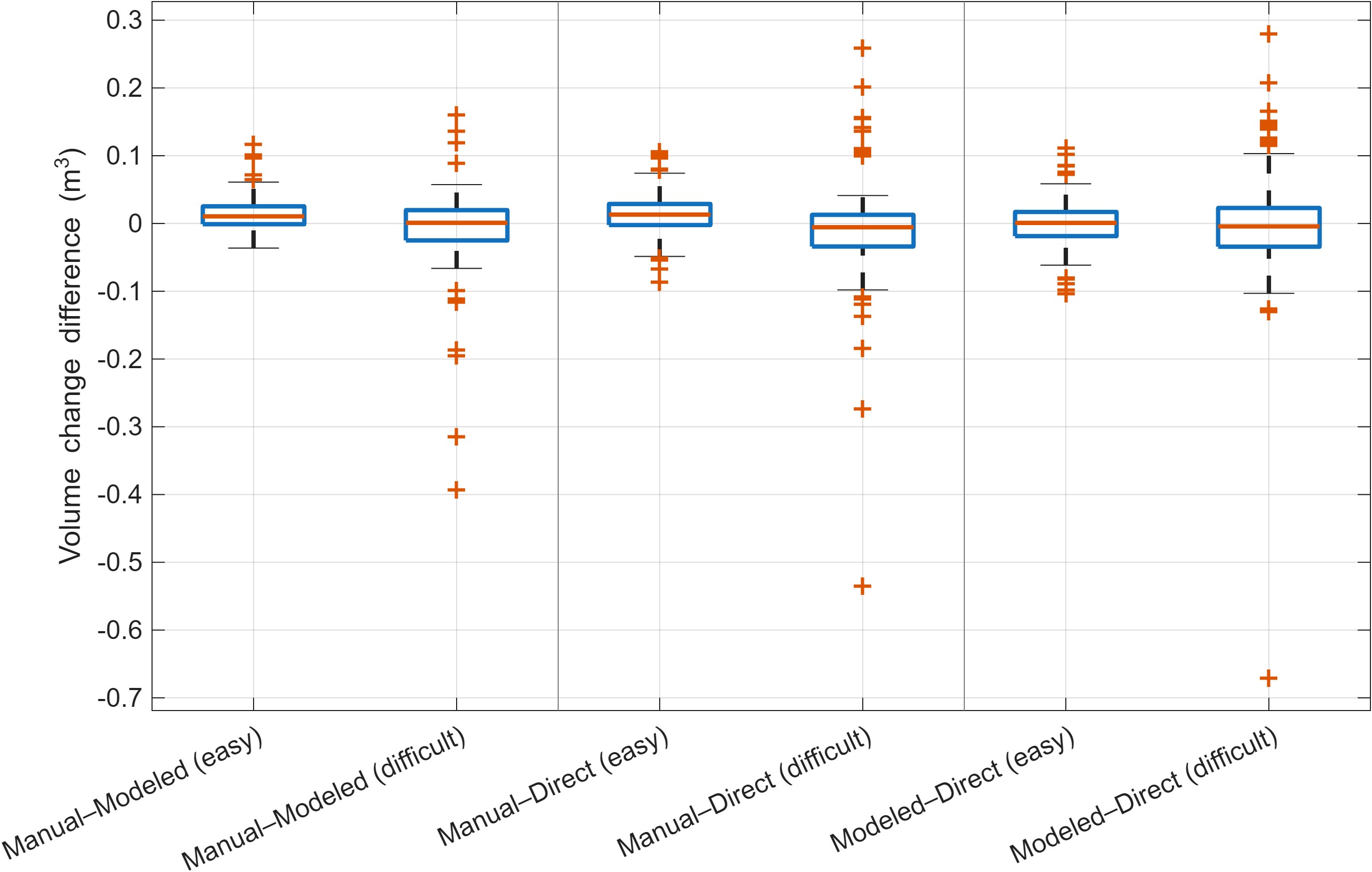}
\end{subfigure}
\caption{\ref{fig:error_boxplots_dbh}: Comparison of the differences in estimated 2014--2024 DBH change between the manual, model-based, and direct approaches in easy and difficult plots. Boxes represent the interquartile range (25th--75th percentiles) with the median indicated by the central orange line. Black whiskers extend to the maximum and minimum values that are not considered outliers, with orange crosses illustrating the outliers. \ref{fig:error_boxplots_vol}: The same for stem volume.}
\end{figure*}

Figs. \ref{fig:error_boxplots_dbh} and \ref{fig:error_boxplots_vol} further illustrate the distributions of differences between the approaches for 10-year DBH and stem volume growth using boxplots. For both attributes, the spread of differences between the methods was larger in difficult plots. These plots also contained a larger number of outliers, indicating increased uncertainty and variability under more complex forest conditions. The boxplots further highlight the systematic overestimation of DBH growth by the direct approach relative to the other methods. In addition, the model-based-manual comparison exhibited the smallest spread of differences, further emphasizing the strong agreement between these approaches.

Additional results describing the accuracy of one-time DBH and stem volume estimation, as well as modeled estimates, for individual years are provided in Appendix \ref{sec:one_time_accuracy}.

\begin{table*}[htp]
\centering
\caption{Comparison of DBH growth estimates for individual trees over the periods 2014--2024 and 2019--2024 in easy and difficult plots. Pairwise differences between the manual, model-based, and direct approaches are summarized using bias, RMSE, MAE, R, and $R^2$. The results in bold indicate the pairing with the highest agreement between the attribute estimates.}
\label{tab:dbh_growth_accuracies}
\begin{tabular*}{\textwidth}{@{\extracolsep\fill}lllll}
\toprule
\textbf{DBH growth} & \textbf{Model-based vs manual} & \textbf{Direct vs manual} & \textbf{Direct vs model-based} \\
\midrule
\textbf{Easy plots} \\
\midrule
\textbf{2014--2024} \\
\midrule
Bias (cm) & \textbf{$\mathbf{-}$0.1 ($\mathbf{-}$2.3\%)} & 1.1 (39.3\%) & 1.2 (42.6\%) \\
RMSE (cm) & \textbf{0.7 (26.1\%)} & 1.6 (56.6\%) & 1.7 (61.9\%) \\
MAE (cm) & \textbf{0.5 (16.0\%)} & 1.2 (40.6\%) & 1.1 (40.7\%) \\
R & \textbf{0.77} & 0.59 & 0.51 \\
$R^2$ & \textbf{0.59} & 0.35 & 0.26 \\
\midrule
\textbf{2019--2024} \\
\midrule
Bias (cm) & 0.5 (35.6\%) & 0.9 (61.9\%) & \textbf{0.4 (19.4\%)} \\
RMSE (cm) & \textbf{0.8 (54.6\%)} & 1.3 (87.3\%) & 1.0 (50.9\%) \\
MAE (cm) & \textbf{0.5 (36.5\%)} & 0.9 (60.4\%) & 0.6 (28.1\%) \\
R & \textbf{0.48} & 0.45 & 0.42 \\
$R^2$ & \textbf{0.23} & 0.20 & 0.17 \\
\midrule
\textbf{Difficult plots} \\
\midrule
\textbf{2014--2024} \\
\midrule
Bias (cm) & $-0.4$ ($-18.2\%$) & \textbf{0.3 (12.9\%)} & 0.7 (38.0\%) \\
RMSE (cm) & \textbf{1.6 (66.9\%)} & 2.5 (105.9\%) & 2.5 (129.1\%) \\
MAE (cm) & \textbf{0.8 (31.4\%)} & 1.0 (41.1\%) & 1.3 (65.2\%) \\
R & \textbf{0.57} & 0.28 & 0.33 \\
$R^2$ & \textbf{0.32} & 0.08 & 0.11 \\
\midrule
\textbf{2019--2024} \\
\midrule
Bias (cm) & \textbf{0.1 (5.6\%)} & 0.7 (56.1\%) & 0.6 (47.8\%) \\
RMSE (cm) & \textbf{1.4 (110.8\%)} & 2.3 (184.4\%) & 2.5 (187.2\%) \\
MAE (cm) & \textbf{0.7 (55.4\%)} & 0.9 (74.6\%) & 1.1 (83.7\%) \\
R & 0.16 & \textbf{0.23} & 0.10 \\
$R^2$ & 0.02 & \textbf{0.05} & 0.01 \\
\bottomrule

\end{tabular*}

\end{table*}

\begin{table*}[htp]
\centering
\caption{Comparison of volume growth estimates for individual trees over the periods 2014--2024 and 2019--2024 in easy and difficult plots. Pairwise differences between the manual, model-based, and direct approaches are summarized using bias, RMSE, MAE, R, and $R^2$. The results in bold indicate the pairing with the highest agreement between the attribute estimates.}
\label{tab:vol_growth_accuracies}
\begin{tabular*}{\textwidth}{@{\extracolsep\fill}llll}
\toprule
\textbf{Volume growth} & \textbf{Model-based vs manual} & \textbf{Direct vs manual} & \textbf{Direct vs model-based} \\
\midrule
\textbf{Easy plots} \\
\midrule
\textbf{2014--2024} \\
\midrule
Bias ($\mathrm{m}^3$) & $\mathbf{-}$\textbf{0.006} \textbf{(}$\mathbf{-}$\textbf{3.7\%)} & 0.027 (18.1\%) & 0.033 (22.7\%) \\
RMSE ($\mathrm{m}^3$) & \textbf{0.031 (21.1\%)} & 0.051 (33.9\%) & 0.053 (36.8\%) \\
MAE ($\mathrm{m}^3$) & \textbf{0.015 (9.9\%)} & 0.028 (18.8\%) & 0.031 (21.6\%) \\
R & \textbf{0.93} & 0.89 & 0.90 \\
$R^2$ & \textbf{0.86} & 0.79 & 0.81 \\
\midrule
\textbf{2019--2024} \\
\midrule
Bias ($\mathrm{m}^3$) & 0.014 (15.1\%) & 0.014 (15.2\%) & \textbf{0.001 (0.5\%)} \\
RMSE ($\mathrm{m}^3$) & \textbf{0.027 (30.5\%)} & 0.033 (35.9\%) & 0.033 (31.5\%) \\
MAE ($\mathrm{m}^3$) & \textbf{0.016 (17.2\%)} & 0.016 (18.1\%) & 0.018 (17.7\%) \\
R & 0.87 & \textbf{0.88} & 0.84 \\
$R^2$ & 0.76 & \textbf{0.77} & 0.70 \\
\midrule
\textbf{Difficult plots} \\
\midrule
\textbf{2014--2024} \\
\midrule
Bias ($\mathrm{m}^3$) & $-0.038$ ($-24.9\%$) & $\mathbf{-}$\textbf{0.018} \textbf{(}$\mathbf{-}$\textbf{12.0\%)} & 0.020 (17.2\%) \\
RMSE ($\mathrm{m}^3$) & \textbf{0.102 (66.8\%)} & 0.147 (96.2\%) & 0.147 (128.3\%) \\
MAE ($\mathrm{m}^3$) & \textbf{0.019 (12.2\%)} & 0.023 (15.2\%) & 0.024 (20.9\%) \\
R & \textbf{0.90} & 0.77 & 0.76 \\
$R^2$ & \textbf{0.82} & 0.59 & 0.58 \\
\midrule
\textbf{2019--2024} \\
\midrule
Bias ($\mathrm{m}^3$) & $-0.009$ ($-10.8\%$) & $-0.009$ ($-10.8\%$) & \textbf{0.000(02) (0.03\%)} \\
RMSE ($\mathrm{m}^3$) & \textbf{0.069 (87.1\%)} & 0.086 (109.4\%) & 0.094 (134.0\%) \\
MAE ($\mathrm{m}^3$) & \textbf{0.023 (29.3\%)} & 0.024 (30.9\%) & 0.032 (44.8\%) \\
R & \textbf{0.73} & 0.71 & 0.62 \\
$R^2$ & \textbf{0.53} & 0.50 & 0.38 \\
\bottomrule
\end{tabular*}
\end{table*}

\section{Discussion}
\subsection{Deep learning-based segmentation was highly effective}
This study used a deep learning-based segmentation method to delineate individual trees for subsequent attribute and growth estimation. The segmentation performed well in both easy and difficult plots. Trees shorter than 7 m were over-segmented, primarily because the manual reference excluded trees with DBH below the 5 cm threshold and because some false positives may not have been captured by our simple filtering rule. In easy plots, trees above 7 m were segmented accurately with an almost equal distribution compared to the manually measured trees. The tree finding rate was 92\%, or 98\% when the manually measured trees with height below 7.64 m were excluded. In difficult plots, the segmentation also closely followed the reference distribution for trees taller than 7 m, but there were some intervals with over-segmentation. Some of the over-segmented intervals may have resulted from false positives that were not captured by the filter or from inaccuracies in the manual or MLS height measurements, but this was not investigated further. The tree finding rates were 89\% when all manually measured trees were counted, and 101\% if the height threshold was set to 7.64 m also for manually measured trees. The deep learning-based segmentation of the MLS point clouds and the subsequent segmentation transfer did not appear to limit tree growth monitoring across different size classes or canopy positions, including both dominant and suppressed trees. Instead, the primary limiting factor was the reliable detection of tree stems across all MLS and TLS point clouds.

Despite the overall good performance, some limitations in the segmentation process were identified. The most prominent issue was that some trees were assigned to multiple tree segments, with the stem curve algorithm detecting stems with more than one tree ID. These trees were subsequently removed from the analysis and examined in Appendix \ref{app:errors_in_segmentation_and_stem_curve_extraction}. A related issue were cases where multiple stems were detected from one segment by the stem curve algorithm. This can be caused by a forked tree, a group of trees growing from the same spot, or a segmentation error where two or more trunks are in the same segment (Fig. \ref{fig:negative_dbh_2014_1} is an example of the last two cases). Inference on such stems was easier than the multi-ID trees, although linking them to manually measured trees was more complicated.

Stem curve quality could potentially be improved through semantic segmentation, which the ForestFormer3D model used here is capable of \citep{Xiang2025a}, to remove leaf and branch points from the point cloud prior to the application of the stem curve algorithm. The trunk segments could then be used as input for the stem curve algorithm, thereby restricting the algorithm to only points originating from tree stems. This could reduce noise introduced by branch and foliage points and improve the robustness of stem curve estimation.

\subsection{Segmentation transferring was an easy-to-implement and fast way of copying segments}
In this study, segmentation transfer was applied to 128 point clouds acquired using several different sensors over a 12-year period. The segmentation transfer required at most two minutes per plot, which enabled us to quickly segment all the used point clouds. Importantly, the segmentation transfer enabled consistent tree segmentation across point clouds collected at different time points, and greatly simplified the identification of the same trees across datasets as only a reference to the ID of the tree was required for identification. The resulting height time series indicated successful segmentation transfer in easy plots, as evidenced by the smooth height growth trajectories shown in Figs. \ref{fig:time_series1.0} and \ref{fig:time_series2.0}. In difficult plots, successful segmentation transfer was also seen as indicated by trees with smooth height trajectories, although some trees showed abrupt changes. These discontinuities were partly attributable to insufficient point cloud coverage. We did not explicitly quantify the success of segmentation transfer, although some of the abrupt changes in the height growth series may have resulted from one ID depicting different trees. Nevertheless, the smooth DBH and stem volume growth trends suggest that segmentation transfer was adequately successful at the stem level. 

The proposed segmentation transfer approach relied on accurate georeferencing of all point clouds, and the transferring was based on spatial distance between the source and target points. Although this approach did not appear to limit the results in this study, it has inherent theoretical limitations. For example, if the original segmentation contains two trees such that the taller one covers the shorter one, and the segmentation is transferred to a point cloud from a distant future, then over a sufficiently long time period the shorter tree may have grown enough that it obtains segment labels from the original taller tree. This issue could perhaps be alleviated through, for example, majority voting where one tree cluster or point would get the majority of the segment labels from nearby source labels similarly to \citet{Cao2023}.

We also found that the initially segmented source point cloud should cover a larger area than the target point clouds to which the segmentation is transferred to. This is because the trees at the edge of the target point cloud obtain the segment labels from the nearest tree regardless if they reside inside the source point cloud or outside of it. Trees outside the spatial extent of the source point cloud cannot receive their correct labels and instead inherit the label of the nearest available source tree. A similar issue arises for fallen trees that obtain the segment labels from neighboring living trees. However, because fallen trees are typically excluded from growth studies or analyzed separately, this limitation is unlikely to have a substantial practical impact.  

\subsection{The scaling model stabilized in 5--6 years}
The scaling model relied on accurate ALS-derived height time series. In easy plots, height estimates were consistent across years, whereas the difficult plots exhibited less consistent growth trajectories, particularly for small trees. These findings highlight limitations in height estimation using ALS alone, where inconsistent results were shown from different years using different sensors and approaches.

Overall, the scaling model used in this study was robust when evaluated against manual measurements. The RMSE for both DBH and stem volume did not increase rapidly when either forecasting or hindcasting. In addition, the RMSE remained comparable to the initial error of 2014 for DBH and stem volume when compared with manual measurements. When compared to direct stem curve measurements, most RMSE growth occurred within the first 5--6 years. In the forecasted results, RMSE stabilized after approximately 5 years (2014\textrightarrow2019), while in the hindcasted results most RMSE growth occurred over a 6-year period (2025\textrightarrow2019). However, shorter and longer periods were also observed.   

The quality of the point cloud in the starting year influenced the modeled results. The RMSE of the forecasted estimates for both DBH and stem volume was similar to the hindcasted estimates, although slightly higher, particularly when compared to manual measurements. Forecasted values were also more consistently underestimated, whereas hindcasted biases were closer to zero. This difference is likely explained by differences in point cloud quality. The point clouds from 2014, which was the starting year for forecasting, had approximately half the point density of the 2025 point clouds and were acquired using only five scanner positions (center + four auxiliary). In contrast, the hindcasted results were derived from an MLS sensor that scanned all sides of the tree and produced a denser point cloud.

The study by \citet{hyyppa2020comparison} compared different scanner carriers for point cloud acquisition and evaluated the stem curve algorithm performance on the point clouds. The algorithm obtained DBH RMSE values of 4--5\% in an easy plot and 4--8\% in a medium plot using similar under-canopy MLS systems as in this study against a reference measured manually from a point cloud. The scaling model in our study achieved better accuracy for approximately 1.5--2 years calculated from the hindcasted values when compared with direct measurements of stem curve. After that, the performance of both forecasted and hindcasted values was not substantially worse. For stem volume, the corresponding RMSE values were 10--15\% and 9--12\%. Our model achieved comparable accuracy in easy plots across the full 12-year period, while the accuracy in difficult plots was worse.

The most similar comparison of individual tree attribute accuracies was done in \citet{muhojoki2024benchmarking}. In their study, they compared multiple scanners, with the reference measurements calculated in the same way as we did with the direct measurements of stem curve. Their most comparable method (a handheld ZEB Horizon without bias correction) yielded an RMSE of 6\% in easy plots, which is the same as our model for almost the whole study period, and 7\% in difficult plots, which is lower than our corresponding estimates. For stem volume, they achieved an RMSE between 20--23\%. Our scaling model achieved better results for the whole study period, both when forecasting and hindcasting the stem volume. These comparisons with other studies suggest that the scaling model can be used for attribute predictions and produces attribute estimates with accuracy (RMSE-wise) comparable to those obtained from other scanning devices, provided that the initial scanner used for obtaining the stem curve is a suitable one, such as a high-quality MLS or TLS scanner.

Finally, the 2019 sparse point clouds were found to be suitable for the scaling method. However, as with other point clouds, the quality of the initial scan should be assured. In particular, bias in the 2019-derived estimates depended on the starting year and sensor. Overall, the results are promising, as they indicate that even a sparse, large-area ALS acquisition can support scaling-based change detection under appropriate conditions.

\subsection{Including stem taper change information did not improve the model}
The stem taper model used in our study did not improve the accuracy of the scaling model. The stem taper functions used in this study were parametrized only by species, DBH, and height information. Furthermore, we used the scaled DBH ($b\cdot\mathrm{DBH}$) as the input for the predicted stem taper, which propagated the error in DBH prediction into the modeled stem taper. The equations also do not account for individual tree characteristics such as canopy metrics, leaving opportunities for developing future models that incorporate richer laser scanning-based features for predicting individual tree stem taper.

The scaling model could perhaps be improved by calibrating the growth factor $b$ using some auxiliary data, as we assumed that height growth translates directly into diameter growth, which may be an oversimplification. On the other hand, we were not able to improve the predictions in the present study by incorporating stem taper information, so this remains an open question for future research. Further improvements in stem curve modeling could also be achieved by incorporating the model for reducing diameter bias caused by non-zero laser beam width proposed by \citet{muhojoki2024benchmarking}.

\subsection{Model-based growth estimation outperformed direct approach}
Comparing manual, model-based, and direct approaches for DBH and volume growth estimation showed that the model-based and manual methods consistently exhibited the highest agreement across plot difficulties and time spans. This was reflected in the lowest RMSE and MAE values as well as highest $R^2$ in almost all cases. For DBH growth in particular, the direct approach obtained a much higher RMSE relative to the manual estimates as well as a large positive bias, indicating systematic overestimation. In contrast, the differences between the methods for volume growth estimation were smaller. However, the model-based approach still demonstrated the strongest agreement with the manual estimates, especially in difficult plots. Therefore, these results indicate that the model-based approach provides more reliable growth estimates compared to the direct approach, particularly for DBH.

An advantage of the model-based approach is that it reduces the number of independent error sources due to fewer point clouds used in change detection. Differencing two independently estimated attributes, as was done in the direct approach, has four error sources (two under-canopy scans and two ALS scans). In contrast, the model-based approach removes the need of the second under-canopy scan, and associates growth to only height difference, which was measured only from two point clouds. The modeling also inherently limits negative stem growth, with the outliers obtained using this method being more modest. Meanwhile, negative stem growth due to measurement errors was possible to obtain using the other approaches. Consequently, some outliers from the direct approach had very large negative growth estimates. 

In comparison to previous studies, the model-based approach achieved accuracy relative to manual measurements comparable to or exceeding that reported by \citet{Tavi2026}. Accuracy comparable to that reported by \citet{wang2025forest} was observed in easy plots, while higher errors were observed in difficult plots. The direct method produced DBH growth estimates with generally lower agreement with the manual method than those obtained by \citet{yrttimaa2022exploring} for 5-year TLS-based direct change estimation. However, the results for volume growth of this study had lower RMSE and higher $R^2$ in easy plots, while there were higher errors or more comparable results in difficult plots. Compared to long-term ALS-based growth estimation by \citet{Soininen2022}, the present study achieved higher or comparable $R^2$ values for DBH growth. In addition, volume change was estimated with higher accuracy and higher $R^2$ values in all cases, highlighting the benefit of incorporating MLS data for stem attribute estimation alongside ALS-derived heights, as opposed to relying on ALS data only.

The direct and manual approaches used the same height measurements for both time points to calculate volume growth. The modeled approach then used these same height estimates for the 2024 heights, but used fully ALS-based heights for the past estimates in 2014 and 2019, in order to follow the framework of reconstructing past attributes without the use of under-canopy MLS data. The reason for using the same height estimates in most cases was due to a lack of alternative data available, in addition to the combination of ALS with MLS/TLS considered to provide the best possible height estimates, especially in more complex forest environments with more occlusion and suppressed trees. The height-time series produced in this study also show that the specific years that were used for the growth analysis obtained comparatively consistent and robust height estimates using both ALS as well as combined ALS and MLS/TLS for the estimation, although the use of the combined point clouds improved the height estimates in the difficult plots.

\subsection{Modeled attributes were more similar to the direct measurements in the one-time study but more similar to the manual measurements in the growth study}
There appears to be a discrepancy with the results between the one-time attribute estimation study and the growth study. In the one-time study, the modeled attributes were more similar to the direct stem curve measurements. This is because both methods were based on the same stem curve algorithm. In contrast, the manual measurements of stem volume were based on the Laasasenaho allometric equations, which used only height and DBH as fitting parameters and therefore represented a coarser approximation of the stem volume. The manual DBH measurements could be considered the most accurate among the available DBH data.

In the growth study, particularly over shorter timescales, the model-based stem volume estimates benefited from fewer fitting parameters. Stem volume growth derived from the difference of two Laasasenaho-based estimates was a function of only four variables: DBH\textsubscript{\textit{tgt}}, DBH\textsubscript{\textit{src}}, $h_\textit{tgt}$ and $h_\textit{src}$. Similarly, the growth predicted using the scaling model was a function of only two variables: $h_\textit{tgt}$ and $h_\textit{src}$ in addition to the rest of the fitting process of the stem curve algorithm. The direct measurements of the stem curve, in contrast, were influenced by many more differences in the value-derivation process, and arguably the most significant difference was the point cloud quality. As a consequence, it is easier to obtain correlating results with the more parsimonious methods. Additionally, the height measurements from the combined ALS and MLS/TLS point clouds were in some cases equal to the height measurements from the ALS-only point clouds, which made the simpler methods more similar.

Across all growth comparisons, higher agreement between growth measuring methods was consistently observed for stem volume growth than for DBH growth. This is likely because all three DBH growth measurements methods were very dissimilar whereas the growth in stem volume was heavily associated with height growth in the manual and model-based approaches. Nevertheless, the results of the growth study can be interpreted so that the scaling method is superior to direct stem curve measurements for DBH growth, as the manual DBH measurements can be considered as an absolute reference of DBH. The results for stem volume growth indicate that the methods using fewer parameters produce more similar values, but due to lack of absolute stem volume reference, such as manual derivation of stem volume from the point clouds, it is not possible to say which method is the most accurate in an absolute sense.

\subsection{Sources of error and limitations}
Despite good accuracy of the model-based approach relative to the manual measurements, the model-based approach also introduces limitations. DBH size was incorporated into the modeling in the form of the $b$ parameter. However, other variables such as species, age, climatic variability such as temperature and precipitation, as well as competition effects related to the position and social status of the tree were not considered, even though they can impact the height-diameter relationship of trees \citep{irwin2025prioritizing, han2025age, liu2025modeling, ahmed2024neighborhood, fortin2019evidence}. In addition, although changes in stem taper over time were examined, their effect on growth estimates was minimal and therefore excluded from the further analysis into attribute growth. As a result, the modeling framework overall represents a simplified approximation of tree growth, and therefore also limits the ability to capture individual tree growth dynamics, non-linear growth patterns, or anomalous behavior.

Plot complexity was consistently shown to impact results throughout this study, with higher errors observed for the estimation of one-time attributes, modeled attributes, and growth in difficult plots. These plots were characterized by higher tree density, greater species diversity, more small trees, and increased understory vegetation. High stem density and understory vegetation contribute to occlusion, reducing the visibility of tree stems and the quality of stem measurements derived from the MLS/TLS data. Additionally, occlusion of tree canopies affects the ALS-based height estimates used for stem volume and modeling, particularly for suppressed and understory trees. Higher tree density in difficult plots could also increase neighborhood competition effects for trees, thereby impacting the height-DBH relationship and introducing additional modeling uncertainty. Species composition also affected errors in difficult plots, as previous studies have shown that attribute estimation and modeling is most accurate for pines, followed by spruces, with birches and other deciduous trees having the highest errors \citep{yrttimaa2022exploring, Tavi2026}. As outlined in Table \ref{tab:plot_stats}, the easy plots contained a much higher proportion of pines, which typically have more visible stems that improve stem measurement quality. Furthermore, previous studies have found higher errors for the estimation of one-time attributes and growth in small trees, specifically trees with DBHs smaller than 15--20 cm \citep{Tavi2026}. Reasons include smaller trees having higher errors in ALS-based height estimation, which affects both volume estimation and modeling, due to small trees being more likely to be suppressed with occluded canopies. Moreover, under-canopy MLS/TLS stem curve extraction for trees with a smaller DBH is challenging due to occlusion from other stems and vegetation as well as a smaller visible stem surface area that makes noisier stem curve fitting more likely.

Additional limitations are related to data acquisition and processing. Differences in the timing of data collection within the growing season, as shown in Table \ref{tab:data_characteristics}, may have introduced variability into the reported one-time attribute accuracies, modeled attribute accuracies, and comparisons between growth estimation approaches. The ALS, MLS/TLS, and manual measurements were not always acquired during the same period of the year, meaning that the datasets did not necessarily represent identical stages of annual tree growth. Consequently, the growth intervals represented by the ALS-derived heights, MLS/TLS-derived stem measurements, and manual measurements were not perfectly aligned. This temporal mismatch is particularly relevant for the ALS-derived height growth that was used to scale stem attributes. As a result, part of the observed differences between methods may reflect acquisition timing rather than methodological performance alone. Furthermore, the manual measurements themselves are subject to measurement uncertainty, introducing an additional source of variability into the reported accuracies.

During the creation of DTMs, empty voxels resulted in holes in the DTM, which affected some point clouds with a sparse set of points labeled as background. We noticed holes in the DTMs only after we had processed most of the point clouds. The holes affected the tree height measurements of some trees in the plot, as the point clouds were not normalized correctly above the holes. Hence, we re-fitted the DTM and the stem curves for those point clouds where the tree height seemed to have increased due to a hole in the DTM below the tree, which corrected the affected plots. However, some errors caused by faulty DTMs may have remained, but they are expected to be only minor errors, as the height growth trajectories in Figs. \ref{fig:time_series1.0}--\ref{fig:time_series2.1} contain only small sharp changes, whereas a hole usually affected tree height by several meters.

Finally, previous work by \citet{Tavi2026}, which employed a similar height growth-based scaling model, demonstrated that small errors in ALS-derived heights had only a limited effect on the resulting error magnitudes of the DBH, stem volume, and growth estimates. This suggests that the proposed framework is relatively robust to minor inaccuracies in height estimates, and consequently, to small differences in data collection timing within the growing season. At the same time, the height time series analyzed in this study highlighted that ALS-derived heights can vary between sensors and acquisitions, particularly in dense forest conditions. For the specific years used in the growth analyses (2014, 2019, and 2024), the ALS point clouds produced relatively consistent height estimates. However, larger discrepancies in ALS-derived heights could propagate through the scaling model and reduce growth estimation accuracy. Therefore, although the framework appears robust to small height errors, its performance remains dependent on the quality and consistency of the ALS data used to derive height growth. If using years where individual ALS point clouds are sparse or otherwise unreliable, height estimates could have also been improved by incorporating information from the full multitemporal height series rather than relying on individual acquisition years.

\section{Conclusion}
This study evaluated a framework for estimating individual tree DBH and stem volume growth by combining single-date under-canopy laser scanning point clouds for stem measurements with multitemporal ALS-derived height observations. Using 136 point clouds acquired between 2014--2025 from 11 different sensors across ALS, MLS, and TLS platforms in eight boreal forest test sites, we assessed the complete workflow from tree segmentation and segment transfer to attribute estimation, temporal scaling of attributes, and growth reconstruction.

The results demonstrated that segmentation transfer from a single high-quality MLS acquisition can maintain reliable tree correspondence across multitemporal point clouds despite substantial differences in point density, sensor characteristics, and acquisition methods. The multitemporal observations enabled the construction of long-term individual-tree attribute time series and the evaluation of the proposed height growth-based scaling model for reconstructing stem attributes forwards and backwards in time.

The height growth-based scaling model was found to be temporally robust, despite being applied to point clouds originating from different sensors and acquisitions. When evaluated against manual measurements, errors in modeled one-time DBH and volume remained relatively stable over time. When compared to direct point cloud-based estimates, most error accumulation occurred during the first 5--6 years, after which RMSE growth largely stabilized. These findings suggest that ALS-derived height growth can be used to reconstruct past or future stem attributes over operationally relevant monitoring periods without substantial loss of accuracy. However, error propagation in the scaling model depended on the quality of the initial point cloud data. Additionally, the variability observed in the ALS-derived height time series highlighted that differences between sensors and acquisitions can sometimes lead to unreliable height estimates, particularly in dense forest stands, emphasizing the importance of the choice and availability of ALS datasets. Incorporating stem taper variation into the scaling model did not significantly improve the results.

A key finding of this study is that the proposed model-based framework for growth estimation consistently outperformed direct growth estimation, which was based on differencing independently estimated attributes from separate point clouds. Across both 5- and 10-year DBH and volume growth, and across both plot complexities,  comparisons between manual, direct, and model-based growth estimates showed that the model-based and manual approaches exhibited the highest agreement. This was reflected in lower RMSE values, higher $R^2$ values, and fewer extreme outliers between these two approaches. The direct growth estimation was more sensitive to measurement noise and produced larger errors and outliers.

The results further demonstrated that reliable tree growth estimation does not require historical under-canopy MLS or TLS data. Instead, a single under-canopy dataset, combined with multitemporal ALS observations, is sufficient to reconstruct past tree attributes and estimate growth with accuracy comparable to, or exceeding, approaches based on repeated direct stem measurements. This substantially improves the practicality of large-scale retrospective growth analyses, as historical MLS/TLS data are rarely available. The results also emphasize that the accuracy of this approach relies on reliable tree segmentation, high-quality point clouds and algorithms for accurate stem measurement, and high quality ALS data for reliable height estimation. Future work on developing the scaling model should focus on calibrating it using auxiliary growth information, incorporating stem taper change through more sophisticated modeling, and incorporating semantic segmentation of tree trunks to further improve stem curve determination.

\section{CRediT authorship statement}
\textbf{DT}: Conceptualization, Methodology, Formal analysis, Investigation, Writing - Original Draft, Visualization. \textbf{VS}: Conceptualization, Methodology, Software, Formal analysis, Data Curation, Writing - Original Draft, Visualization. \textbf{LR}: Methodology, Software, Writing - Original Draft. \textbf{JM}: Investigation, Software, Data Curation, Writing - Review \& Editing. \textbf{JH}: Writing - Review \& Editing, Supervision, Project administration, Funding acquisition.

\section{Acknowledgements}
We gratefully acknowledge the Research Council of Finland (RCF) for the following projects "High-performance computing allowing high-accuracy country-level individual tree carbon sink and biodiversity mapping" (RCF 359203) and "Enhanced Wood Tracing Systems for Sustainable Forestry and Improved Wood Utilization" (decision number 373290). The work was under RCF forestry flagship "Forest-Human-Machine Interplay - Building Resilience, Redefining Value Networks and Enabling Meaningful Experiences" (RCF 359175) using research infrastructure "Measuring Spatiotemporal Changes in Forest Ecosystem" (decision number 346382). We thank Antero Kukko, Harri Kaartinen, Teemu Hakala, and Anttoni Jaakkola for collecting the point clouds used in this study. We also thank Ville Luoma, Tuomas Yrttimaa, Jiri Pyörälä, Osmo Suominen, and Otto Saikkonen for collecting the manual measurements used.

\section{Declaration of generative AI and AI-assisted technologies in the manuscript preparation process}
Microsoft Copilot was used in creating some parts of the illustrations in Fig. \ref{fig:workflow}. During the revision of this work, ChatGPT (OpenAI) was used solely to assist with language editing and improving the clarity of parts of the text. All suggestions were reviewed and, where appropriate, incorporated into the manuscript by the authors, who take full responsibility for the content of the publication.

\section{Declaration of competing interests}

The authors declare that they have no known competing financial interests or personal relationships that could have appeared to influence the work reported in this paper.

{
\footnotesize
\bibliography{references.bib}
}

\clearpage
\normalsize
\begin{appendices}

\section{One-time attribute estimation errors underlying the growth analysis}
\label{sec:one_time_accuracy}
Table \ref{tab:individual_accuracies_growth} summarizes the accuracies of one-time DBH and stem volume estimates in the easy and difficult plots. The results include both attributes directly estimated from the MLS/TLS point clouds and attributes hindcasted from the 2024 estimates using the proposed scaling model. DBH estimates were evaluated against manual measurements, while stem volume estimates were evaluated against reference volumes calculated using the Laasasenaho allometric model with manually measured DBH and species information together with heights obtained from the combined ALS and MLS/TLS point clouds. These one-time attribute estimates correspond directly to the growth estimation results presented in Section \ref{sec:growth_results}, as the same attribute estimates and reference values were used in both analyses.

In the easy plots, both directly estimated and modeled DBH values exhibited relatively consistent accuracies across years. DBH RMSE values were approximately 1.2--1.6 cm (5--7\%). Similarly, stem volume RMSE values were consistent at approximately 0.07--0.08 $\mathrm{m}^3$ (15--18\%). For DBH, the modeled estimates were slightly more accurate than the directly estimated values. For stem volume, the direct estimates achieved higher accuracy in 2014, whereas the modeled values were slightly more accurate in 2019.

In the difficult plots, DBH RMSE values were approximately 2.7--3.2 cm (11--14\%). The modeled estimates in 2019 were slightly more accurate than the directly estimated values, whereas the direct estimates achieved higher accuracy in 2014. Stem volume RMSE values ranged from approximately 0.14 to 0.18 (21--26\%). Similar to the easy plots, the modeled estimates were more accurate in 2019, whereas the 2014 direct estimates produced lower errors than the corresponding modeled estimates.

\section{Errors in segmentation and stem curve extraction}
\label{app:errors_in_segmentation_and_stem_curve_extraction}
This section examines errors in tree segmentation and attribute estimation that are often caused by segmentation anomalies. We observed the segmentation errors arising from the deep learning-based segmentation and the subsequent segmentation transfer to be such that they appear unnatural to a human interpreter. These errors are a drawback to using a deep learning method for segmenting. However, they produce errors that are easily observable (e.g. negative DBH) and repairable through post-processing (e.g. combining points from different segments). Thus, these errors are unlikely to pose major practical problems.

\subsection{One tree in multiple segments}
\begin{figure*}[tb]
    \begin{subfigure}{0.38\textwidth}
        \caption{\label{fig:multiple_segments1}}
        \includegraphics[width=\textwidth]{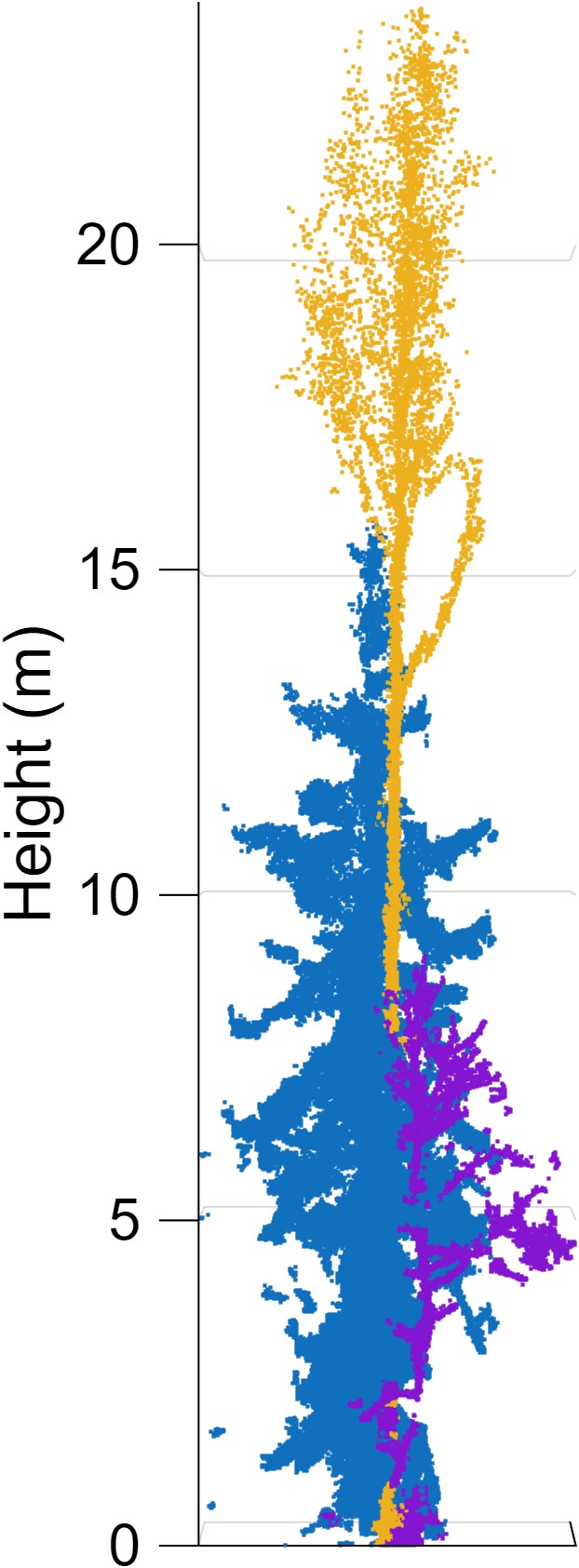}
    \end{subfigure}
    \begin{subfigure}{0.49\textwidth}
        \caption{\label{fig:multiple_segments2}}
        \includegraphics[width=\textwidth]{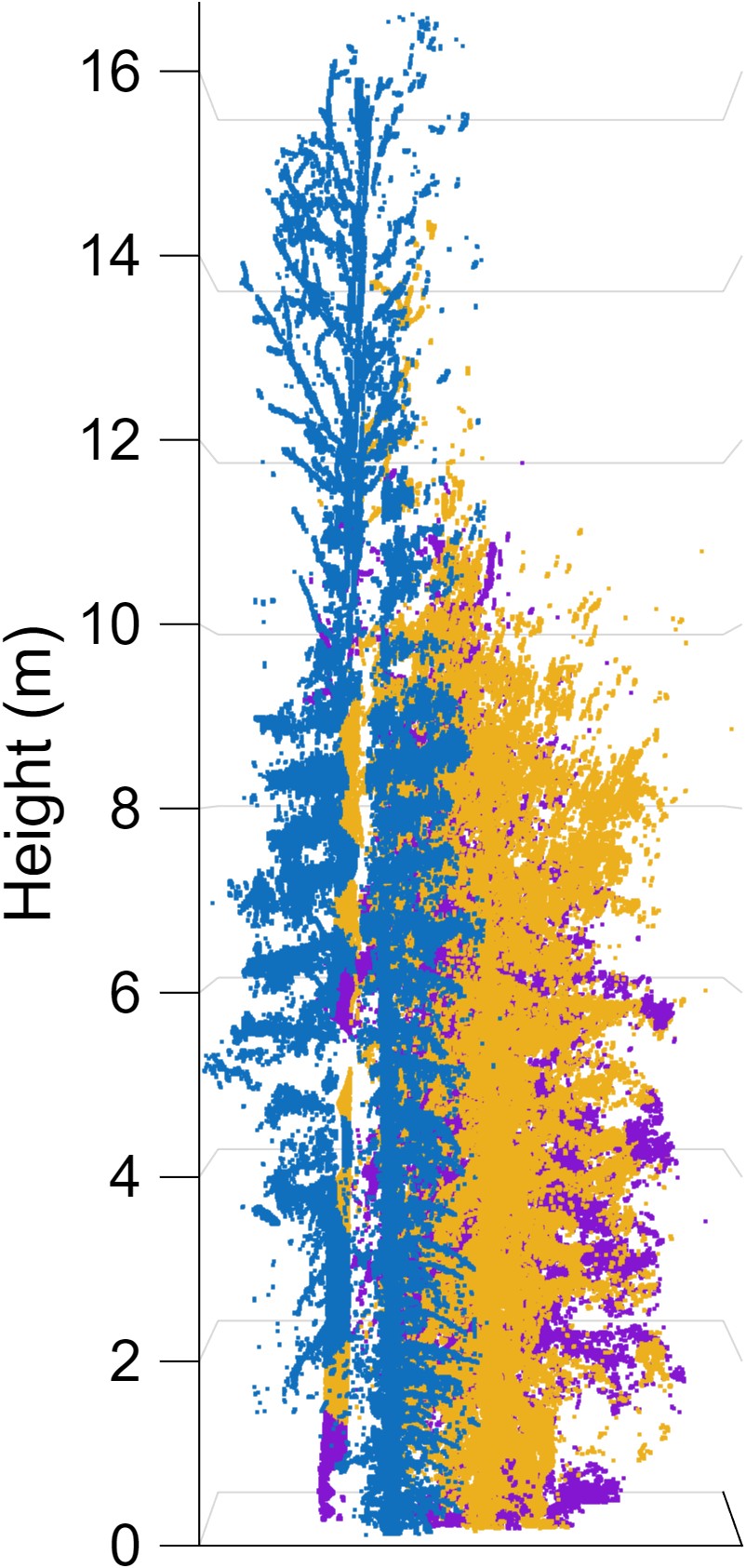}
    \end{subfigure}
    \caption{\label{fig:multiple_segments} Each color represents one segment. \ref{fig:multiple_segments1}: Three trees from the preliminarily segmented 2020 MLS point cloud in a difficult plot. The blue and purple trees consist of a single segment, although parts of the segments belong to the nearby orange tree. The orange tree in the middle, on the other hand, consists of three segments (orange, purple, blue mixed with the background) and was therefore removed from the analysis. \ref{fig:multiple_segments2}: Another example from the 2014 TLS point cloud in a difficult plot. Here the leftmost tree consists of three different segments.}
\end{figure*}

The deep learning-based segmentation or segmentation transfer occasionally assigned a single tree to multiple segments. It would be difficult to determine the correct segment for height determination without additional steps in point cloud processing or visual inspection of point clouds and segments. These trees were excluded from our analysis for simplicity. In principle, however, it would be possible to assign the points from different segments to the correct tree using the detected stem location. Fig. \ref{fig:multiple_segments} shows two example of this type of error.

Fig \ref{fig:multiple_segments1} shows a tree whose stem curve was divided between three segments in 2020 by the deep learning model. On the other hand, the tree in \ref{fig:multiple_segments2} had fallen between 2014 and 2020. As a result, the tree in 2014 was assigned the segment labels from the closest trees in 2020, which resulted in a multi-segment tree. This type of error was due to the segment transferring method.

\begin{table*}[tbp]
\centering
\caption{Accuracy of one-time DBH, stem volume, and modeled attribute estimates in easy and difficult plots. The results include both direct MLS/TLS-based estimates and modeled values hindcasted from the 2024 estimates using the proposed scaling model. The starting year (2024), marked with an asterisk, served as the model input and was therefore not modeled. DBH estimates were evaluated against manual measurements. Stem volume estimates were evaluated against reference volumes calculated using the Laasasenaho allometric model that incorporated manually measured DBH and species information, as well as heights estimated from the combined ALS and MLS/TLS point clouds.}
\label{tab:individual_accuracies_growth}
\begin{tabular*}{\textwidth}{@{\extracolsep\fill}llllll}
\toprule
\textbf{Easy plots} & \textbf{Bias} & \textbf{RMSE} & \textbf{MAE} \\
\midrule
DBH 2024* (cm) & 0.3 (1.4\%) & 1.2 (5.0\%) & 0.4 (1.7\%) \\
DBH 2019 (cm) & $-0.6$ ($-2.5\%$) & 1.5 (6.4\%) & 0.6 (2.7\%) \\
DBH 2014 (cm) & $-0.8$ ($-3.5\%$) & 1.6 (7.1\%) & 0.8 (3.7\%) \\ 
Modeled DBH 2019 (cm) & $-0.2$ ($-0.8\%$) & 1.2 (5.3\%) & 0.5 (2.1\%) \\
Modeled DBH 2014 (cm) & 0.4 (1.8\%) & 1.4 (6.2\%) & 0.7 (3.0\%) \\ 
\hline
Volume 2024* ($\mathrm{m}^3$) & 0.023 (4.3\%) & 0.079 (14.7\%) & 0.026 (4.8\%) \\
Volume 2019 ($\mathrm{m}^3$) & 0.009 (2.0\%) & 0.072 (16.3\%) & 0.023 (5.3\%) \\
Volume 2014 ($\mathrm{m}^3$) & $-0.004$ ($-1.0\%$) & 0.065 (16.9\%) & 0.022 (5.7\%) \\
Modeled volume 2019 ($\mathrm{m}^3$) & 0.010 (2.2\%) & 0.069 (15.6\%) & 0.022 (4.9\%) \\
Modeled volume 2014 ($\mathrm{m}^3$) & 0.029 (7.5\%) & 0.070 (18.1\%) & 0.027 (7.1\%) \\
\midrule
\textbf{Difficult plots} & \textbf{Bias} & \textbf{RMSE} & \textbf{MAE} \\
\midrule
DBH 2024* (cm) & $-0.5$ ($-2.1\%$) & 3.2 (12.6\%) & 0.6 (2.5\%) \\
DBH 2019 (cm) & $-1.2$ ($-5.2\%$) & 3.2 (13.5\%) & 0.8 (3.3\%) \\
DBH 2014 (cm) & $-0.8$ ($-3.7\%$) & 2.6 (11.3\%) & 0.7 (2.9\%) \\
Modeled DBH 2019 (cm) & $-0.6$ ($-2.5\%$) & 3.0 (12.5\%) & 0.8 (3.4\%) \\
Modeled DBH 2014 (cm) & $-0.09$ ($-0.4\%$) & 2.7 (11.8\%) & 0.8 (3.7\%) \\
\hline
Volume 2024* ($\mathrm{m}^3$) & $-0.020$ ($-2.6\%$) & 0.175 (23.4\%) & 0.019 (2.5\%) \\
Volume 2019 ($\mathrm{m}^3$) & $-0.011$ ($-1.7\%$) & 0.164 (24.6\%) & 0.028 (4.1\%) \\
Volume 2014 ($\mathrm{m}^3$) & $-0.001$ ($-0.2\%$) & 0.144 (24.2\%) & 0.025 (4.3\%) \\
Modeled volume 2019 ($\mathrm{m}^3$) & $-0.011$ ($-1.7\%$) & 0.142 (21.2\%) & 0.026 (3.9\%) \\
Modeled volume 2014 ($\mathrm{m}^3$) & 0.018 (3.1\%) & 0.156 (26.3\%) & 0.020 (3.3\%) \\
\bottomrule
\end{tabular*}
\end{table*}

\subsection{Negative DBH and jump in stem volume}
\label{app:removed_trees_dbh_vol}
\begin{figure*}[tb]
    \centering
    \begin{subfigure}{0.45\textwidth}
        \caption{\label{fig:negative_dbh_2014_1}}
        \includegraphics[width=\textwidth]{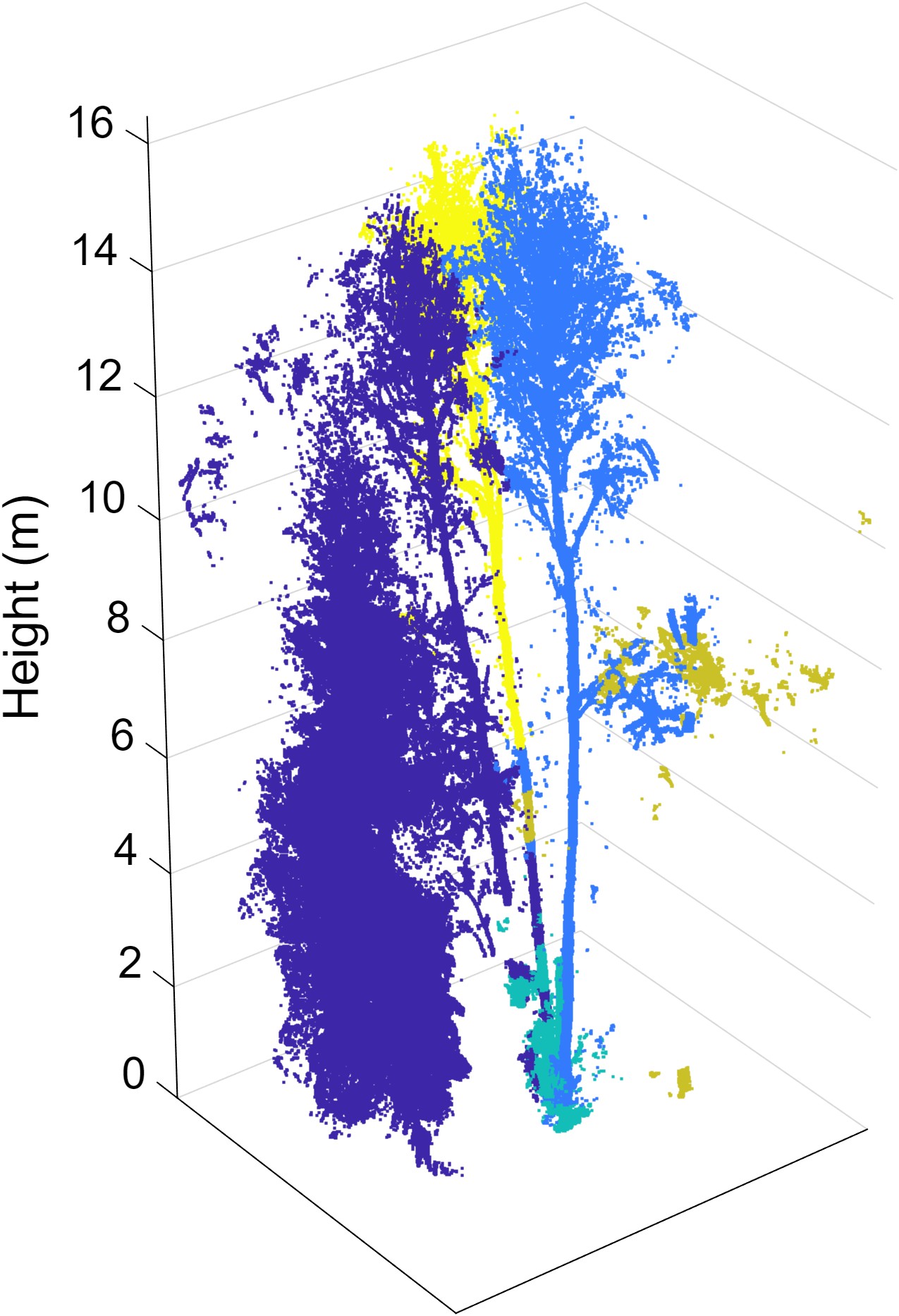}
    \end{subfigure} \\
    \begin{subfigure}{\textwidth}
        \centering
        \caption{\label{fig:negative_dbh_2014_2}}
        \includegraphics[width=0.7\textwidth]{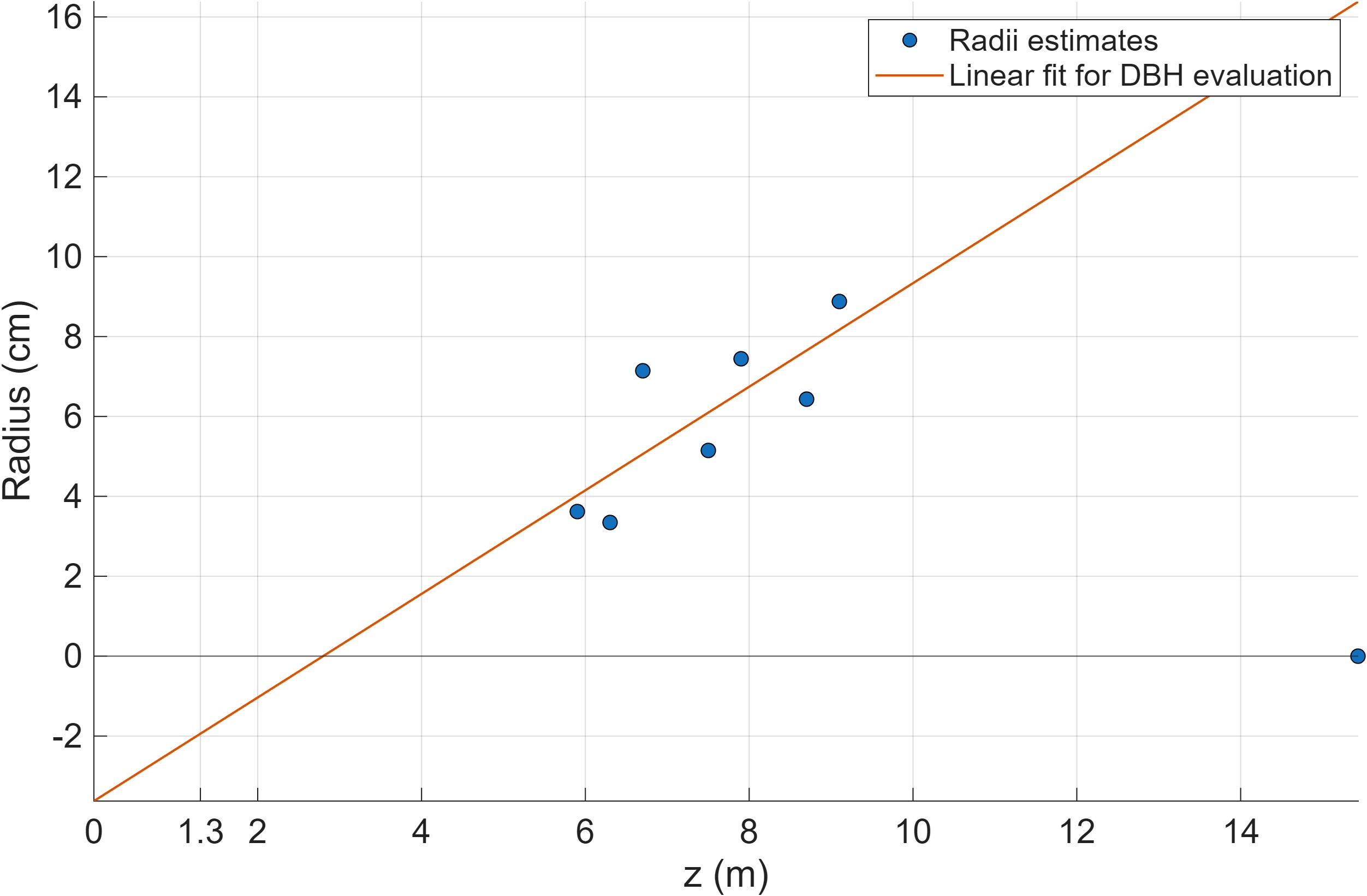}
    \end{subfigure}
    \caption{\ref{fig:negative_dbh_2014_1}: The yellow tree in the middle had a negative DBH in 2014. Its stem was split into multiple segments such that the main segment started at around 6 m, which led to a lack of data points for fitting. \ref{fig:negative_dbh_2014_2}: The radii estimates of the corresponding tree and the DBH extrapolation fit. The DBH at $z = 1.3\mathrm{\, m}$ is marked with a vertical line.}
\end{figure*}

The tree that produced the negative DBH estimate in 2014 (Fig. \ref{fig:time_series4}) was split into multiple segments by the segmentation method and subsequently during the segmentation transferring phase. This caused the main segment to start at a height of approximately 6 m (yellow tree in Fig. \ref{fig:negative_dbh_2014_1}). This led to a negative DBH estimate because the stem diameter estimates $r_i$ at heights $z_i \in [0, h_\mathit{tree}]$ had an increasing trend. The DBH estimation is shown in Fig. \ref{fig:negative_dbh_2014_2}.

\begin{figure*}[tb]
    \centering
    \begin{subfigure}{\textwidth}
        \caption{\label{fig:jump_in_stemvolume_2024_1}}
        \includegraphics[width=\textwidth]{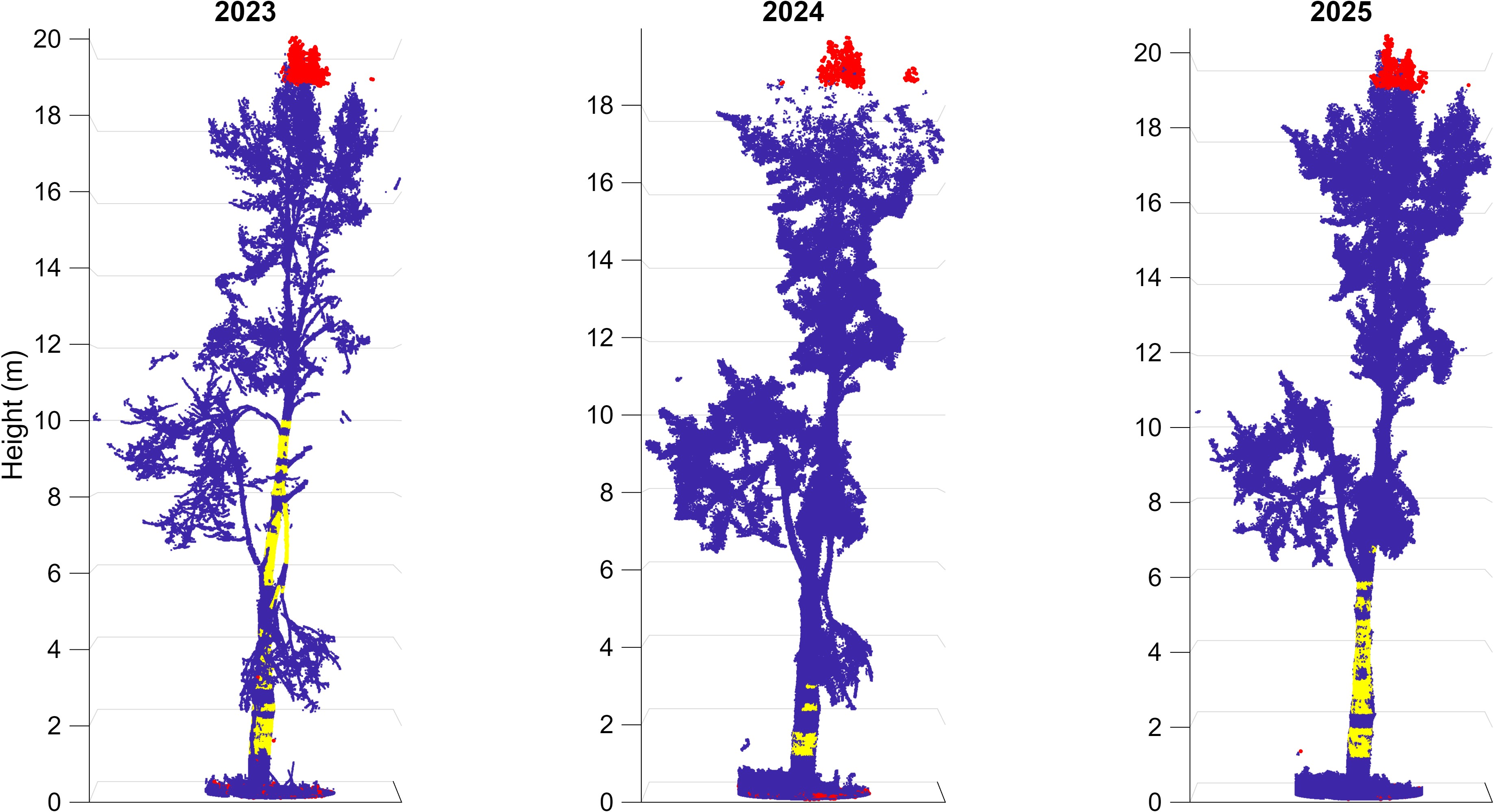}
    \end{subfigure} \\
    \begin{subfigure}{\textwidth}
        \centering
        \caption{\label{fig:jump_in_stemvolume_2024_2}}
        \includegraphics[width=0.7\textwidth]{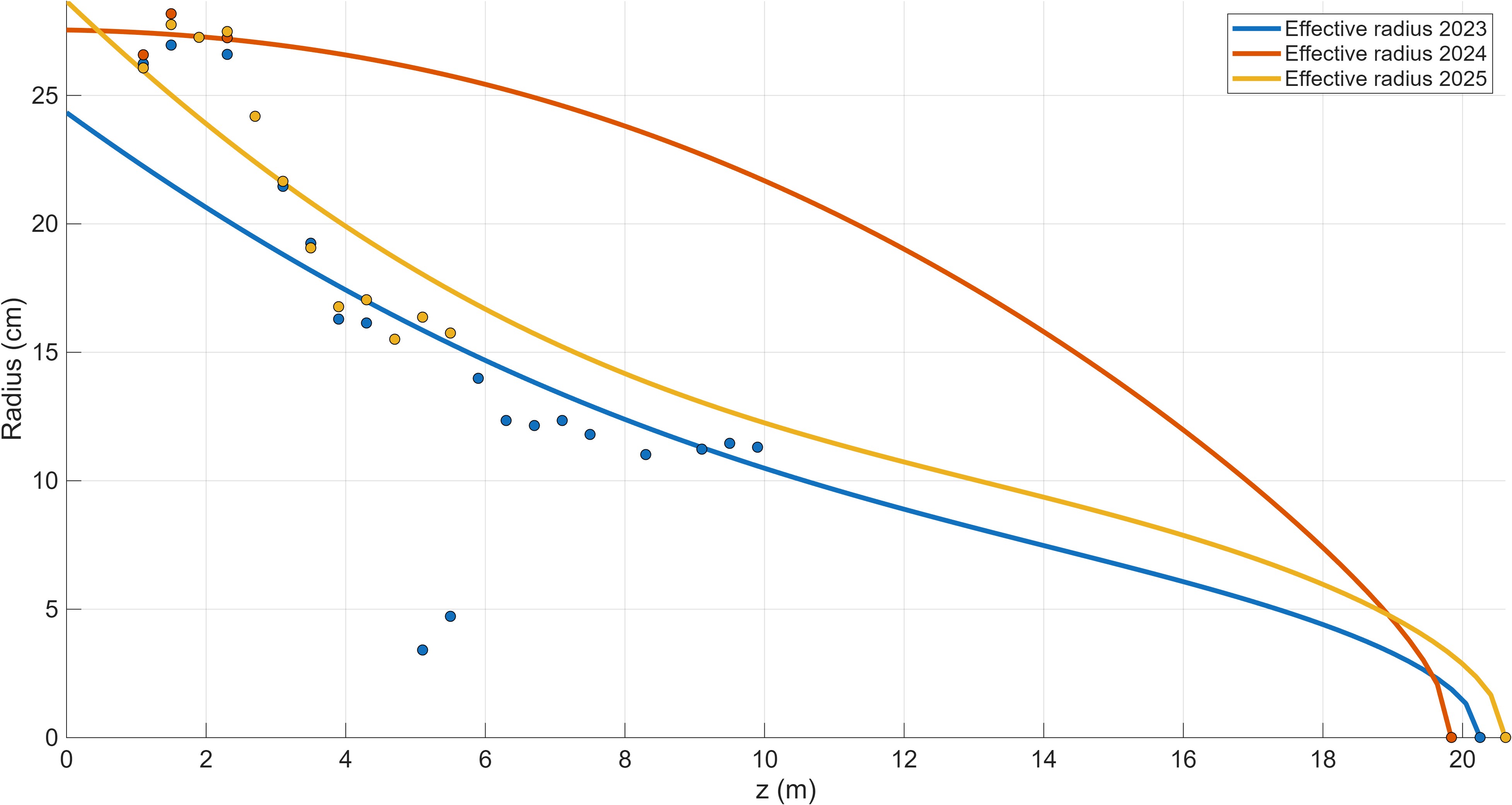}
    \end{subfigure} 
    \caption{\ref{fig:jump_in_stemvolume_2024_1}: The tree whose stem volume was overestimated in 2024. The dark blue points are from the MLS/TLS scanners, and the red points are from the ALS scanner. The yellow points represent the points from which the stem curve algorithm found arcs and constructed the stem radii estimates. In 2023 and 2025, the stem curves were found from a wider interval than in 2024. \ref{fig:jump_in_stemvolume_2024_2}: The effective radius $R_\textit{eff}$ of Eq. \eqref{eq:volume} plotted against the stem radii estimates. The limited number of radii estimates in 2024 resulted in an erroneous fit.}
\end{figure*}

The tree whose stem volume was overestimated in 2024 (Fig. \ref{fig:time_series6}) is plotted in Fig. \ref{fig:jump_in_stemvolume_2024_1}. The segmentation of the tree was successful, and the error originated from the stem curve algorithm. The algorithm excluded much of the points above 3 m, which caused an erroneous radius fit for stem volume estimation (Fig. \ref{fig:jump_in_stemvolume_2024_2}).

\end{appendices}
\end{document}